\PassOptionsToPackage{nameinlink}{cleveref}
\documentclass[mlmain,onecolumn,cleveref]{jmlr}

\jmlrproceedings{PMLR}{Proceedings of Machine Learning Research}

\usepackage{booktabs}
\usepackage{longtable}
\usepackage{multirow}
\usepackage{tikz}

\newcommand{\abs}[1]{\left|#1\right|}
\newcommand{\norm}[1]{\left\lVert#1\right\rVert}

\jmlrvolume{}
\firstpageno{1}
\editors{List of editors' names}
\jmlryear{2026}
\jmlrworkshop{Symmetry and Geometry in Neural Representations}

\title[The Tangled Web They Weave]{The Tangled Web They Weave: Exploring
Neural Representations with Directed Simplicial Complexes}

\author{\Name{Michael Bleher} \Email{mbleher@mathi.uni-heidelberg.de}\\
\addr Institute for Mathematics, Heidelberg University, Heidelberg, Germany}

\begin{document}

\maketitle

\begin{abstract}
Understanding how artificial neural networks (ANNs) represent and combine features internally is a central challenge in mechanistic interpretability.
Existing neural circuit-discovery methods represent neuron interactions as pairwise edges in a graph.
But distributed neural representations consist of large groups of neurons that are inherently linked by higher-order causal dependencies: suppressing a subgroup of neurons in concert may shift the activation of other neurons in the feature representation.
We propose a method to construct filtered directed simplicial complexes (DSCs) that give a topological representation of such dependencies, built from the activation profiles of an ANN on a given set of inputs.
Our approach treats ANNs as structural causal models and uses counterfactual ablations to test for causal links between groups of internal neurons within specific input contexts.
As a proof of concept, we apply the approach to a pair of small multilayer perceptron (MLP) architectures in a synthetic two-dimensional quadrant classification task.
In these examples, DSCs recover the designed feature pathways and coordinated class transitions, and the complexes of trained networks are higher-dimensional than those of their handcrafted references.
As validation, we compare our method with graph-based circuit-discovery baselines and find that their 1-skeletons agree within the relations that a weight graph can represent.
These results support DSCs as context-dependent descriptions of multi-neuron computation and may thus provide a novel, orthogonal framework to study the phenomenon of distributed feature representations and polysemanticity.

\end{abstract}

\begin{keywords}
distributed neural representations, feature superposition, directed
algebraic topology, simplicial complexes, causal inference
\end{keywords}

\section{Introduction}
\label{sec:introduction}

Understanding how information propagates and organises within artificial neural networks (ANNs) is the main goal in interpretability research.
While ANNs demonstrably encode semantic information through intricate patterns of neuron activations \citep{Alain2017,Kim2018a}, how they arise and interact remains an area of active investigation.
Methods that can identify and characterise such patterns can offer deep insights into how ANNs represent abstract concepts and perform complex computations.

A central challenge in understanding this \textit{neural code} arises from the efficient encoding of features on a finite number of neurons.
More often than not, features in the data have \textit{distributed representations} in the neural network, relying on coordinated activity across an ensemble of neurons \citep{Hinton1986,Olshausen1997,Pouget2000}.
Conversely, an individual neuron may participate in representing multiple distinct features; such a neuron is called \textit{polysemantic}\footnote{From the Greek \textit{poly} (many) and \textit{semantic} (meaning).} \citep{Olah2017,Olah2020}.
To highlight the duality of these concepts, we will call a feature \textit{polytopic}\footnote{From the Greek \textit{poly} (many) and \textit{topos} (place).} when its representation is distributed across multiple neurons: a polysemantic neuron participates in many features, whereas a polytopic feature is represented across many neurons.
When the representations of two or more polytopic features overlap on shared polysemantic neurons, the features are said to be in \textit{superposition} \citep{Elhage2022}.

Polytopic features are realised by coordinated activity across groups of neurons.
As a consequence, reducing the activation of one or more neurons in the group can affect the feature representation as a whole in a way that is not captured by perturbing the members of the group separately.
Such \emph{higher-order interactions} are naturally represented by directed hypergraphs or directed simplicial complexes \citep{Abiad2026}.
A general hypergraph construction would have to test for a causal link between every pair of disjoint subsets of the $N$ internal neurons, yielding $O(3^N)$ candidate relations.
We therefore rely on the subclass of \textit{directed simplicial complexes} (DSCs), which can be constructed inductively from the vertex set $V$ and require the slightly more amenable number of $O(N^{k+1})$ candidates when probing interactions of order at most $k$.

\begin{figure}
    \centering
    \includegraphics[width=\textwidth, height=.3\textheight, keepaspectratio]{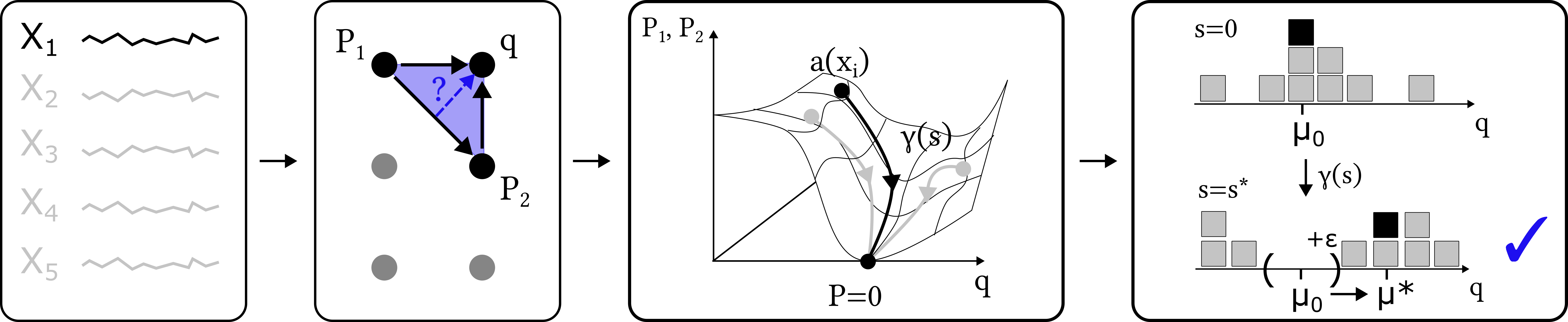}
    \caption{Construction of a filtered directed simplicial complex $\mathcal{K}_X$ from the activations of an input sample $X=\{x_b\}_{b=1}^B$.
    \textbf{Panel 1.} We start with the input sample $X$ and map it to the internal activation profiles $\{a(x_b)\}_{b=1}^B$.
    \textbf{Panel 2.} The internal neurons are the vertex set $V=\{n_i\}_{i\in I}$, from which candidate simplices are constructed inductively.
    \textbf{Panel 3.} For an admitted simplex with predictor set $P$ we test whether it's causal link extends to a target neuron $n_q$ by analysing its activation values along a coherent counterfactual ablation (CCA) trajectory $\gamma_{a(x),P}(s)$ that progressively suppresses $P$.
    \textbf{Panel 4.} If the activation changes significantly, the candidate simplices are admitted and labelled by the last persistent crossing to significance $s^*$; lower values indicate more immediate functional interactions in the context of $X$.}
    \label{fig:figure1}
\end{figure}

In this work, we introduce a method to construct \textit{filtered directed simplicial complexes} (DSCs) $\mathcal{K}_X$ from the activation profiles and parameters of a pre-trained feedforward multilayer perceptron (MLP) for a given set of input data $X=\{x_b\}_{b=1}^B$.\footnote{Code is available at \url{https://github.com/subthaumic/neural-dsc} \citep{Bleher2026code}.}
Its vertices are the internal neurons $V$, while its directed edges and higher-order simplices represent higher-order causal relations between neuron activations in the context of $X$.
For causal inference we rely on the framework of structural causal models by \citet{Pearl2010}.
More concretely, we introduce \emph{coherent counterfactual ablations} (CCAs) on the internal neurons of the network, trajectories of neuron activations that satisfy the networks constraints (see \Cref{subsec:counterfactual_generation}).
On the one hand, if an input $x$ activates multiple features, ablating a group of neurons $P\subseteq V$ that are part of a single feature representation should primarily affect other neurons $n_q\in V\setminus P$ within that same feature's processing pathway.
So, as long as off-feature pathways remain largely undisturbed, a directed edge $P\to n_q$ is evidence that $n_q$ is part of the same polytopic feature as the neurons in $P$.
On the other hand, ablating a polysemantic neuron is expected to influence multiple pathways simultaneously, which suggests that a large number of outgoing edges could be a signature of polysemanticity.
We assign a filtration value to each simplex $\sigma$ that quantifies the required coordinated reduction of activations over $P$ to induce a significant change in the activation of the target neuron $n_q$.
\Cref{fig:figure1} summarises the resulting construction.

Our main contributions are:
\begin{itemize}
    \item We construct filtered DSCs $\mathcal{K}_X$ from the activation data of MLPs with respect to input data $X$ using causal inference.
    \item We apply the method to a synthetic two-dimensional classification task with known reference computations, and show that the resulting DSCs recover the expected higher-order interactions between neurons. 
    \item We compare the 1-skeleton of $\mathcal{K}_X$ with the circuits found by established methods, namely ACDC \citep{Conmy2023}, EAP \citep{Syed2023}, and EAP-IG \citep{Hanna2024}, and find that they agree on those edges that can appear in both.
\end{itemize}

\subsection*{Related work}

\paragraph{Circuit discovery.}
A large body of work seeks the subnetwork responsible for a given task behaviour by pruning or scoring the connections of the network.
ACDC removes edges one at a time by iterative activation patching, keeping those whose removal degrades the behaviour \citep{Conmy2023}, while attribution methods replace that iteration with a gradient estimate of each edge's patch effect --- EAP from a single forward and backward pass \citep{Syed2023}, and EAP-IG by integrating that estimate along the path from a corrupted to a clean input \citep{Hanna2024}.
These are the closest empirical comparators to our construction, and we evaluate our method against all three (\Cref{sec:methods_comparisons}).

\paragraph{Causal methods in interpretability.}
Probing a network's internal state through causal reasoning is a well-established method.
Causal mediation identifies internal variables that mediate an input--output effect \citep{Vig2020}, while causal abstraction tests whether network states implement a hypothesised causal model, including variables represented in learned subspaces \citep{Geiger2024,Geiger2025}.
Manifold-aligned counterfactual explanations likewise constrain perturbations to produce only attainable data states \citep{Tsiourvas2024}.
Our approach relates to these approaches but differs in its use of counterfactual paths to suppress internal neurons and then constructs joint relations bottom-up rather than testing a supplied concept or causal abstraction.

\paragraph{Higher-order neural structures and topology.}
The phenomenon of distributed representations and polysemanticity, and thus higher-order interactions, also arises in biological neural networks (BNNs), where individual neurons may respond to various different tasks, a phenomenon called \textit{mixed selectivity} \citep{Fusi2016}.
Directed simplicial complexes in the connectome have revealed high-dimensional organisation in cortical microcircuits \citep{Reimann2017,EgasSantander2024}.
\citet{Chung2026} use counterfactually reasoning about network topology, modelling disruption and intervention in brain networks as perturbations of network flow and using Hodge theory to separate dissipative from persistent communication.
For artificial networks, \citet{Chowdhury2019} characterise the path homology of feedforward architectures, separating the topology implied by a network's wiring from the structure its parameters realise.
More broadly, topological analyses of neural systems are surveyed by \citet{Curto2025} for BNNs and \citet{Ballester2026} for ANNs.
Most such constructions derive topology from connectivity, activations, or correlations; our simplices are admitted by counterfactual relevance tests and therefore encode context-conditioned functional relations.

\section{Background}
\label{sec:background}

\subsection{Neural nets and neuron activations}
\label{subsec:neural_activations}

Consider a feedforward multilayer perceptron (MLP) with $L$ layers, indexed $l=0,\ldots,L$.
Layer $l$ comprises $N_l$ neurons.
The input layer is $l=0$, with activation vector $a^{(0)}(x)=x\in\mathbb{R}^{N_0}$.
The activation vectors $a^{(l)}(x)\in\mathbb{R}^{N_l}$ are then computed recursively via
\begin{align}
    \label{eq:mlp_activation}
    a^{(l)}(x) &= \sigma^{(l)}(W^{(l)} a^{(l-1)}(x) + b^{(l)} ),
\end{align}
where $W^{(l)}\in\mathbb{R}^{N_l\times N_{l-1}}$ are the weights, $b^{(l)}\in\mathbb{R}^{N_l}$ are the biases, and $\sigma^{(l)}$ are non-linear, element-wise activation functions.
Typically the activation functions are the same across layers (e.g., ReLU), except for the output layer $\sigma^{(L)}$ (e.g., softmax for classification).
The output layer's activation vector $a^{(L)}(x)$ represents the final decision or prediction of the network.

Our analysis focuses on the internal state of the network when evaluated on a given input $x$, represented by the concatenated activation vector of all hidden neurons:
\begin{align}
    a(x)^T = (a^{(1)T}(x), a^{(2)T}(x), \ldots, a^{(L-1)T}(x)) \in \mathbb{R}^N,
\end{align}
where $N=\sum_{l=1}^{L-1}N_l$ is the total number of hidden neurons.
For a fixed network, the set of attainable internal activation vectors forms a subset $\mathcal{M}=\{a(x):x\in\mathbb{R}^{N_0}\}\subseteq\mathbb{R}^N$, which we call the \textit{neural manifold}\footnote{In our setting, the neural manifold is a piecewise-linear manifold, not a smooth one.} following \citet{Stopfer2003}.

The vector $a(x)\in\mathcal{M}$ is the network's internal representation of the input $x$ and must encode the features and their interactions that are relevant to the network's task.
The activation vector $a(x)$ of a complex input may therefore reflect a superposition of multiple, interacting feature representations \citep{Elhage2022}.

\subsection{Structural causal models and counterfactual reasoning}
\label{subsec:pearl}

To reason about the influence of one neuron on another within the network's operational constraints, we rely on Pearl's framework for causal inference \citep{Pearl2010}.
This framework formalizes structural causal models (SCMs), which consist of a directed acyclic graph (DAG) whose nodes correspond to exo- or endogenous (random) variables and whose edges denote direct causal relationships, together with structural equations of the form $Z=f_Z(\mathrm{Parents}(Z))$ specifying how the value of a variable $Z$ depends on the values of its parent nodes and noise.
MLPs map naturally onto (deterministic) SCMs: the architecture is a DAG, the input nodes are exogenous variables, the internal neuron activations $a^{(l)}_i$ are endogenous variables, and the functions $f_Z$ are determined by $W^{(l)}$, $b^{(l)}$, and $\sigma^{(l)}$ as in \eqref{eq:mlp_activation}.

To investigate the causal effects between variables, Pearl's framework introduces two key tools: \textit{interventions} and \textit{counterfactuals}.
An \emph{intervention} is a controlled manipulation in the real world to test an unknown SCM; if an intervention on $Z$ results in a change in some downstream variable $Y$, one can infer a structural causal effect of $Z$ on $Y$.
In an ANN, interventions on individual neurons reveal a causal link to all downstream neurons in the network, thus reproducing the network's underlying architecture.

In this article we instead rely on \textit{counterfactuals}, which explore hypothetical outcomes of individual observations under altered conditions (antecedents): given the observed values of the endogenous variables, what would have happened to the value of $Y$ if $Z$ had been $z'$ instead of $z$?
Counterfactual reasoning proceeds in three steps: \textit{abduction} identifies the most likely values of the exogenous variables that produced the observed outcome, the \textit{action} step modifies the value of a set of chosen endogenous variables, and \textit{prediction} propagates these altered values through the SCM.
The resulting counterfactual state then serves as the basis for subsequent reasoning, e.g., to decide whether the proposed change in $Z$ would have significantly altered $Y$.
The coherent counterfactual ablations of \Cref{subsec:counterfactual_generation} implement exactly these three steps, by moving the input and propagating it through the unchanged network.

\subsection{Filtered directed simplicial complexes}
\label{subsec:directed_simplicial_complexes}

A \textit{directed $k$-simplex} on a vertex set $V$ is an ordered $(k+1)$-tuple $\sigma=(v_0,\dots,v_k)$ of distinct elements of $V$.
A \textit{face} of $\sigma$ is a directed $j$-simplex $\tau=(v_{r_0},\dots,v_{r_j})$, where $j<k$ and $0\leq r_0<\dots<r_j\leq k$.
A \textit{directed simplicial complex} $\mathcal{K}$ on $V$ is a collection of directed simplices closed under taking faces.
Vertices $v\in V$ are 0-simplices, and a directed edge $v\to w$ corresponds to the 1-simplex $(v,w)$.

A \textit{filtration} on $\mathcal{K}$ assigns a real number $f_\sigma$ to each simplex $\sigma\in\mathcal{K}$, satisfying $f_\tau\leq f_\sigma$ whenever $\tau$ is a face of $\sigma$.
This induces a family of nested subcomplexes $\mathcal{K}_{\le s}=\{\sigma\in\mathcal{K}:f_\sigma\leq s\}$ indexed by the filtration value $s$.

\section{Methods}
\label{sec:methods}

For a given set of inputs $X=\{x_b\}_{b=1}^B$, let $I=\{1,\ldots,N\}$ and write $V=\{n_i\}_{i\in I}$ for the set of internal neurons.
We construct a filtered DSC $\mathcal{K}_X$ on $V$ that models the distributed representation of $X$'s features in a trained MLP.
Conceptually, it probes the activation patterns $\{a(x_b)\}_{b=1}^B$ for polytopic and polysemantic behaviour by ablating a set of neurons and observing the rest of the network: a neuron dragged along by the ablation is evidence that it and the ablated set jointly encode a common feature of $X$.

Setting activations to zero directly produces an activation vector off the neural manifold $\mathcal{M}$.
Such an intervention will distort the delicate coordination of neuron activations in a feature representation: altering a single neuron may deform the coordinated pattern representing a polytopic feature beyond the network's recognition, and ablating a polysemantic neuron may disturb several features at once.
We therefore introduce \textit{coherent counterfactual ablations} (CCA) -- on-manifold paths of hypothetical activation states in $\mathcal{M}$ along which the activation on the set $P$ is progressively suppressed.

The construction of $\mathcal{K}_X$ involves three main operations, detailed in \Cref{sec:construction}:
\begin{enumerate}
    \item \textbf{Coherent counterfactual ablation (CCA) (\Cref{subsec:counterfactual_generation}):}
    For each $x\in X$ and predictor set $P$, generate an on-manifold trajectory $\gamma_{a(x),P}(s)\in\mathcal{M}$ that progressively suppresses $P$.
    The normalised coordinate $s\in[0,1]$ records the fraction of predictor activation removed, with $s=1$ denoting full ablation.
    \item \textbf{Relevance testing (\Cref{subsec:relevance_test}):}
    Test whether the paired change in a target neuron $n_q$ exceeds a target-scaled relevance margin across $X$.
    A link $P\to n_q$ is admitted when its full-ablation test survives dimension-wise Bonferroni correction \citep{Dunn1961}; its birth value $s^*(P\to n_q)$ is the last persistent crossing of that cutoff along the trajectory.
    \item \textbf{Construction of filtered DSC (\Cref{subsec:dsc_construction}):}
    Initialise $\mathcal{K}_X$ with vertex set $V$ at filtration value zero and proceed inductively in increasing dimension.
    At dimension $k\geq1$, extend each admitted $(k-1)$-simplex $(n_{i_0},\dots,n_{i_{k-1}})$ by a candidate target $n_q$, put $P=\{n_{i_0},\dots,n_{i_{k-1}}\}$, and apply the two operations above to $P\to n_q$.
    Admit $\sigma=(n_{i_0},\dots,n_{i_{k-1}},n_q)$ only when that link is admitted and all codimension-one faces of $\sigma$ are already present, then set $f_\sigma$ to the maximum of $s^*(P\to n_q)$ and the filtration values of those faces.
    Repeat until no candidate is admitted or the chosen dimension limit is reached.
    The test depends on $P$ only as a set, so the admitted orderings of a given predictor set differ only in which faces they require; simplex counts therefore count ordered tuples rather than distinct joint interventions.
\end{enumerate}

We compare DSCs built from different contexts or networks on the same architecture by a filtration-integrated Jaccard distance $d_J$ (\Cref{subsec:jaccard_distance}).

\section{Results}
\label{sec:results}

We test our method on a synthetic two-dimensional classification task of points in the plane, with four classes corresponding to the four quadrants of the plane.
We study two handcrafted multilayer perceptrons whose internal computations are known exactly: the \emph{monosemantic} $[2,4,4,4]$ network, whose four first-layer neurons detect the four elementary features\footnote{This assignment presupposes that the \emph{objectively relevant features} for the task at hand are the signs of $x$ and $y$. Relative to this choice, a neuron computing $\operatorname{relu}(x)$ is monosemantic, whereas $\operatorname{relu}(x+y)$ is polysemantic as it is active for both $x>0$ and $y>0$; in a rotated input basis the same weights would support the opposite reading. Any assignment of semantics is necessarily ours.} $x>0$, $x<0$, $y>0$ and $y<0$ separately, and the \emph{polysemantic} $[2,3,4,4]$ network, whose narrower first layer forces its three units to encode mixtures of both coordinates instead (\Cref{subsec:quadrant_networks}).
We compare them with trained and randomly initialised networks of the same two architectures (\Cref{subsec:quadrant_trained,subsec:quadrant_untrained}).

\paragraph{Sensitivity to hyperparameters.}
The DSC construction requires us to specify a number of hyperparameters that control the input distribution, the spacing of the ablation values, the relevance test, and the multiple-testing correction.
We swept these parameters for the handcrafted neural nets and found that the resulting DSCs are reasonably robust to the spacing of the ablation values and to the relevance margin, while the multiple-testing correction occasionally reduces the number of admitted edges, with considerable knock-on effects on the number of higher-dimensional simplices.
The construction is somewhat sensitive to the choice of input distribution, which is to be expected given that the input distribution directly determines the 'features' the network must represent and react to.
See \Cref{sec:statistical_sensitivity} and \Cref{sec:sensitivity} for full details.

\begin{figure}[t]
	\centering
	\begin{tikzpicture}[x=\linewidth,y=\linewidth,>=latex]
		{ 
			\draw[draw=black!25,fill=black!1,line width=.55pt,rounded corners=5pt]
				(0.005,0.410) rectangle (0.995,0.410+0.395);
			\node[anchor=north,font=\small\bfseries]
				at (0.500,0.410+0.386) {(a) Monosemantic classifier};
			\node[anchor=south,font=\footnotesize]
				at (0.125,0.410+0.318) {$Q_1$ input set};
			\node[anchor=south,font=\footnotesize]
				at (0.440,0.410+0.318) {Mean neuron activations};
			\node[anchor=south,font=\footnotesize]
				at (0.812,0.410+0.318) {Filtered DSC $\mathcal{K}_{Q_1}$};

			\node[anchor=south west,inner sep=0pt]
				at (0.020,0.410+0.055)
				{\includegraphics[width=.210\linewidth]{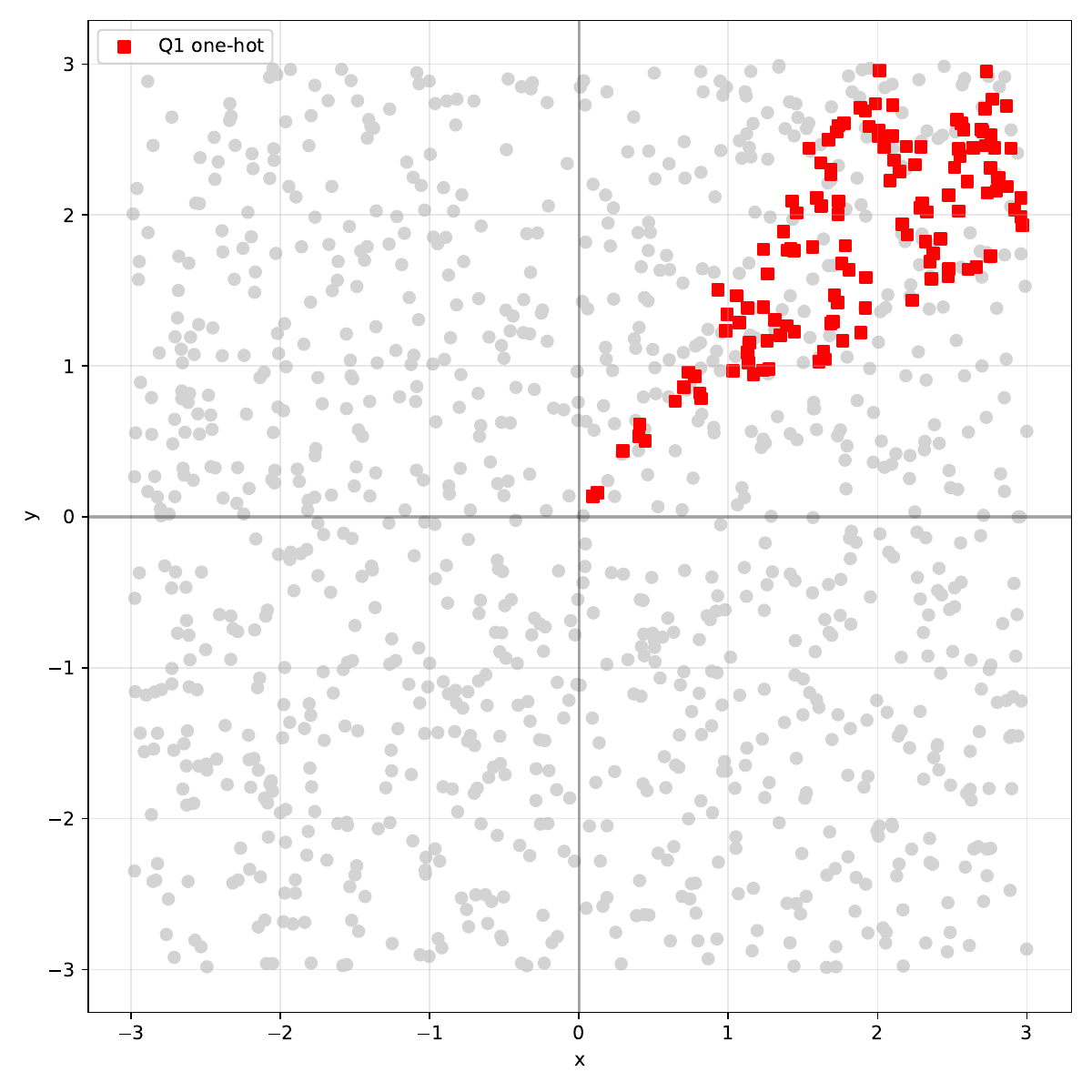}};
			\node[anchor=south west,inner sep=0pt]
				at (0.285,0.410+0.008)
				{\includegraphics[width=.310\linewidth]{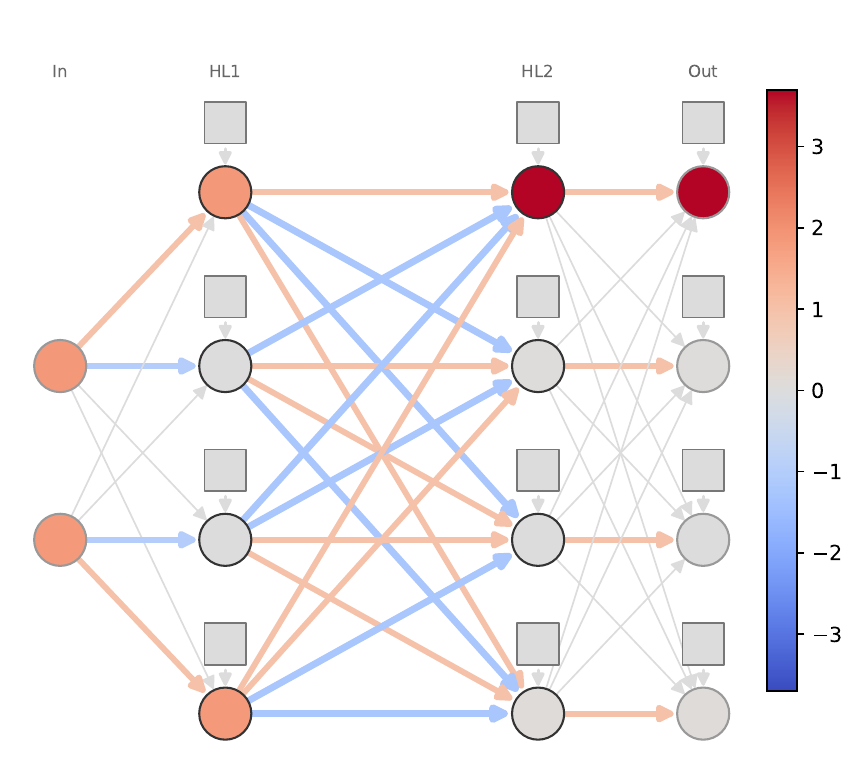}};
			\node[anchor=south west,inner sep=0pt]
				at (0.700,0.410+0.005)
				{\includegraphics[width=.245\linewidth]{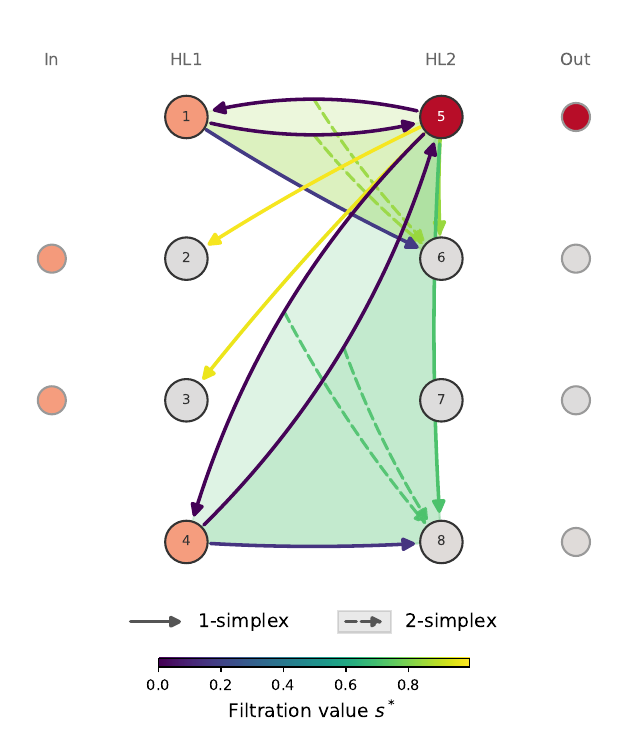}};

			\draw[->,line width=.8pt,draw=black!65]
				(0.237,0.410+0.175) -- (0.275,0.410+0.175);
			\draw[->,line width=.8pt,draw=black!65]
				(0.607,0.410+0.175) -- (0.685,0.410+0.175);
		}
		{ 
			\draw[draw=black!25,fill=black!1,line width=.55pt,rounded corners=5pt]
				(0.005,0.000) rectangle (0.995,0.000+0.395);
			\node[anchor=north,font=\small\bfseries]
				at (0.500,0.000+0.386) {(b) Polysemantic classifier};
			\node[anchor=south,font=\footnotesize]
				at (0.125,0.000+0.318) {$Q_1$ input set};
			\node[anchor=south,font=\footnotesize]
				at (0.440,0.000+0.318) {Mean neuron activations};
			\node[anchor=south,font=\footnotesize]
				at (0.812,0.000+0.318) {Filtered DSC $\mathcal{K}_{Q_1}$};

			\node[anchor=south west,inner sep=0pt]
				at (0.020,0.000+0.055)
				{\includegraphics[width=.210\linewidth]{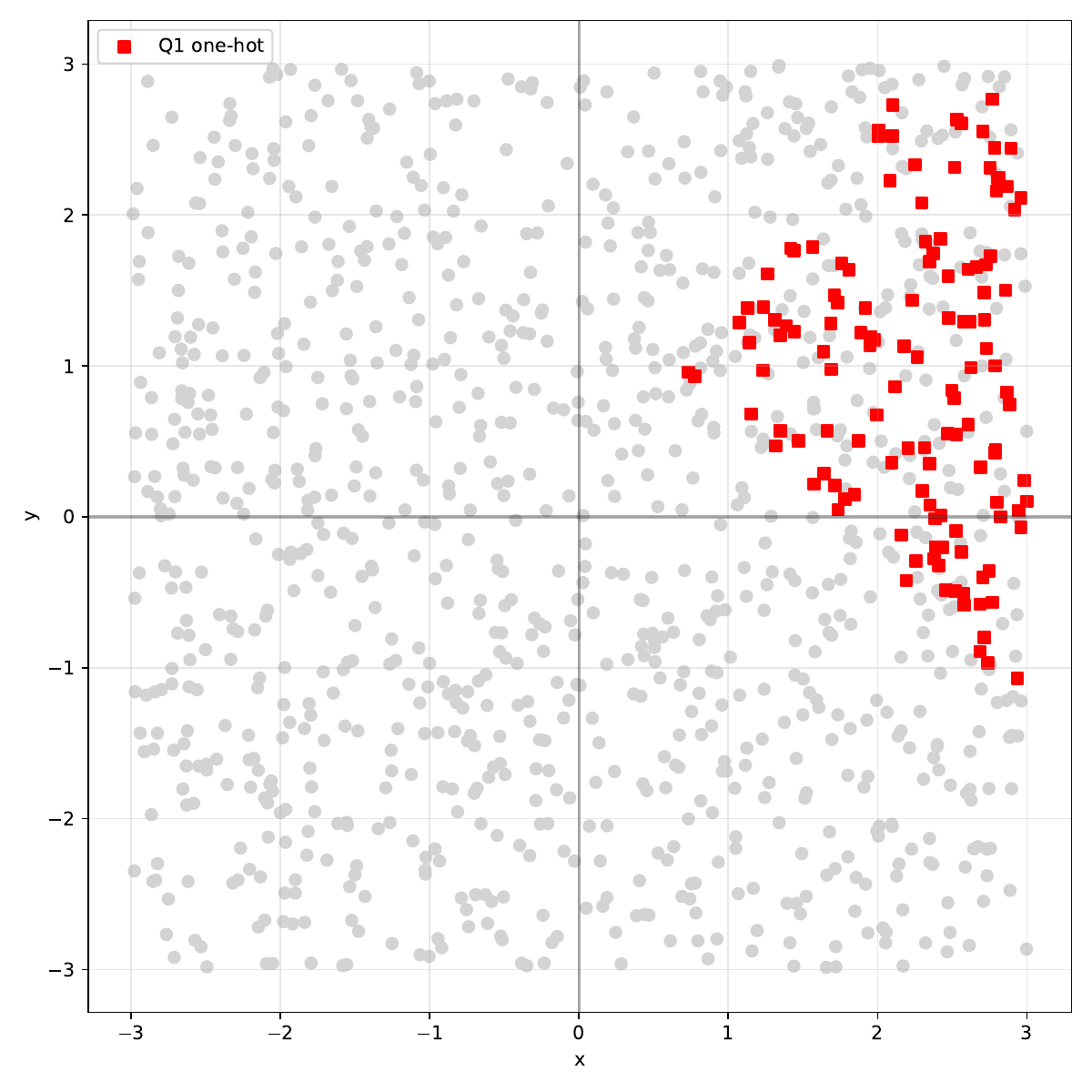}};
			\node[anchor=south west,inner sep=0pt]
				at (0.285,0.000+0.008)
				{\includegraphics[width=.310\linewidth]{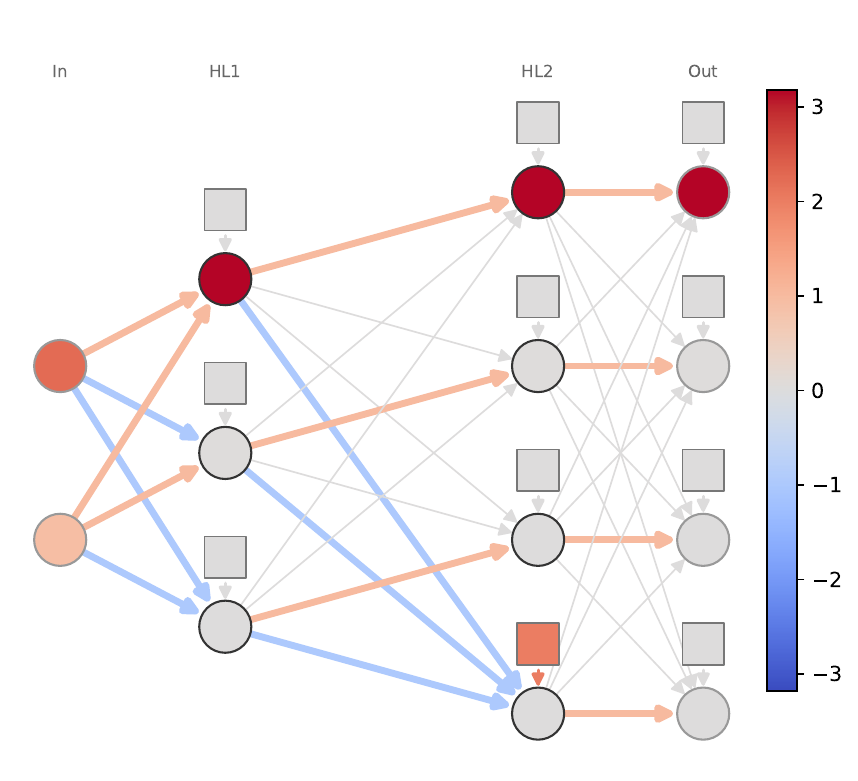}};
			\node[anchor=south west,inner sep=0pt]
				at (0.700,0.000+0.005)
				{\includegraphics[width=.245\linewidth]{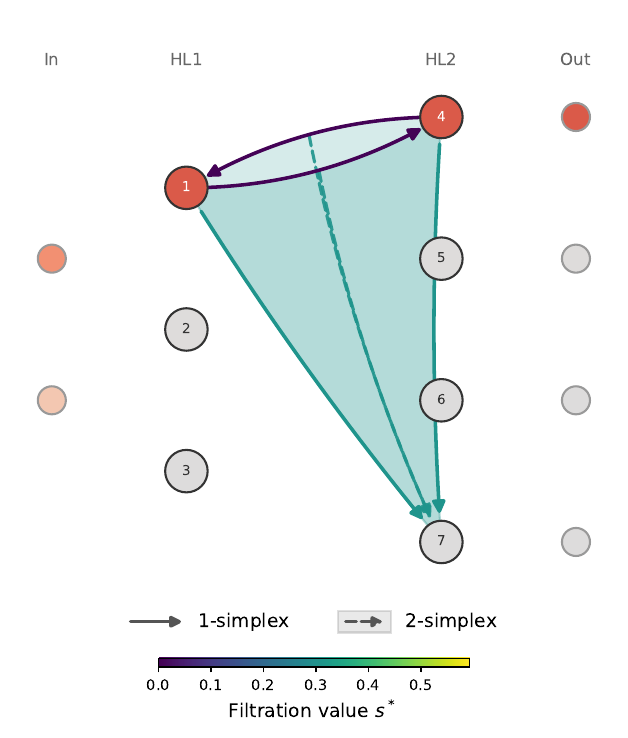}};

			\draw[->,line width=.8pt,draw=black!65]
				(0.237,0.000+0.175) -- (0.275,0.000+0.175);
			\draw[->,line width=.8pt,draw=black!65]
				(0.607,0.000+0.175) -- (0.685,0.000+0.175);
		}
	\end{tikzpicture}
	\caption{Illustrative results for the handcrafted quadrant classifiers.
	Each panel shows the $Q_1$ input set, the mean activations of the neurons on these inputs, and the resulting filtered DSC.
	In the monosemantic classifier, the edges with the lowest filtration values connect the class neuron of $Q_1$ in both directions with the two first-layer neurons that detect $x>0$ and $y>0$.
	In the polysemantic classifier, they connect the active first-layer neuron with its copy in the second layer.
	See \Cref{subsec:handcrafted_dsc_logic} for a thorough discussion of these edges.}
	\label{fig:results_quadrants}
\end{figure}

\paragraph{DSCs recover designed computations in handcrafted networks.}
For the monosemantic network, the DSC of each one-hot input set (\Cref{subsec:probe_contexts}) contains ten edges and four 2-simplices, and the four DSCs coincide up to a relabelling of the neurons (\Cref{subsec:handcrafted_dsc_logic}).
The four edges with the lowest filtration values connect the class neuron of the quadrant in both directions with the two first-layer neurons that detect the signs of $x$ and $y$ there (\Cref{fig:results_quadrants}).
The 2-simplices enter later and describe how the ablation moves the inputs towards the adjacent quadrants.
For the polysemantic network, the edges that enter first connect each active first-layer neuron with its copy in the second layer, and every 2-simplex of a one-hot DSC has as its target the second-layer neuron of $Q_4$.
Its one-hot DSCs contain between four and twelve edges and between two and four 2-simplices, because the inputs of $Q_2$ and $Q_4$ activate different neurons on the two sides of $x+y=0$.

\paragraph{Jaccard distance between DSCs identifies class mixtures.}
In both handcrafted networks, the DSC of the two-hot inputs of two adjacent classes is closer to the one-hot DSCs of these two classes than to those of the other two, with a separation of at least $0.16$ in the monosemantic and $0.22$ in the polysemantic network (\Cref{subsec:twohot_distances}).
The Jaccard distance therefore identifies the two classes that are mixed in a two-hot input set.
For the trained networks, it does so for only three of the four two-hot input sets.

The Jaccard distance between DSCs does not simply reflect the distance between the mean activations of the input sets.
In the monosemantic network, for example, the cosine distance between the mean activations is smaller for adjacent than for opposite quadrants, while the Jaccard distance is the same for all pairs, because the four one-hot DSCs share no simplex (\Cref{subsec:onehot_distances}).

\paragraph{Trained and untrained networks have higher-order structure.}
The DSCs of both trained networks reach dimension four for two of their four one-hot input sets, whereas those of the handcrafted networks have no simplices above dimension two.
The trained DSCs are also larger, with up to $365$ simplices against at most $23$ for the handcrafted networks (\Cref{subsec:quadrant_trained}).
Unlike those of the handcrafted monosemantic network, the DSCs of the trained $[2,4,4,4]$ network differ between the quadrants, with $24$, $220$, $132$ and $38$ simplices.
However, randomly initialised networks give DSCs of comparable size, and three of their $24$ DSCs reach dimension four (\Cref{subsec:quadrant_untrained}).
For the $[2,4,4,4]$ architecture, five of the twelve untrained DSCs are even larger than the trained DSC of the same quadrant.
The size and dimension of a DSC therefore do not show that its structure was learned.

\paragraph{The 1-skeleton is compatible with established methods.}
We compared the handcrafted DSC 1-skeletons with ACDC \citep{Conmy2023}, EAP \citep{Syed2023}, and two EAP-IG variants \citep{Hanna2024}, keeping for each method the same number of connections.
A circuit can contain an edge only if its predictor feeds its target directly, and such edges make up $40\%$ of the edges of the monosemantic DSCs and between $33\%$ and $50\%$ of those of the polysemantic DSCs (\Cref{subsec:representable_ceiling}).
In $26$ of the $32$ comparisons, the method keeps all of these edges.
The scores of the methods also rank these edges above the other connections, with a mean AUC of $0.86$ and a minimum of $0.65$.
The remaining DSC edges, and all simplices of dimension two or higher, have no direct analogues in established methods.
Further details are in \Cref{sec:methods_comparisons}.

\section{Discussion}
\label{sec:discussion}

In this article we have introduced a method to construct filtered DSCs from the internal activations of a feedforward multilayer perceptron (MLP) in response to a set of inputs.
We expected a simplex to correspond to a group of neurons jointly encoding one feature of the input, and its dimension to reflect across how many neurons that feature is distributed.
In the handcrafted networks, we found that the simplices with low filtration values follow the designed computation, while those that enter later describe how the ablation moves the inputs towards neighbouring classes.
A DSC therefore describes a network's functional response to a distribution of inputs.
Moreover, the DSCs of a fixed network differ markedly between one-hot (single-class), two-hot (mixed-class) and random inputs.
In the handcrafted networks, a two-hot DSC is not the union of the DSCs of its two classes, but it is closer to both of them than to those of the other two classes (\Cref{subsec:handcrafted_dsc_logic} and \Cref{sec:input_set_comparison}).
Finally, the circuits found by established circuit-discovery methods agree with the DSC on those edges that can appear in both methods (\Cref{sec:methods_comparisons}).

However, two of the questions that motivated this construction remain open.
The first is whether the DSCs identify polysemantic neurons.
In the polysemantic network, the neuron that belongs to the most maximal simplices is the one that represents $Q_4$, whereas the six neurons that mix both coordinates belong to fewer (\Cref{subsec:handcrafted_dsc_logic}).
Since this neuron is inhibited by all three first-layer neurons, its participation in many simplices could be read as polysemanticity with respect to the features they detect, which departs from our original choice of elementary features.
The second is whether higher-dimensional simplices show higher-order interactions between neurons.
Such simplices record the effect of a joint ablation of several predictors, which no circuit can represent (\Cref{subsec:representable_ceiling}).
But the relevance test we implemented checks only that the joint ablation has an effect, without comparing it with the effects of the individual predictors.
Moreover, a simplex is only tested if its faces are admitted, so a joint effect is missed when there is no edge between the predictors, as for the AND gate of $Q_1$ in the monosemantic network (\Cref{subsec:handcrafted_dsc_logic}).
Answering it requires a relevance test that compares the joint effect with the effects of the individual predictors, and both questions require further architectures and tasks for which the semantics of the neurons is known.

The study is confined to small ReLU MLPs on a synthetic task with known reference computations.
The number of candidate simplices grows as $O(N^{k+1})$, which severely limits the size of networks and the dimension of simplices, although restricting the candidates might lift this limitation (\Cref{sec:computational-complexity}).
For related reasons, the multiple-testing procedures are applied separately for each dimension of a DSC, so the error rate over all dimensions, input sets and networks together is not controlled.
The CCA trajectories are computed from the local Jacobian, so at the boundaries between ReLU regions they can fail to continue even when a fully ablated state exists elsewhere on the neural manifold (\Cref{subsec:counterfactual_generation}).

The next steps are to apply the construction to larger networks and to natural data.

\acks{%
This work was funded by the Deutsche Forschungsgemeinschaft (DFG, German Research Foundation) under Germany's Excellence Strategy EXC 2181/1 - 390900948 (the Heidelberg STRUCTURES Excellence Cluster) and finalized with the support of the European Research Council (ERC) under the European Union's Horizon 2020 research and innovation programme (project PEPS, no. 101071786).
The author acknowledges the use of compute resources from de.NBI Cloud at the Zentrum f\"ur Datenverarbeitung of the University of T\"ubingen and the German Federal Ministry of Education and Research (BMBF) through grant no 031 A535A.
Part of this research was performed while the author was visiting the Institute for Mathematical and Statistical Innovation (IMSI), which is supported by the National Science Foundation (Grant No. DMS-1929348).
I thank Freya Jensen for helpful discussions.

}

\bibliography{literature}

\clearpage

\appendix
\crefalias{section}{appendix}
\crefalias{subsection}{subappendix}
\jmlrpretitle Appendix\jmlrposttitle

\noindent
\Cref{sec:notation} provides a summary of the notation and conventions used throughout the paper.
The methodological and computational details are given in \Crefrange{sec:construction}{sec:computational-complexity}: the construction of the complexes (\Cref{sec:construction}), the quadrant classifiers and the input sets used in the experiments (\Cref{sec:quadrant_setup}), sensitivity to the relevance test (\Cref{sec:statistical_sensitivity}), sensitivity to numerical integration and input selection tolerance (\Cref{sec:sensitivity}), and computational complexity (\Cref{sec:computational-complexity}).
The experiments and their results are presented in \Crefrange{sec:quadrant_classifiers_details}{sec:methods_comparisons}: the complexes of the quadrant classifiers (\Cref{sec:quadrant_classifiers_details}), comparisons of complexes across input sets and with activation distances (\Cref{sec:input_set_comparison}), and comparisons with other interpretability methods (\Cref{sec:methods_comparisons}).
The code that generates all results, tables and figures of this article is available at \url{https://github.com/subthaumic/neural-dsc} \citep{Bleher2026code}.

\section{Notation and conventions}
\label{sec:notation}

\begin{longtable}{c|p{0.85\linewidth}}
	\caption{Symbols used in the main text and appendices.}
	\label{tab:appendix_notation}\\
	\toprule
	symbol & meaning \\
	\midrule
	\endfirsthead

	\caption[]{Symbols (continued).}\\
	\toprule
	symbol & meaning \\
	\midrule
	\endhead

	\midrule
	\multicolumn{2}{r}{\small Continued on next page}\\
	\endfoot

	\bottomrule
	\endlastfoot

	\addlinespace[0.75em]
	$N_0$ & number of input neurons \\
	$N$ & number of internal (hidden-layer) neurons; input and output
	layers are excluded throughout \\
	$I$ & internal-neuron index set, $I=\{1,\ldots,N\}$ \\
	$n_i$ & $i$-th internal neuron, $i\in I$ \\
	$V$ & set of internal neurons, $V=\{n_i\}_{i\in I}$ \\
	$x$ & input vector, $x\in\mathbb{R}^{N_0}$ \\
	$a_i(x)$ & activation of neuron $n_i$ at input $x$ \\
	$a(x)$ & vector of all internal neuron activations, $a(x)=(a_i(x))_{i\in I}\in\mathbb{R}^N$ \\
	$\mathcal{M}$ & neural manifold: attainable internal activation states
	$\{a(x):x\in\mathbb{R}^{N_0}\}\subseteq\mathbb{R}^N$ \\
	\addlinespace[0.75em]
	$X$ & set of inputs used to construct the complex \\
	$B$ & number of inputs, $B=\abs{X}$ \\
	$\eta$ & selection tolerance of the input set \\
	$P$ & predictor-neuron set $P\subseteq V$, suppressed jointly \\
	$n_q$ & target neuron, $q\in I$ and $n_q\in V\setminus P$ \\
	\addlinespace[0.75em]
	$\gamma_{a(x),P}$ & coherent counterfactual ablation (CCA) trajectory starting at
	$a(x)$ that suppresses $P$ \\
	$s$ & ablation coordinate, $s\in[0,1]$; $s=1$ is the fully ablated
	counterfactual \\
	$\mathrm{D}a(x)$ & derivative of the activation map at $x$, represented
	by its Jacobian \\
	$J$ & Jacobian of the internal activations with respect to the input \\
	$\Delta t$ & step size of the numerical integration of $x(t)$ \\
	$K$ & ablation values $s_k=k/K$, $k=0,\ldots,K$, at which the relevance test is evaluated \\
	\addlinespace[0.75em]
	$\Delta_{q,x}(s)$ & change in the activation of $n_q$ at ablation $s$ for input $x$ \\
	$\Delta_{q,X}(s)$ & changes in the activation of $n_q$ over the input set $X$ \\
	$\operatorname{med}_{\mathrm{HL}}$ & Hodges--Lehmann pseudomedian \\
	$\varepsilon$ & relevance margin, in units of the standard deviation of the
	target activation over $X$ \\
	$\varepsilon_q$ & relevance margin for target $n_q$ \\
	$\alpha$ & test level, applied separately at each dimension \\
	$c_d$ & cutoff for the p-values of the candidates at dimension $d$ \\
	$s^*$ & smallest ablation value from which the link $P\to n_q$ stays relevant
	up to full ablation \\
	\addlinespace[0.75em]
	$\sigma,\tau$ & simplices and their faces \\
	$f_\sigma$ & filtration value of $\sigma$ \\
	$\mathcal{K}$ & (filtered) directed simplicial complex\\
	$\mathcal{K}_{\le s}$ & its subcomplex of simplices with $f_\sigma\leq s$ \\
	$\mathcal{K}_X$ & the filtered directed simplicial complex constructed from the input set
	$X$ \\
	$\delta_J,\,d_J$ & Jaccard distance and its filtration-integrated form
	\eqref{eq:jaccard_distance} \\
\end{longtable}

\clearpage

\section{Construction of the directed simplicial complex}
\label{sec:construction}

Fix a feedforward multilayer perceptron with ReLU activations, input dimension $N_0$ and internal neurons $V=\{n_i\}_{i\in I}$, where $I=\{1,\ldots,N\}$.
Let $a:\mathbb{R}^{N_0}\to\mathbb{R}^N$ be its internal activation map and $\mathcal{M}=a(\mathbb{R}^{N_0})$ its neural manifold.
Given a finite, nonempty set of inputs $X\subseteq\mathbb{R}^{N_0}$, with $B=\abs{X}$, we construct a filtered directed simplicial complex (DSC) $\mathcal{K}_X$ on $V$ that encodes causal relationships between neuron activations in the context of $X$.

Let $P\subseteq V$ be a nonempty set of \emph{predictor} neurons and let $n_q\in V\setminus P$ be a \emph{target} neuron.
For each $x\in X$, we seek a \emph{coherent counterfactual ablation} (CCA) trajectory $\gamma_{a(x),P}: [0,1] \to \mathcal{M}$ that starts at $\gamma_{a(x),P}(0) = a(x)$ and decreases the predictor activations $a_P(x)$ at equal rates while remaining on the neural manifold (\Cref{subsec:counterfactual_generation}).
For each $s\in[0,1]$, the differences between the initial target neuron activation $a_q(x)$ and its value along the trajectory $\gamma_{a(x),P}(s)$ are tested across $x\in X$ against a relevance margin (\Cref{subsec:relevance_test}).
A candidate directed simplex $\sigma$, with the predictors in $P$ followed by the target neuron $n_q$, is included in $\mathcal{K}_X$ only when there is a statistically relevant causal link $P\to n_q$ after full ablation and all its faces are present.
Its filtration value $f_\sigma$ is the maximum of the ablation value $s^*(P\to n_q)$ from which the link stays relevant up to full ablation and the filtration values of its faces (\Cref{subsec:dsc_construction}).
Filtered DSCs $\mathcal{K}$ and $\mathcal{L}$ on $V$ are compared using a version of the Jaccard distance $d_J(\mathcal{K},\mathcal{L})$ (\Cref{subsec:jaccard_distance}).

\subsection{Generating coherent counterfactual ablation (CCA) trajectories}
\label{subsec:counterfactual_generation}

Fix an input $x_0\in X \subseteq \mathbb{R}^{N_0}$ and a nonempty set of predictor neurons $P\subseteq V$.
Let $\pi_P:\mathbb{R}^N\to\mathbb{R}^{\abs{P}}$ denote the projection onto the activations indexed by $P$.
We write $a_P=\pi_P\circ a$ for the activation vector restricted to the predictor set.
We seek a trajectory $\gamma:[0,1] \to \mathcal{M}$, starting at $a(x_0)$, along which the predictor activations $a_P$ are reduced to zero.
These conditions do not determine a unique trajectory, and the construction below introduces additional choices that make it deterministic.

Let the total activation of the predictor neurons be denoted by
\begin{align}
	h_P(x)
	&=\mathbf{1}^{\top}a_P(x)
	=\sum_{n_i\in P}a_i(x).
	\label{eq:predictor_height}
\end{align}
As a sum of ReLU network activations, $h_P$ is continuous and piecewise affine in the input.
Since ReLU activations are non-negative, $h_P(x)=0$ exactly when every predictor in $P$ is off.
At each point of the trajectory, let $P'\subseteq P$ contain the predictors that remain active and write $m=\abs{P'}$.
To reduce $h_P$ to zero, we use its normalised negative gradient with respect to these activations,
\begin{align*}
	u=-\frac{\mathbf{1}_m}{\sqrt{m}},
\end{align*}
which decreases the active predictors at equal rates and prevents the easiest-to-change predictor from dominating the descent.

Prescribing $u$ does not yet determine a unique trajectory on $\mathcal{M}$, since different trajectories can have the same velocity when projected onto the active predictor coordinates.
Since our intention is to `ablate a feature in the input', we construct an input trajectory $x(t)$, starting at $x(0)=x_0$, which determines the internal activations $a(x(t))$ through the network equations.

We first assume that $x$ lies in an open neighbourhood on which $a$ is differentiable as a function of the input.
Set $J_{P'}=\mathrm{D}a_{P'}(x)$, the Jacobian of the predictor activations with respect to the input.
By the chain rule, the velocity of the active predictor activations along $x(t)$ is $\frac{\mathrm{d}}{\mathrm{d}t}a_{P'}(x(t))=J_{P'}\frac{\mathrm{d}x(t)}{\mathrm{d}t}$.
Ordinary input-space gradient descent on $h_P$ would use $-\nabla h_P(x)=-J_{P'}^{\top}\mathbf{1}_m$.
To realise the prescribed velocity $u$, the input trajectory must instead satisfy
\begin{align}
	J_{P'}\frac{\mathrm{d}x(t)}{\mathrm{d}t}&=u.
	\label{eq:cca_desired_velocity}
\end{align}
Whenever this velocity is attainable, it gives $\frac{\mathrm{d}}{\mathrm{d}t}h_P(x(t))=\mathbf{1}_m^{\top}u=-\sqrt{m}$.
To minimise incidental input displacement, we choose the input velocity of smallest Euclidean norm among those that realise the prescribed activation velocity.

Since $J_{P'}:\mathbb{R}^{N_0}\to\mathbb{R}^m$ is generally rectangular and may be rank deficient, the solution is expressed through the Moore--Penrose pseudoinverse \citep{Penrose1955}:
\begin{align}
	v:=\frac{\mathrm{d}x(t)}{\mathrm{d}t}
	&=J_{P'}^+u
	=J_{P'}^{\top}(J_{P'}J_{P'}^{\top})^+u, \\
	J_{P'}v
	&=\operatorname{proj}_{\operatorname{im}J_{P'}}u, \\
	v
	&=\underset{w\in\arg\min_z\norm{J_{P'}z-u}_2}{\arg\min}
	\norm{w}_2.
\end{align}
In summary, the descent uses the unique minimum-norm exact solution when $u\in\operatorname{im}J_{P'}$ and the minimum-norm least-squares solution otherwise.
The pseudoinverse removes arbitrary motion in $\ker J_{P'}$, but it does not assert that activations outside $P$ remain fixed.

We now consider points where $a$ is not differentiable, which occur on the boundaries of the polyhedral regions in input space on which a fixed set of neurons is active.
To continue the descent, we choose a Jacobian of an adjacent region from the B-subdifferential.
Let $\mathcal{D}$ be the differentiability set of $a_{P'}$.
The B-subdifferential consists of the limiting Jacobians \citep{Scholtes2012}:
\begin{align}
	\partial_B a_{P'}(x)
	&=\left\{\lim_{r\to\infty}\mathrm{D}a_{P'}(x_r):
	x_r\in\mathcal{D},\ x_r\to x\right\}.
\end{align}
For each region we form $v=G^+u$ with $G\in\partial_B a_{P'}(x)$, retain it only when the directional changes of its boundary pre-activations make $v$ enter that same region, and choose the minimum-input-norm direction among the self-consistent continuations.

Finally, to compare trajectories across inputs, the integration parameter $t$ is replaced by a normalised ablation parameter $s$.
When the predictor activations decrease monotonically along a completed CCA, the decrease in $h_P$ is the $\ell^1$-arclength traversed by $a_P(x(t))$.
The normalised trajectory parameter is therefore
\begin{align}
	s
	&=1-\frac{h_P(x(t))}{h_P(x_0)}
	=1-\frac{\sum_{n_i\in P}a_i(x(t))}
	{\sum_{n_i\in P}a_i(x_0)}\in[0,1].
	\label{eq:ablation_coordinate}
\end{align}
Reparameterising $a(x(t))$ by $s$ gives the CCA trajectory $\gamma_{a(x_0),P}$, with $s=0$ the factual state and $s=1$ the fully ablated counterfactual.
A predictor set with $h_P(x_0)=0$ has the constant trajectory $\gamma_{a(x_0),P}(s)=a(x_0)$.

\subsection{Relevance testing}
\label{subsec:relevance_test}

For a predictor set $P\subseteq V$, target neuron $n_q\in V\setminus P$, input $x\in X$, and ablation $s\in[0,1]$, let the change in target activation be
\begin{align}
	\Delta_{q,x}(s)
	&=a_q(x)-\left[\gamma_{a(x),P}(s)\right]_q.
\end{align}
Here $[\gamma_{a(x),P}(s)]_q$ is the activation of the target neuron $n_q$ along the trajectory.
We write $\Delta_{q,X}(s)=\{\Delta_{q,x}(s):x\in X\}$ for the changes in target activation across the input set.

Let $\operatorname{std}_X(a_q)$ denote the standard deviation of the unperturbed activations $\{a_q(x):x\in X\}$.
To account for differences in activation scale, a common dimensionless relevance parameter $\varepsilon$ is used to define the target-specific margin
\begin{align}
	\varepsilon_q
	&=\varepsilon\max\!\left(
	\operatorname{std}_X(a_q),
	\frac{0.05}{N}\sum_{r=1}^N\operatorname{std}_X(a_r)
	\right).
	\label{eq:target_relevance_margin}
\end{align}
The second term ensures a nonzero relevance margin also for targets whose activation is constant over $X$, provided that other neurons vary over $X$.

Let $\operatorname{med}_{\mathrm{HL}}(\Delta_{q,X}(s))$ denote the Hodges--Lehmann pseudomedian \citep{Hodges1963} of the changes in target activation.
To test whether the effect exceeds $\varepsilon_q$ in either direction, we use a two-sided Wilcoxon signed-rank relevance test \citep{Wilcoxon1945} with hypotheses
\begin{align}
	H_0:\ \abs{\operatorname{med}_{\mathrm{HL}}(\Delta_{q,X}(s))}
	\leq \varepsilon_q,\quad
	H_1:\ \abs{\operatorname{med}_{\mathrm{HL}}(\Delta_{q,X}(s))}
	>\varepsilon_q.
\end{align}
The two directions are tested using one-sided signed-rank tests of $\Delta_{q,X}(s)-\varepsilon_q$ against positive changes and $\Delta_{q,X}(s)+\varepsilon_q$ against negative changes.
We denote the smaller of the two p-values by $p_{P\to n_q}(s)$.
The link is declared relevant at $s$ when $p_{P\to n_q}(s)\leq c$, where the cutoff $c$ is determined from the test level $\alpha$ by the multiple-testing correction in \Cref{subsec:dsc_construction}.

\subsection{Construction of the filtered directed simplicial complex}
\label{subsec:dsc_construction}

We build $\mathcal{K}=\mathcal{K}_X$ by induction on dimension.

\begin{enumerate}
	\item Initialise the complex with the vertex set $V=\{n_i\}_{i\in I}$ and assign every vertex filtration value zero.

	\item At dimension $d\geq1$, take every admitted $(d-1)$-simplex $\rho=(n_{i_0},\ldots,n_{i_{d-1}})$, set $P=\{n_{i_0},\ldots,n_{i_{d-1}}\}$, and extend it by each neuron $n_q\in V\setminus P$ to form the candidate $\sigma=(n_{i_0},\ldots,n_{i_{d-1}},n_q)$.
	Only candidates whose codimension-one faces have all been admitted are tested and counted for the multiple-testing correction.

	\item For each candidate, jointly ablate $P$ and test $P\to n_q$ as in \Cref{subsec:relevance_test}.

	\item Let $m_d$ denote the number of candidates tested at dimension $d$.
	For a test level $\alpha$, let $c_d$ be the cutoff obtained by applying the chosen multiple-testing correction to the family of $m_d$ p-values at full ablation.
	Admit $\sigma$ exactly when $p_{P\to n_q}(1)\leq c_d$.
	For the reported experiments, we use Bonferroni correction, giving $c_d=\alpha/m_d$, with $\alpha=0.01$.
	The choice of correction is investigated in \Cref{subsec:joint_sensitivity}.

	\item For each admitted candidate, evaluate the relevance test at ablation values $s_k=k/K$, $k=0,\ldots,K$.
	Define $s^*(P\to n_q)$ as the smallest $s_j$ for which the link is found to be relevant at every subsequent ablation value, including full ablation:
	\begin{align}
		s^*(P\to n_q)
		&=\min\left\{s_j:
		p_{P\to n_q}(s_\ell)\leq c_d
		\ \text{for every }\ell\geq j\right\}.
	\end{align}
	This corresponds to testing in decreasing order of $s$, stopping at the first non-rejection.

	\item Let $\partial\sigma$ denote the codimension-one faces of an admitted candidate and assign
	\begin{align}
		f_\sigma
		&=\max\left(
		s^*(P\to n_q),
		\max_{\tau\in\partial\sigma}f_\tau
		\right).
		\label{eq:simplex_filtration}
	\end{align}
	This choice prevents a simplex from appearing before any of its proper faces.

	\item Stop if dimension $d$ admits no simplex, since no higher-dimensional simplex can then have all its faces admitted.
	Otherwise continue at dimension $d+1$, unless the chosen maximum dimension has been reached.
\end{enumerate}

By construction, $\mathcal{K}_{\le s}=\{\sigma\in\mathcal{K}:f_\sigma\leq s\}$ defines a filtered directed simplicial complex.
A small value of $s^*$ records an immediate functional dependence, while a larger value records one that becomes detectable only after more of $P$ has been removed.

The influence of the relevance margin, sample size, multiple-testing correction, test level, and spacing of the ablation values on $\mathcal{K}$ is analysed in \Cref{sec:statistical_sensitivity}.

\subsection{Comparing filtered complexes}
\label{subsec:jaccard_distance}

For filtered DSCs $\mathcal{K}$ and $\mathcal{L}$ on the same vertex set $V$, we compare their sublevel sets without vertices by integrating the ordinary Jaccard distance $\delta_J$ over the filtration,
\begin{align}
	d_J(\mathcal{K},\mathcal{L})
	=\int_0^1\delta_J\!\left(\mathcal{K}_{\le s}\setminus V,\,\mathcal{L}_{\le s}\setminus V\right)\,\mathrm{d}s,
	\label{eq:jaccard_distance}
\end{align}
where $\delta_J(A,B)=1-\abs{A\cap B}/\abs{A\cup B}$, with $\delta_J(\varnothing,\varnothing)=0$.
Vertices are excluded because both complexes contain all of $V$ for all filtration values, so including them would add the same $\abs{V}$ elements to every intersection and union and pull all distances towards zero.
Integrating over the filtration distinguishes complexes with the same simplices but different filtration values.

\clearpage

\section{Neural network specifications and input sets}
\label{sec:quadrant_setup}

This section describes the artificial neural networks used in the experiments of this article and the input sets $X$ from which we construct associated filtered directed simplicial complexes (DSCs) $\mathcal{K}_X$ as described in \Cref{sec:construction}.

\subsection{Network specifications}
\label{subsec:quadrant_networks}

All networks are multilayer perceptrons (MLPs), with rectified linear unit (ReLU) activations in the hidden layers, that classify points of the input domain $[-3,3]^2$ into four classes.
The four classes are the quadrants $Q_1$ ($x\geq0,y\geq0$), $Q_2$ ($x<0,y\geq0$), $Q_3$ ($x<0,y<0$) and $Q_4$ ($x\geq0,y<0$).
We use two architectures, $[2,4,4,4]$ and $[2,3,4,4]$, and for each study three kinds of network, differing in how their weights and biases are obtained.
On the one hand, we consider handcrafted networks, for which the internal computations are known exactly and can be compared to the constructed DSCs.
On the other hand, we consider trained networks that have been optimised to perform the classification task and randomly initialised networks to investigate the structures of DSCs obtained in more realistic settings.
We report the performance of the networks for a fixed evaluation set of $1{,}000$ points drawn uniformly from the input domain.

In the handcrafted $[2,4,4,4]$ network, each of the four first-layer neurons detects exactly one of the elementary features $x>0$, $x<0$, $y>0$ and $y<0$, while each of the three first-layer neurons of the handcrafted $[2,3,4,4]$ network responds to a combination of both coordinates.
We therefore call the first the \emph{monosemantic} network and the second the \emph{polysemantic} network.
Note, however, that whether a single neuron counts as monosemantic depends on the choice of elementary features, and in an input basis rotated by $45^\circ$ the readings of the two networks would be exchanged.

\begin{figure}[p]
	\centering
	\textbf{Handcrafted classifiers}\par
	\includegraphics[width=.5\textwidth, height=.18\textheight, keepaspectratio]{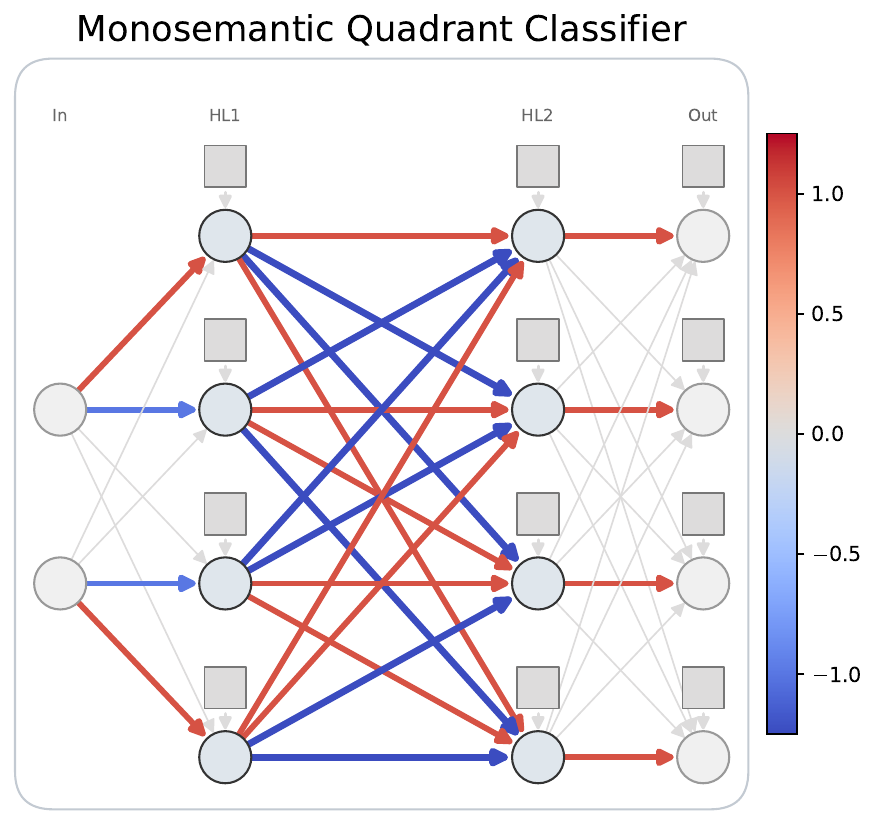}
	\includegraphics[width=.4\textwidth, height=.18\textheight, keepaspectratio]{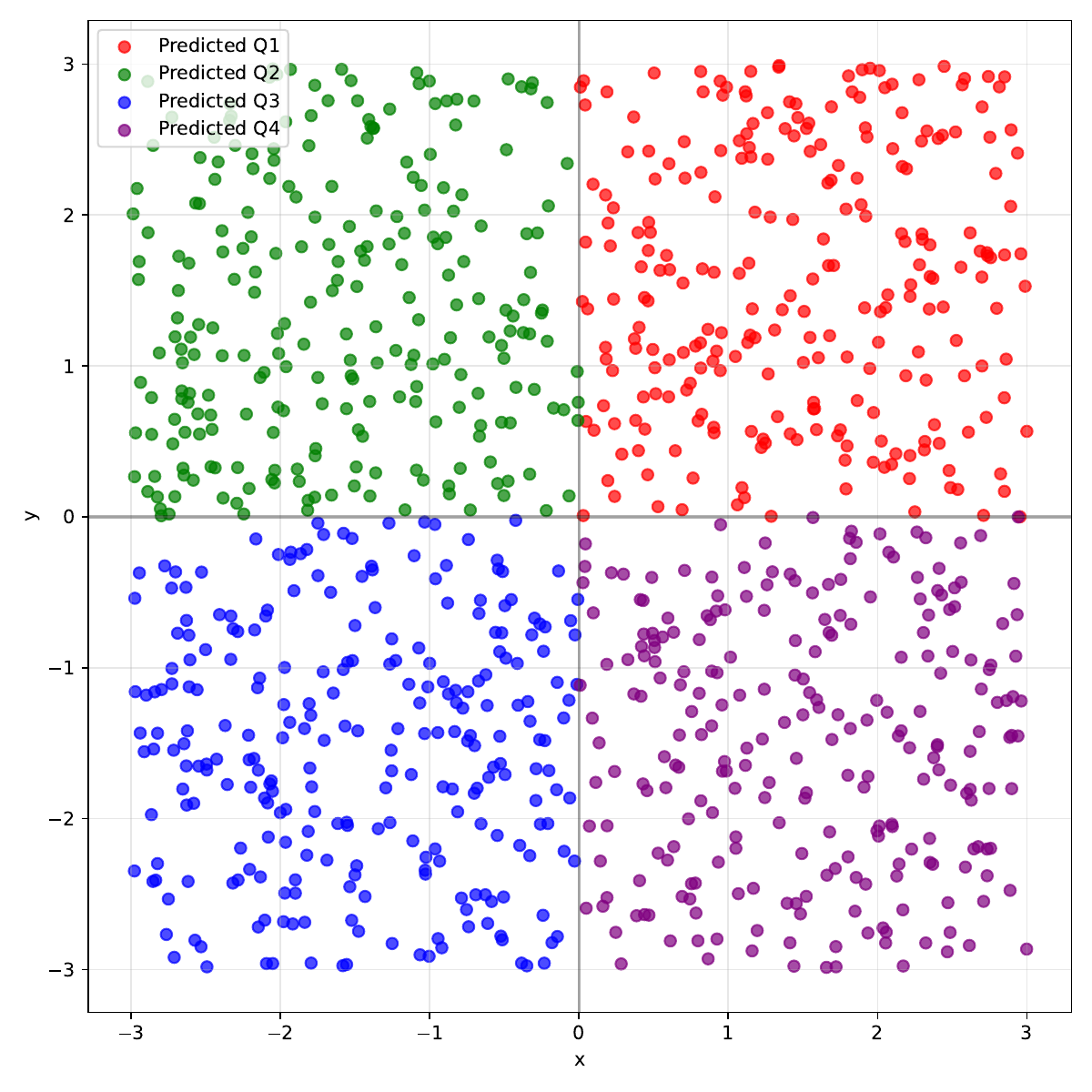}

	\vspace{.1cm}
	\includegraphics[width=.5\textwidth, height=.18\textheight, keepaspectratio]{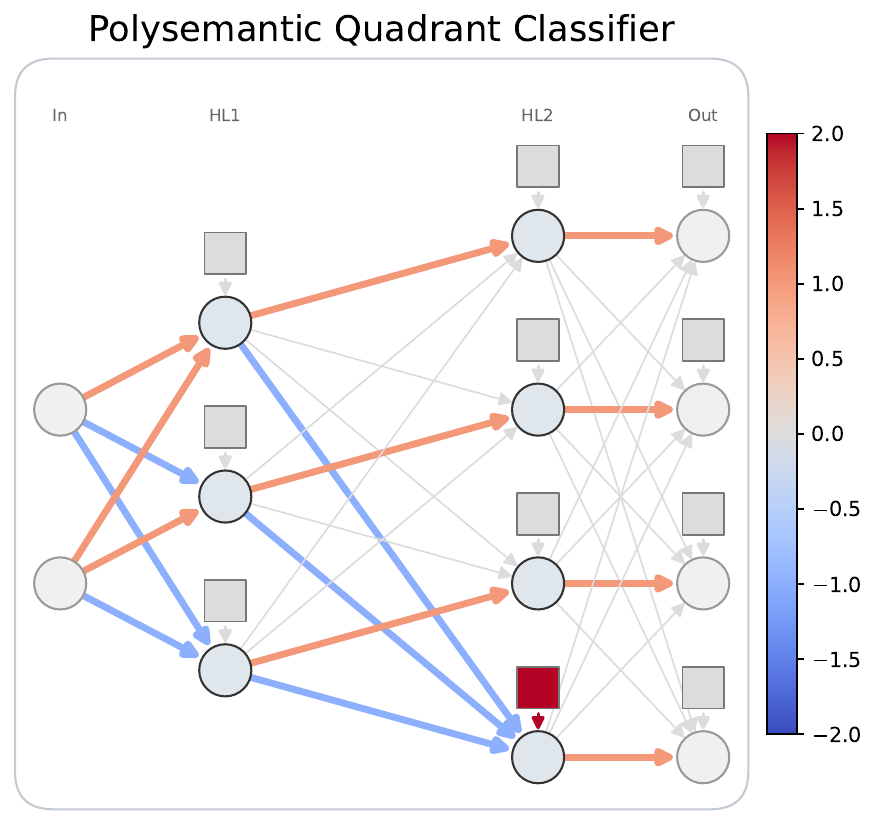}
	\includegraphics[width=.4\textwidth, height=.18\textheight, keepaspectratio]{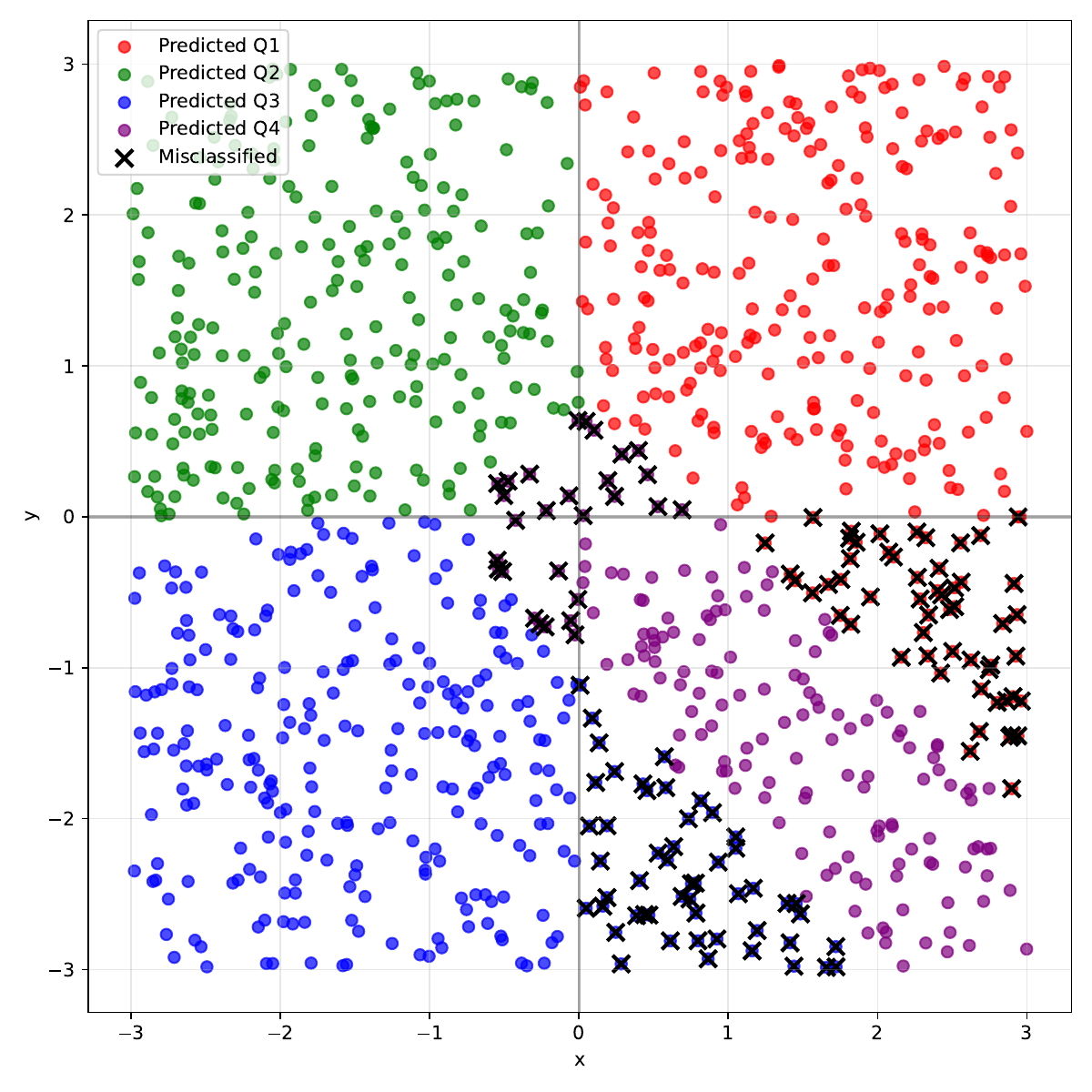}

	\vspace{.2cm}
	\textbf{Trained classifiers}\par
	\includegraphics[width=.5\textwidth, height=.18\textheight, keepaspectratio]{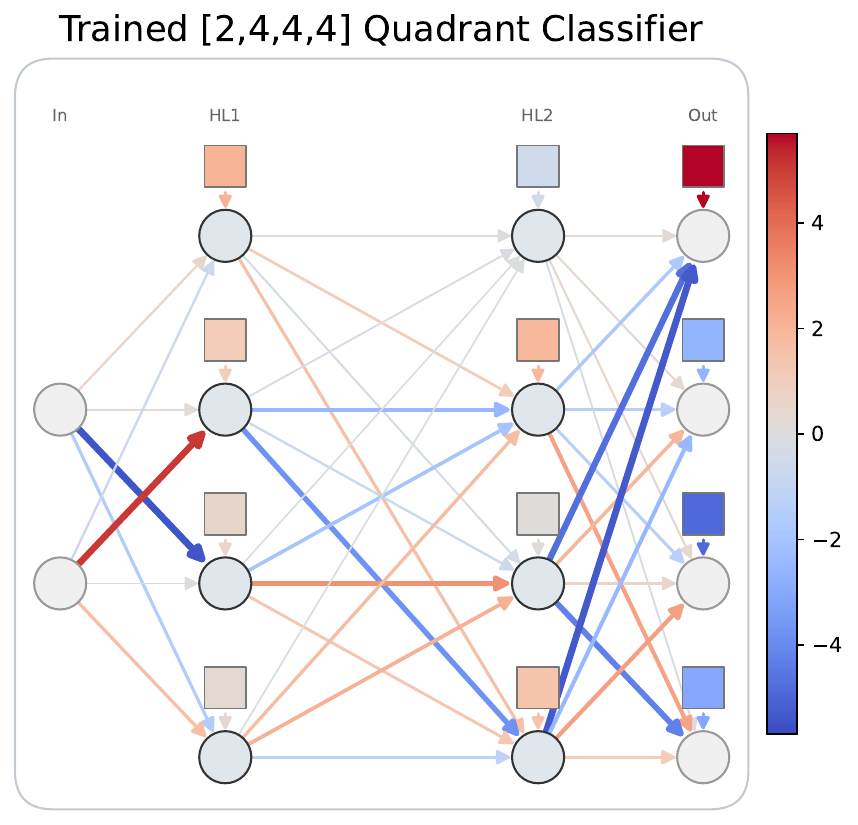}
	\includegraphics[width=.4\textwidth, height=.18\textheight, keepaspectratio]{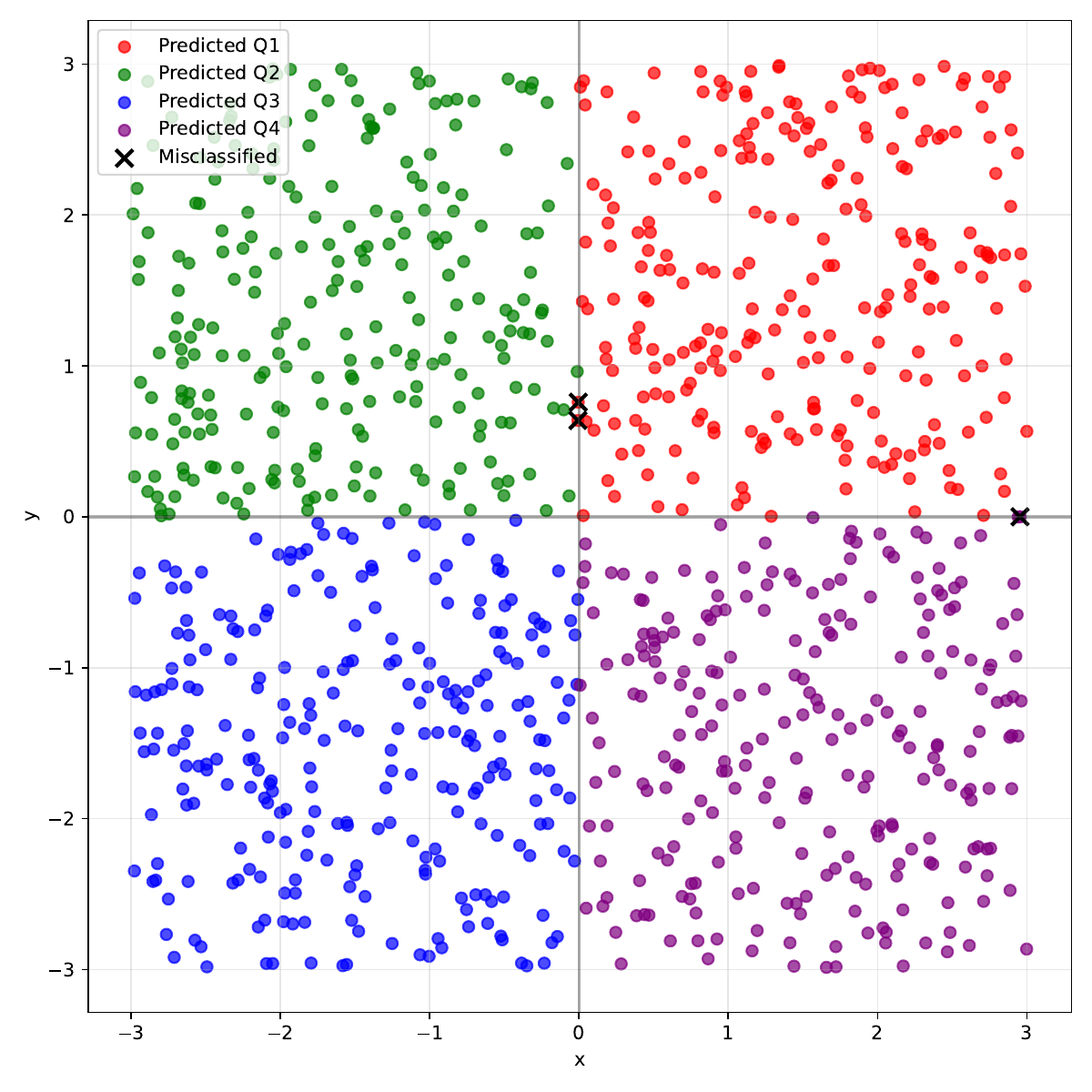}

	\vspace{.1cm}
	\includegraphics[width=.5\textwidth, height=.18\textheight, keepaspectratio]{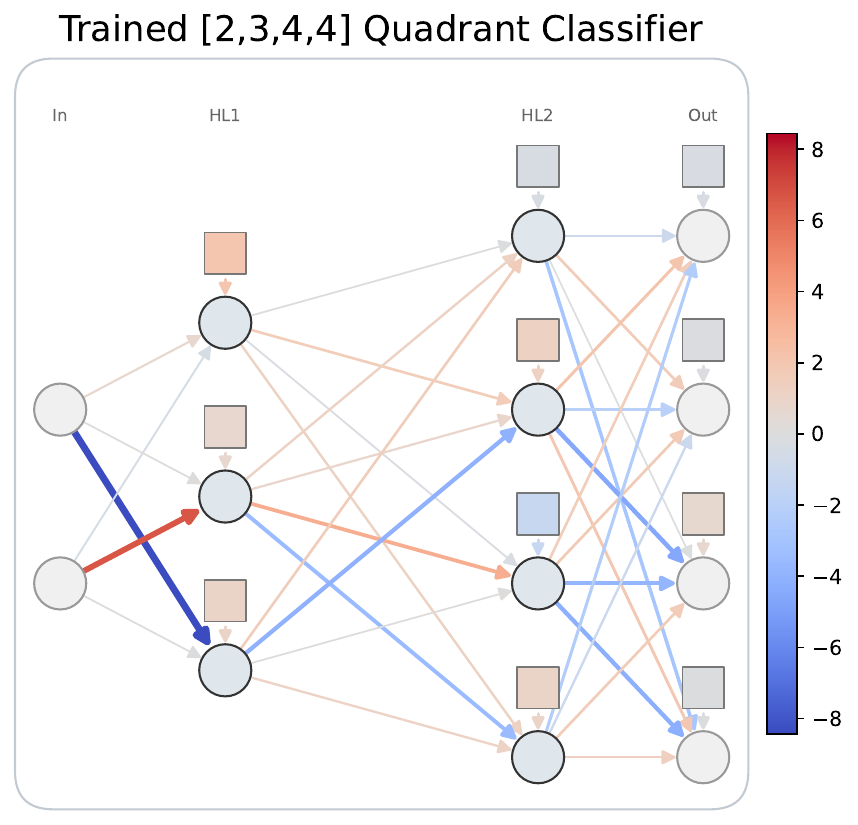}
	\includegraphics[width=.4\textwidth, height=.18\textheight, keepaspectratio]{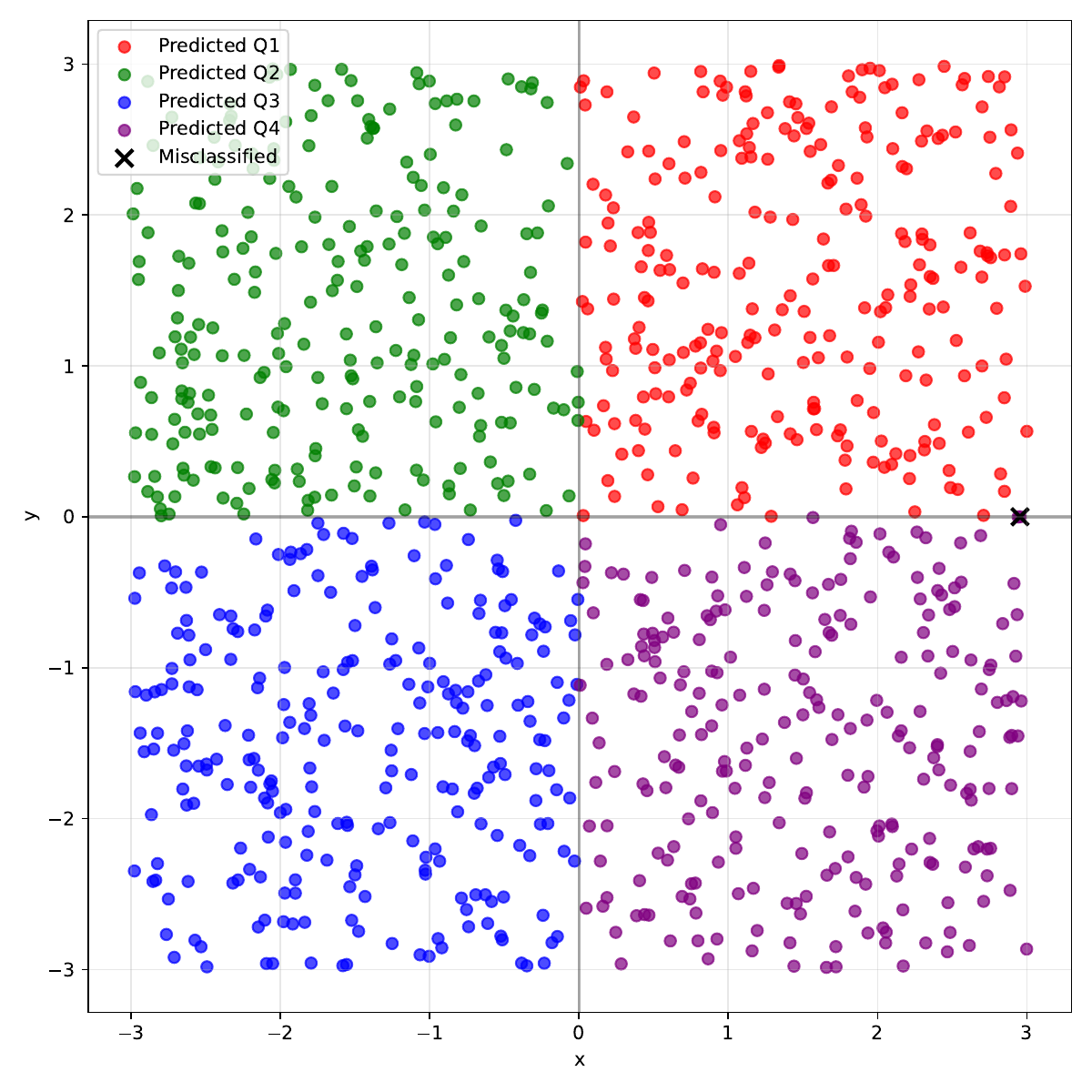}
	\caption{Architectures and decision regions of the quadrant classifiers.
	Each row shows the parameterised network (left) and its predicted classes on the evaluation set (right).
	In the network diagrams, circles are neurons, squares are biases and arrows are weights.
	The colour of a square or arrow shows its value on the colour bar of the panel, and the width of an arrow grows with the magnitude of the weight.
	From top to bottom: handcrafted monosemantic $[2,4,4,4]$, handcrafted polysemantic $[2,3,4,4]$, trained $[2,4,4,4]$ and trained $[2,3,4,4]$ networks.}
	\label{fig:appendix_quadrant_networks_scatter}
\end{figure}

\paragraph{Trained networks.}
We trained networks on both architectures with Adam \citep{Kingma2014} at learning rate $0.005$, batch size $256$, for $200$ epochs of $10{,}000$ draws each ($2\times10^6$ draws in total).
For the loss we used the cross-entropy between the softmax of the outputs and the quadrant label.
Each minibatch is an independent uniform draw from $[-3,3]^2$.
Their accuracy on the evaluation set is $0.997$ for the $[2,4,4,4]$ network and $0.999$ for the $[2,3,4,4]$ network.
The bottom two rows of \Cref{fig:appendix_quadrant_networks_scatter} show the learned parameters and decision regions.
In the trained $[2,4,4,4]$ network, two first-layer neurons depend mainly on one input coordinate and two mix both coordinates.
The trained $[2,3,4,4]$ network has two first-layer neurons of the first kind and one of the second.
We therefore refer to the trained networks by their architecture only.

\paragraph{Monosemantic network.}
The monosemantic classifier is a $[2,4,4,4]$ MLP with two four-unit hidden layers and four outputs.
All biases are zero.
The first-layer weights are
\begin{align}
	W^{(1)}
	&= \begin{pmatrix}
		1.0 & 0.0 \\
		-1.0 & 0.0 \\
		0.0 & -1.0 \\
		0.0 & 1.0
	\end{pmatrix}.
\end{align}
After the ReLU nonlinearity, neurons $n_1$--$n_4$ encode $x>0$, $x<0$, $y<0$, and $y>0$, respectively.
The second-layer weights are
\begin{align}
	W^{(2)}
	&= \begin{pmatrix}
		1.0 & -1.25 & -1.25 & 1.0 \\ 
		-1.25 & 1.0 & -1.25 & 1.0 \\ 
		-1.25 & 1.0 & 1.0 & -1.25 \\ 
		1.0 & -1.25 & 1.0 & -1.25  
	\end{pmatrix}.
\end{align}
This layer implements piecewise-linear conjunctions of the first-layer features.
For example, neuron $n_5$ is active in $Q_1$ when neurons $n_1$ and $n_4$ are active, while negative weights suppress contributions from incompatible sign features.
The output weights are $W^{(3)}=\operatorname{Id}_4$.
The first row of \Cref{fig:appendix_quadrant_networks_scatter} shows this network and its decision regions.
Its accuracy on the evaluation set is $1.0$.

\paragraph{Polysemantic network.}
The polysemantic classifier is a $[2,3,4,4]$ MLP.
Its three-unit first layer forms a bottleneck that induces feature sharing.
Its weights are
\begin{align}
	W^{(1)}
	&= \begin{pmatrix}
		1.0 & 1.0 \\  
		-1.0 & 1.0 \\ 
		-1.0 & -1.0  
	\end{pmatrix}.
\end{align}
The first-layer biases are zero.
Under ReLU, these directions map the corners of a square centred at the origin into three positive codes and one all-zero code.
Consequently, several input regions share units, while the fourth is represented by their coordinated inactivity.

The second-layer parameters are
\begin{align}
	W^{(2)}
	&= \begin{pmatrix}
		1.0 & 0.0 & 0.0 \\   
		0.0 & 1.0 & 0.0 \\   
		0.0 & 0.0 & 1.0 \\   
		-1.0 & -1.0 & -1.0  
	\end{pmatrix}, \\
	b^{(2)}
	&= \begin{pmatrix} 0.0 \\ 0.0 \\ 0.0 \\ 2.0 \end{pmatrix}.
\end{align}
Here neurons $n_4$, $n_5$, and $n_6$ copy the three first-layer codes, while the biased neuron $n_7$ represents $Q_4$ by activating when their total activity is small.
The output layer has weights $W^{(3)}=\operatorname{Id}_4$ and zero bias.
The second row of \Cref{fig:appendix_quadrant_networks_scatter} shows this network and its decision regions.
Its accuracy on the evaluation set is $0.866$, and all errors involve $Q_4$.
Within $Q_4$, the network assigns only the points with $\abs{x+y}<1$ to $Q_4$ and the others to $Q_1$ or $Q_3$.
Conversely, some points of the other quadrants close to the origin are assigned to $Q_4$.

\paragraph{Untrained networks.}
As a control, we use randomly initialised networks of both architectures, with the default PyTorch initialisation and seeds $0$, $1$ and $2$.
Their accuracy on the evaluation set lies between $0.243$ and $0.258$, close to the level of pure chance of $0.25$.
In fact, five of the six networks assign all points to the same quadrant, and the sixth assigns a region in the upper right corner of the input domain to $Q_2$ and all other points to $Q_3$.
Each network thus classifies the points of exactly one quadrant correctly, and its accuracy is the share of this quadrant in the evaluation set.

\subsection{Input sets}
\label{subsec:probe_contexts}

A DSC $\mathcal{K}_X$ describes the network in the context of a fixed set of inputs $X$, because the relevance test relies on these inputs.
We can only conclude that the network has detected a particular shared feature in the input set $X$ if it produces similar outputs for all of these inputs.
For a target vector $h\in\mathbb{R}^4$, we thus construct a set of inputs $X$ by selecting points $x$ at which the output $F(x)$ of the network is close to the desired output $h$.
We quantify closeness by the cosine similarity $\cos(F(x),h)$.

The outputs of the handcrafted networks are non-negative by construction, and we can use them unchanged, $F(x)=a^{(L)}(x)$.
The outputs of the trained networks are signed logits, and adding a constant to all of them would change their cosine with $h$ but not the classification, so for these networks we additionally apply a softmax $F(x)=\operatorname{softmax}(a^{(L)}(x))$ in line with their training loss.
For a given selection tolerance $\eta$, an input $x$ is accepted if $\cos(F(x),h)\geq1-\eta$.
Thus $\eta=0$ accepts only inputs at which $F(x)$ points in the direction of $h$, while a larger $\eta$ accepts inputs from a larger region of input space.
We draw inputs uniformly from the input domain and keep the accepted ones until we have collected $B$ of them.

For a quadrant $Q_i$, the \emph{one-hot} inputs $X_{Q_i}$ are selected with the one-hot target vector $h$ that has entry $1$ in position $i$ and $0$ elsewhere, so that the network assigns them unambiguously to $Q_i$.
For two adjacent quadrants $Q_i$ and $Q_j$, the \emph{two-hot} inputs $X_{Q_iQ_j}$ are selected with the target vector $h$ that has entries $1$ in positions $i$ and $j$ and $0$ elsewhere.
This gives the four sets $X_{Q_1Q_2}$, $X_{Q_1Q_4}$, $X_{Q_2Q_3}$ and $X_{Q_3Q_4}$.
These inputs lie close to the decision boundary between $Q_i$ and $Q_j$, where the network is undecided between the two quadrants.
In the monosemantic network, for example, the two-hot inputs $X_{Q_1Q_2}$ have $y>0$, the feature shared by both quadrants, while $x$ is close to zero and might be either positive or negative.
We do not consider two-hot inputs for the diagonally opposite pairs $Q_1,Q_3$ and $Q_2,Q_4$, because these quadrants meet only at the origin, so that the networks typically have no inputs at which their output is close to the corresponding two-hot target vector.
The \emph{random} inputs $X_{\mathrm{rand}}$ are 128 points sampled uniformly from the evaluation set without any selection.
Such a set mixes inputs from all four quadrants and therefore does not single out any of them.
For the untrained networks, we use the one-hot inputs of the trained network with the same architecture (\Cref{subsec:quadrant_untrained}).

\clearpage

\section{Sensitivity to the relevance test}
\label{sec:statistical_sensitivity}

We analyse how the choices of relevance margin $\varepsilon$, sample size $B$, multiple-testing correction, test level $\alpha$, and the spacing $1/K$ of the ablation values in the construction (\Cref{subsec:relevance_test} and \Cref{subsec:dsc_construction}) influence the resulting complexes, and use this analysis to choose the configuration adopted in the rest of the article.

\subsection{Minimum sample size and test power}
\label{subsec:minimum_sample_size}

The sample size $B$ determines the smallest p-value the relevance test can return and the power of the test.

First, with $B$ inputs, the one-sided signed-rank test cannot return a p-value below $2^{-B}$, the probability under the null hypothesis that all $B$ changes in target activation have the same sign.
A candidate at dimension $d$ is admitted only if $p_{P\to n_q}(1)\leq c_d$ (\Cref{subsec:dsc_construction}), so dimension $d$ stays empty unless $B\geq\log_2(1/c_d)$.
Without multiple-testing correction, we have $c_d=\alpha$, which at $\alpha=0.01$ requires a sample size of $B\geq7$.
Benjamini--Hochberg can raise its cutoff up to $\alpha$, so its bound is the same.
The cutoff of Benjamini--Yekutieli is at most $\alpha/H_{m_d}$, where $H_m=\sum_{i=1}^m 1/i$, which requires $B\geq\log_2(H_{m_d}/\alpha)$.
Bonferroni has the fixed cutoff $c_d=\alpha/m_d$.
This is the smallest of the four cutoffs, and it requires
\begin{align}
	B \geq \log_2\!\frac{m_d}{\alpha}.
	\label{eq:minimum_sample_size}
\end{align}
Under the Bonferroni correction used in the main experiments (\Cref{subsec:joint_sensitivity}), every ordered pair of distinct neurons is a candidate at dimension one, so $m_1=N(N-1)$ and a complex has edges only if $B\geq\log_2(N(N-1)/\alpha)$.
At higher dimensions, the number of candidates is at most $m_d\le N!$, the number of orderings of all neurons, so a sample size of $B\geq\log_2(N!/\alpha)$ covers every dimension.
For the quadrant classifiers at $\alpha=0.01$, edges require $B\geq13$, and the bound holds at every dimension for $B\geq22$ ($N=8$) or $B\geq19$ ($N=7$).
If $B$ is below the bound at some dimension, that dimension is empty because of the sample size.

Second, recall that the power of a statistical test is the probability of rejecting the null hypothesis when the alternative is true.
For the signed-rank test, the power depends on the distribution of the changes in target activation.
Changes in target activation can be zero-inflated, that is, a large fraction of them are exactly zero.
Suppose a fraction $\rho$ of the changes in target activation is much larger than $\varepsilon_q$ and the other changes are zero.
The normal approximation of the signed-rank statistic then gives
\begin{align}
	z
	&\approx\frac{\sqrt{3B}}{2}\bigl(2\rho(2-\rho)-1\bigr).
\end{align}
The test detects the effect only if $\rho>1-1/\sqrt{2}\approx0.29$, which is where the pseudomedian of the changes exceeds $\varepsilon_q$.
Near this value, the sample size needed to reach a given cutoff grows like $(\rho-1+1/\sqrt{2})^{-2}$.
For example, at $\rho=0.5$, the dimension-one Bonferroni cutoff of the quadrant classifiers requires about $70$ inputs, as opposed to the minimum of $13$, to detect the effect with $50\%$ power.

\subsection{Parameter sweep}
\label{subsec:joint_sensitivity}

We performed a parameter sweep to assess the sensitivity of the complexes to the parameters of the relevance test.
We varied the sample size $B$, the relevance margin $\varepsilon$, the test level $\alpha$, and the spacing $1/K$ of the ablation values $s_k=k/K$ at which the relevance test is evaluated (\Cref{subsec:dsc_construction}) over
\begin{align}
	B &\in \{8,16,32,64,128,256,512\},\\
	\varepsilon &\in \{0,0.001,0.0025,0.005,0.01,0.025, 0.05,0.1,0.2,0.4\},\\
	\alpha &\in \{0.001,0.01,0.05\},\\
	K &\in \{250,500,1000\},
\end{align}
and either left the p-values uncorrected or applied the multiple-testing correction of Bonferroni \citep{Dunn1961}, Benjamini--Hochberg \citep{Benjamini1995} or Benjamini--Yekutieli \citep{Benjamini2001}.

The experiments used the handcrafted monosemantic and polysemantic quadrant classifiers (\Cref{subsec:quadrant_networks}).
For each classifier and each quadrant $Q_i$, we used the one-hot inputs $X_{Q_i}$ (\Cref{subsec:probe_contexts}).
For each classifier, each quadrant, and each parameter combination, we built complete filtered directed simplicial complexes (DSCs) up to the maximum possible simplex dimension with fixed integration step $\Delta t=0.01$ of the trajectories (\Cref{sec:sensitivity}) and fixed selection tolerance $\eta=0.01$ for the inputs (\Cref{subsec:probe_contexts}).
The p-values depend only on the classifier, the quadrant, $B$, and $\varepsilon$, because the ablation values for $K=250$ and $K=500$ are a subset of those for $K=1000$.
We therefore computed them once at the ablation values for $K=1000$, for each of the $2\times4\times7\times10=560$ combinations of classifier, quadrant, $B$, and $\varepsilon$.
From these p-values, we built the DSCs for all $3\times4\times3=36$ combinations of $\alpha$, multiple-testing correction, and $K$, which gives $560\times36=20{,}160$ DSCs.

In each of the following paragraphs, we report the effect of varying a single parameter.
Unless stated otherwise, the other parameters are held at $\varepsilon=0.01$, $\alpha=0.01$, Bonferroni correction, $K=1000$ and $B=512$.

\begin{figure}[t]
	\centering
	\includegraphics[width=\linewidth]{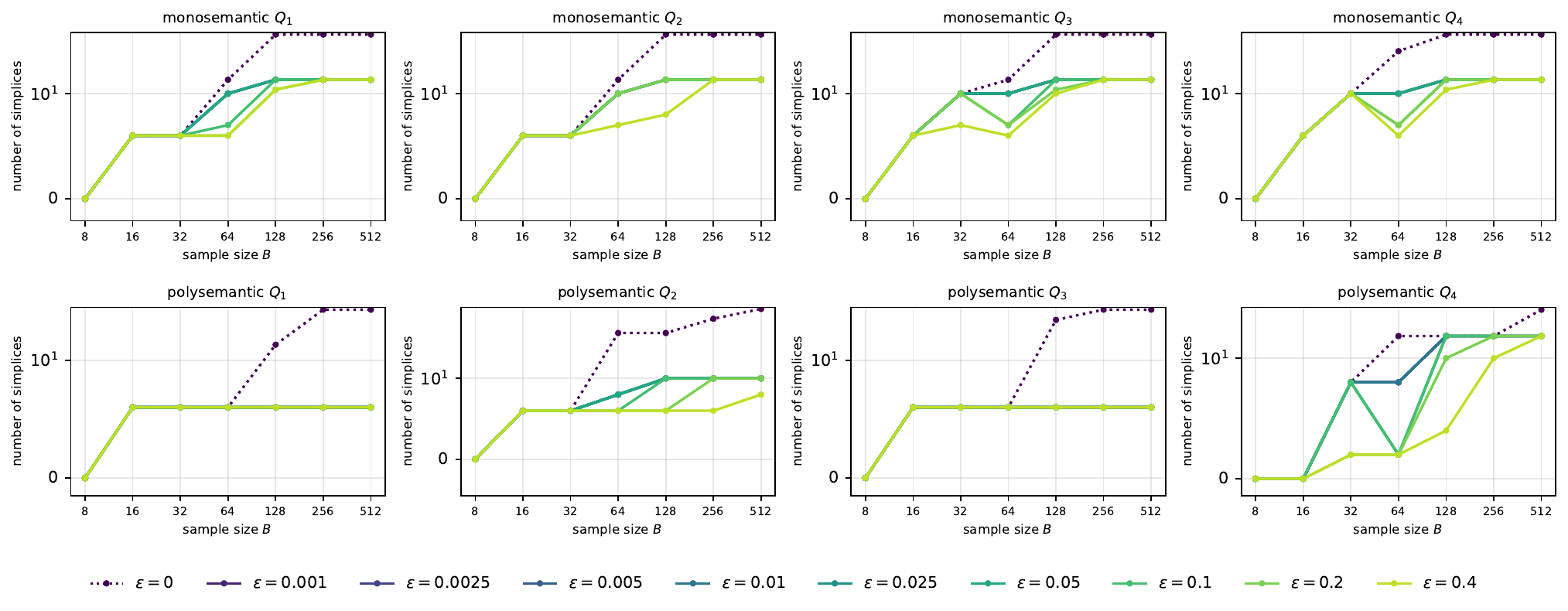}
	\caption{
		Number of simplices, excluding vertices, as a function of the sample size $B$ for both classifiers and the one-hot inputs of all four quadrants, at $\alpha=0.01$ with Bonferroni correction and $K=1000$.
		Colours give the relevance margin $\varepsilon$, while the dotted curve is $\varepsilon=0$.
		At $\varepsilon=0.01$, the simplex sets at $B=128$ and $B=512$ are identical in all eight complexes, whereas $B=64$ gives the $B=512$ simplex set in only two of them.
	}
	\label{fig:inference_probe_convergence}
\end{figure}

\paragraph{Relevance margin.}
At $B=512$, every relevance margin $\varepsilon$ from $0.001$ through $0.2$ produced exactly the same simplex set for every set of inputs, for every $\alpha$ and every multiple-testing correction.
In contrast, at a vanishing relevance margin $\varepsilon=0$, where the relevance test asks whether the target activation changes at all, the complexes were larger for every input set.
They contained between 28 and 90 simplices besides the vertices, compared with 6 to 16 for $\varepsilon>0$.
In particular, three of the polysemantic complexes at $\varepsilon=0$ contained 3-simplices, while no complex at $\varepsilon>0$ contained simplices above dimension two.
For $X_{Q_2}$ and $X_{Q_4}$ of the polysemantic network, the $\varepsilon=0$ complexes were still growing between $B=256$ and $B=512$ (\Cref{fig:inference_probe_convergence}).

This difference is related to the power argument in \Cref{subsec:minimum_sample_size}.
With $\varepsilon>0$, target activations that do not change along the trajectory ($\Delta_{q,x}(s)=0$) count as evidence against relevance.
At $\varepsilon=0$, these cases are dropped from the signed-rank test, so the test reports as relevant as soon as enough of the changes in target activation are nonzero and share a sign.
The $\varepsilon=0$ complexes therefore tend to keep growing as the sample size $B$ increases, since there are more opportunities for nonzero changes in target activation.
A nonzero relevance margin sets a minimum effect size and thus removes links with too few nonzero changes in target activation, which a large enough sample would otherwise always detect.

While the simplex sets of the complexes were invariant to $\varepsilon$ in the range $[0.001,0.2]$, the filtration values of the simplices did change.
Compared with $\varepsilon=0.01$, relevance margins from $0.001$ to $0.025$ changed filtration values by at most $0.051$, and no two simplices changed their order in the filtration.
At $\varepsilon=0.2$, filtration values changed by up to $0.64$, and simplices changed their order in five complexes.
A larger relevance margin delays the filtration values, because a link needs a larger effect to be declared relevant.
We therefore use $\varepsilon=0.01$, which lies well inside the range of relevance margins that give the same simplex set.

\paragraph{Sample size.}

As $B$ increased, the number of simplices in the complexes increased (\Cref{fig:inference_probe_convergence}).
At $B=8$, the complexes contained only vertices, as expected from the minimum sample size of $13$ for edges (\Cref{subsec:minimum_sample_size}).
Every simplex admitted at a smaller $B$ was also admitted at $B=512$, in line with the increasing power of the relevance test.
From $B=128$ on, all eight complexes had the simplex set of $B=512$.

The Jaccard distance $d_J$ (\Cref{subsec:jaccard_distance}) to the $B=512$ complex shrank as $B$ increased.
It ranged from $0.047$ to $0.66$ at $B=64$, from $0.009$ to $0.19$ at $B=128$, and from $0.001$ to $0.14$ at $B=256$.
From $B=128$ on, this distance came from the filtration values alone, because the simplex sets agreed.
A larger $B$ increases the power of the relevance test and therefore tends to move $s^*$ earlier (\Cref{subsec:minimum_sample_size}).
In addition, the input sets for different $B$ were drawn independently, so the filtration values also varied with the sampled inputs.

Since the simplex sets were fixed from $B=128$ on and the remaining variation in the filtration values is expected, we use $B=128$ in all other experiments of this article (\Cref{sec:sensitivity,sec:quadrant_classifiers_details}).

\paragraph{Multiple-testing correction.}

At each dimension, all candidate simplices are tested at once, and the multiple-testing correction turns the test level $\alpha$ into the cutoff $c_d$ for their p-values (\Cref{subsec:dsc_construction}).
Bonferroni correction controls the family-wise error rate \citep{Dunn1961}, so that with probability at least $1-\alpha$ none of the admitted candidates of a dimension has a link that is not relevant.
Benjamini--Hochberg and Benjamini--Yekutieli control the false discovery rate \citep{Benjamini1995}, so that on average at most a fraction $\alpha$ of the admitted candidates have a link that is not relevant.
Two candidates $P\to n_q$ and $P\to n_{q'}$ with the same predictor set $P$ are tested on the same coherent counterfactual ablation trajectories $\gamma_{a(x),P}$, $x\in X$.
In particular, the associated changes in target activation $\Delta_{q,x}(s)$ and $\Delta_{q',x}(s)$ are computed from the same points $\gamma_{a(x),P}(s)$ (\Cref{subsec:relevance_test}), so their p-values are dependent, and the form of this dependence is unknown.
Bonferroni correction and Benjamini--Yekutieli remain valid under any dependence, whereas Benjamini--Hochberg requires positive dependence \citep{Benjamini2001}.

\begin{figure}[htbp]
	\centering
	\includegraphics[width=\linewidth]{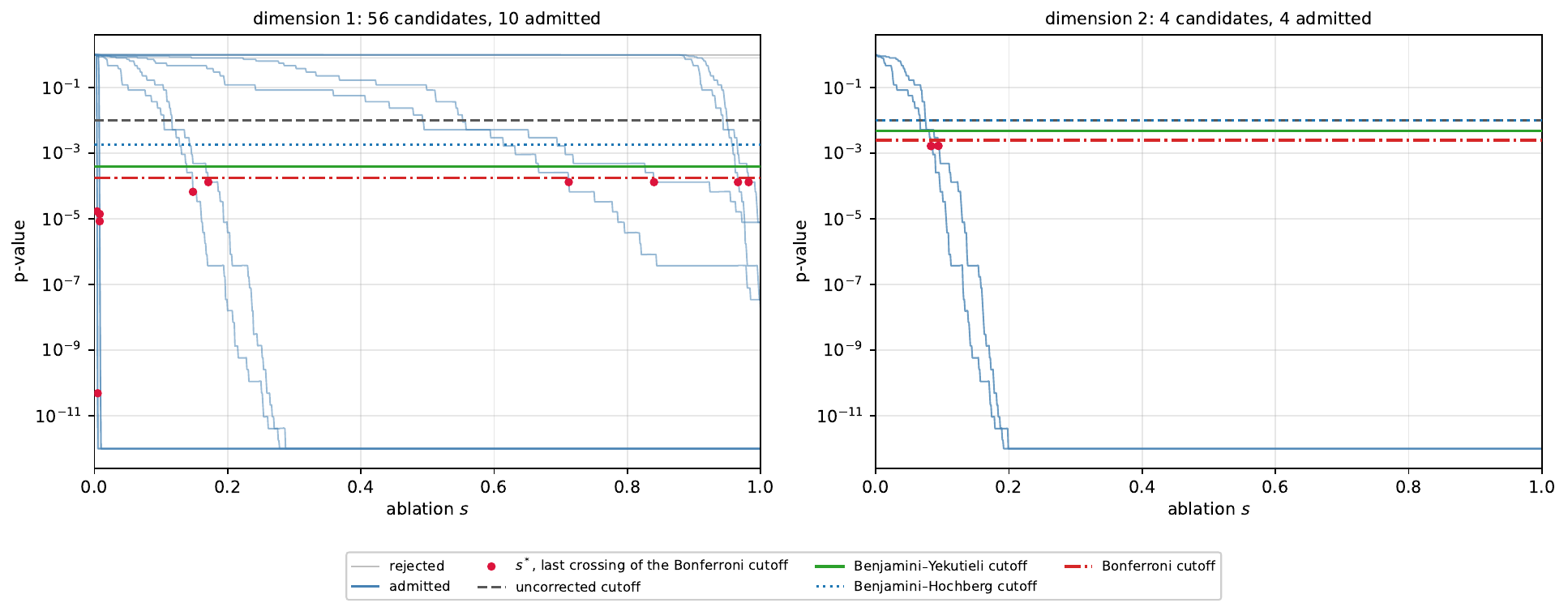}

	\vspace{0.6em}
	\includegraphics[width=\linewidth]{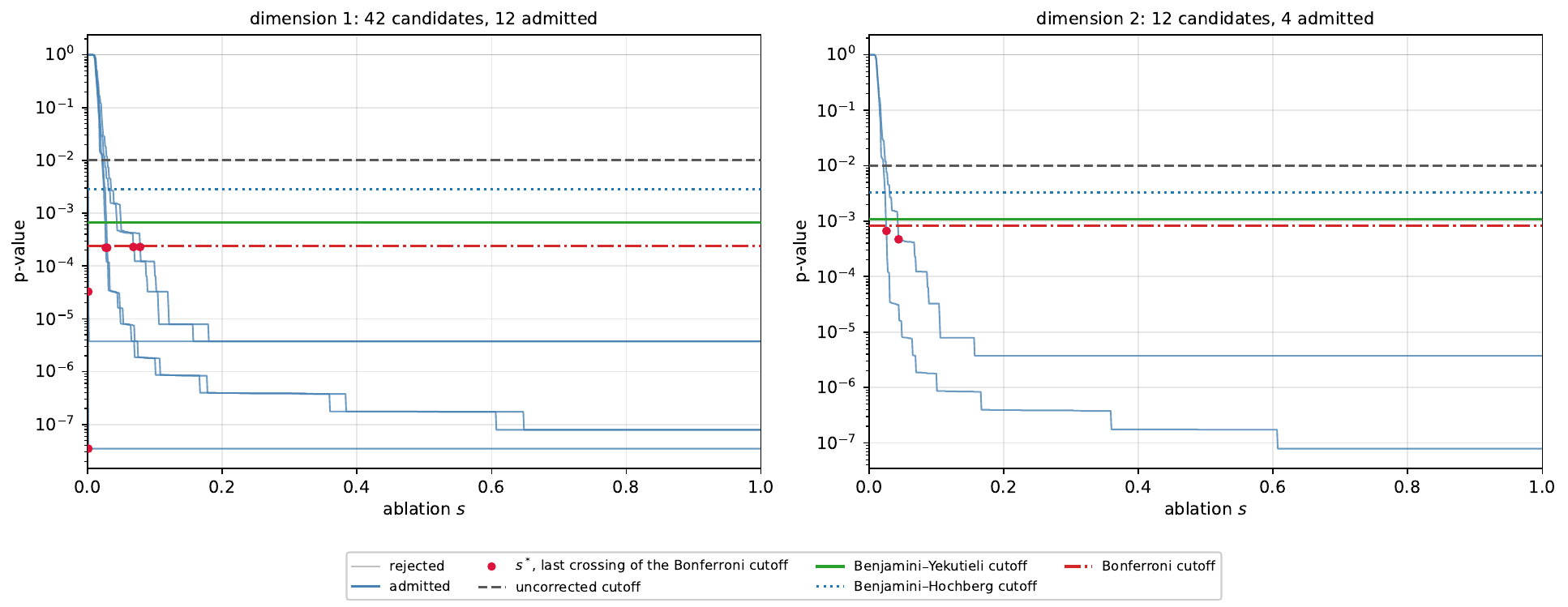}
	\caption{
		P-values along the ablation axis for the one-hot inputs $X_{Q_1}$ of the monosemantic network (top) and $X_{Q_4}$ of the polysemantic network (bottom), at the chosen parameters $B=128$, $\varepsilon=0.01$, $\alpha=0.01$ and $K=1000$.
		The left and right panels show the candidates of dimensions one and two.
		Each curve is the p-value $p_{P\to n_q}(s)$ of one candidate as a function of the ablation $s$.
		Blue curves belong to admitted candidates and grey curves to rejected ones.
		The horizontal lines are the cutoffs $c_d$ under the four multiple-testing corrections.
		At full ablation $s=1$, the p-values of admitted and rejected candidates are separated by several orders of magnitude, and all four cutoffs lie in the gap.
		Red dots mark $s^*$, the last crossing of the Bonferroni cutoff, before the maximum over faces in \eqref{eq:simplex_filtration} is taken.
	}
	\label{fig:appendix_pvalue_trajectories}
\end{figure}

All four options, including uncorrected p-values, returned the same simplex set in all eight complexes.
Changing the multiple-testing correction shifted filtration values by at most $0.14$ and did not change the order of any two simplices in the filtration.
The largest Jaccard distance $d_J$ between two multiple-testing corrections ranged from $0.002$ to $0.073$ across the eight complexes.
The multiple-testing corrections agree because the p-values of admitted and rejected candidates are separated by several orders of magnitude, so all four cutoffs fall into the same gap (\Cref{fig:appendix_pvalue_trajectories}).

Since the choice of multiple-testing correction did not change the simplex sets, we choose the strictest one and use Bonferroni correction in all other experiments of this article (\Cref{sec:sensitivity,sec:quadrant_classifiers_details}).

\paragraph{Test level.}

At $B=512$, all three test levels $\alpha\in\{0.001,0.01,0.05\}$ returned the same simplex set in all eight complexes.
Changing $\alpha$ shifted filtration values by at most $0.14$.
At $B=128$, the three levels still agreed in seven complexes, but for $X_{Q_3}$ of the monosemantic network, $\alpha=0.001$ admitted four fewer simplices than the other two levels.
A smaller $\alpha$ lowers the cutoff $c_d$, so admitting a link with a weak effect requires a larger $B$ (\Cref{subsec:minimum_sample_size}).

We therefore choose $\alpha=0.01$, the strictest test level in the sweep that is consistent with choosing $B=128$, in the remaining experiments in this article (\Cref{sec:sensitivity,sec:quadrant_classifiers_details}).

\paragraph{Spacing of the ablation values.}

A candidate is admitted based on its p-value at full ablation $s=1$, so the spacing $1/K$ cannot change the simplex sets (\Cref{subsec:dsc_construction}).
It can, however, change the filtration values.
If the p-value trajectory is monotone, a finer spacing can only decrease the filtration value by less than the coarser spacing.
If the p-value trajectory crosses back over the cutoff between two ablation values, a finer spacing may detect this and increase the filtration value.

Since it is not known in advance whether the p-value trajectories are monotone, we compared three spacings.
For the parameters chosen above ($B=128$, $\varepsilon=0.01$, $\alpha=0.01$ and Bonferroni correction), no admitted p-value trajectory crossed back over the cutoff (\Cref{fig:appendix_pvalue_trajectories}).
Accordingly, refining the spacing from $1/250$ to $1/1000$ only decreased filtration values, by at most $0.003<0.004 = 1/250$.
In 11 of the 6,720 complexes, a finer spacing revealed a trajectory that crossed back over the cutoff and increased a filtration value by up to $0.026$.
All of them used one of the other two test levels, $\alpha=0.05$ or $\alpha=0.001$.
Their largest increase of $0.026$ is small compared with the changes of up to $0.14$ caused by the choice of test level or multiple-testing correction.

We therefore use the finest tested spacing $K=1000$ in all other experiments of this article (\Cref{sec:sensitivity,sec:quadrant_classifiers_details}).

\clearpage

\section{Sensitivity to numerical integration and input selection tolerance}
\label{sec:sensitivity}

We analyse how the integration step $\Delta t$ of the coherent counterfactual ablation trajectories (\Cref{subsec:counterfactual_generation}) and the selection tolerance $\eta$ for the inputs (\Cref{subsec:probe_contexts}) influence the resulting complexes, and use this analysis to choose the values adopted in the rest of the article.
We integrate the input trajectory $x(t)$ by explicit Euler steps of size $\Delta t$, and shorten a step whenever an active predictor would otherwise pass zero.
The integration step only discretises this trajectory, so the complexes should not depend on it once it is small enough.
The selection tolerance, in contrast, determines the region of input space from which the inputs are drawn, so it is part of the definition of the set of inputs and may have a subtle effect on the resulting complexes.

The experiments used the handcrafted monosemantic and polysemantic quadrant classifiers with the one-hot inputs of all four quadrants, as in \Cref{subsec:joint_sensitivity}.
The parameters of the relevance test were held at the values chosen in \Cref{sec:statistical_sensitivity} ($B=128$, $\varepsilon=0.01$, $\alpha=0.01$, Bonferroni correction, and $K=1000$), and we built complexes up to the maximum possible simplex dimension.
We varied the integration step over
\begin{align}
	\Delta t
	&\in\{0.005,0.01,0.02,0.05,0.1,0.2,0.5\},
\end{align}
with one fixed input set per quadrant, drawn at $\eta=0.01$.
Separately, we varied the selection tolerance over
\begin{align}
	\eta
	&\in\{0,0.001,0.005,0.01,0.05,0.1\}
\end{align}
at $\Delta t=0.01$.
For each value of $\eta$, the input set was drawn from the same sequence of uniformly sampled points, so the input sets for different values of $\eta$ partly overlap.
For each value of $\Delta t$ and of $\eta$, we thus obtained eight complexes, one for each handcrafted network and quadrant.

\begin{figure}[t]
	\centering
	\includegraphics[width=\linewidth]{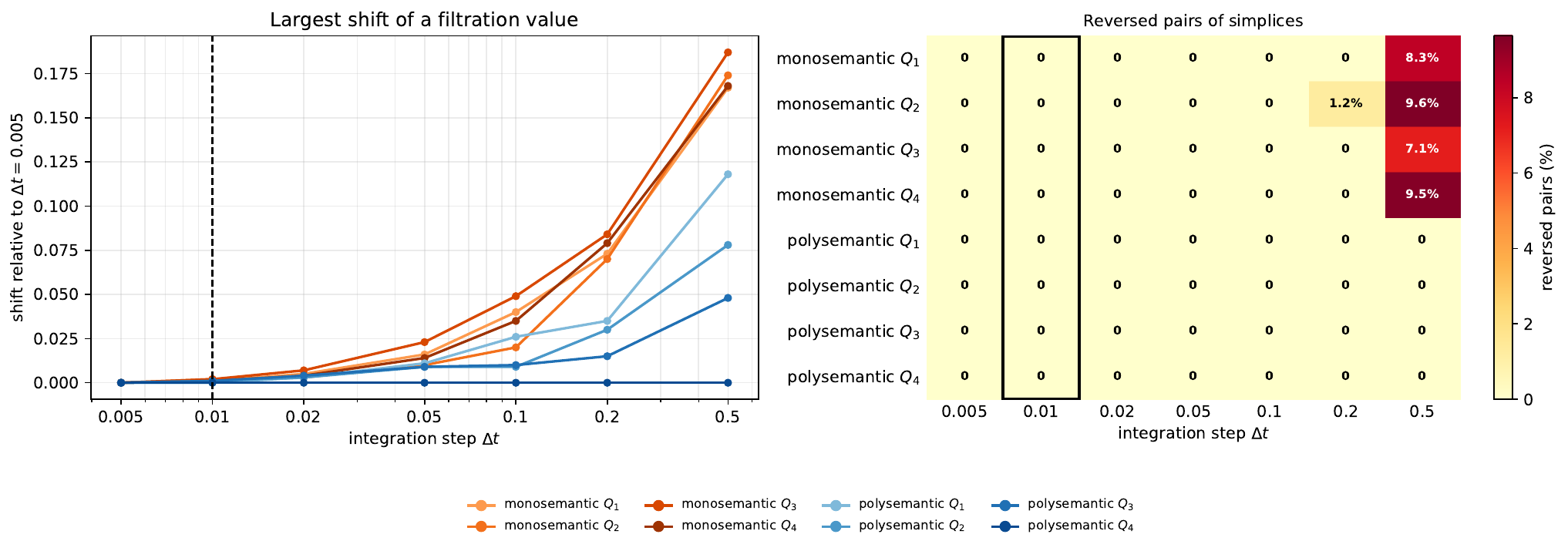}
	\caption{
		Effect of the integration step $\Delta t$ on the filtration values, relative to the finest step $\Delta t=0.005$.
		The simplex sets are the same at every $\Delta t$.
		The left panel shows the largest shift of a filtration value in each of the eight complexes.
		The right panel shows the percentage of pairs of simplices whose order in the filtration is reversed, among the pairs with distinct filtration values at $\Delta t=0.005$.
		The dashed line and the outlined column mark the chosen step $\Delta t=0.01$.
	}
	\label{fig:hyperparameter_dt_stability}
\end{figure}

\paragraph{Integration step.}
All seven integration steps from $\Delta t=0.005$ to $\Delta t=0.5$ returned the same simplex set in all eight complexes.
Compared with the finest step $\Delta t=0.005$, the step $\Delta t=0.01$ shifted filtration values by at most $0.002$ and did not change the order of any two simplices in the filtration.
Coarser steps shifted filtration values further, by up to $0.19$ at $\Delta t=0.5$ (\Cref{fig:hyperparameter_dt_stability}).
From $\Delta t=0.2$ on, they also reversed the order of up to eight pairs of simplices in the monosemantic complexes.

Since halving the step to $\Delta t=0.005$ changed neither the simplex sets nor the order of the simplices, we use $\Delta t=0.01$ in all other experiments of this article (\Cref{sec:statistical_sensitivity,sec:quadrant_classifiers_details}).

\paragraph{Selection tolerance.}
For each network and quadrant, we compared the complex at each $\eta$ with the complex at the smallest $\eta$ at which $B=128$ inputs could be collected within 128,000 draws (\Cref{subsec:probe_contexts}).
This is $\eta=0$ except for the input sets $X_{Q_2}$ and $X_{Q_4}$ of the polysemantic network, where it is $\eta=0.001$.

\begin{figure}[t]
	\centering
	\includegraphics[width=\linewidth]{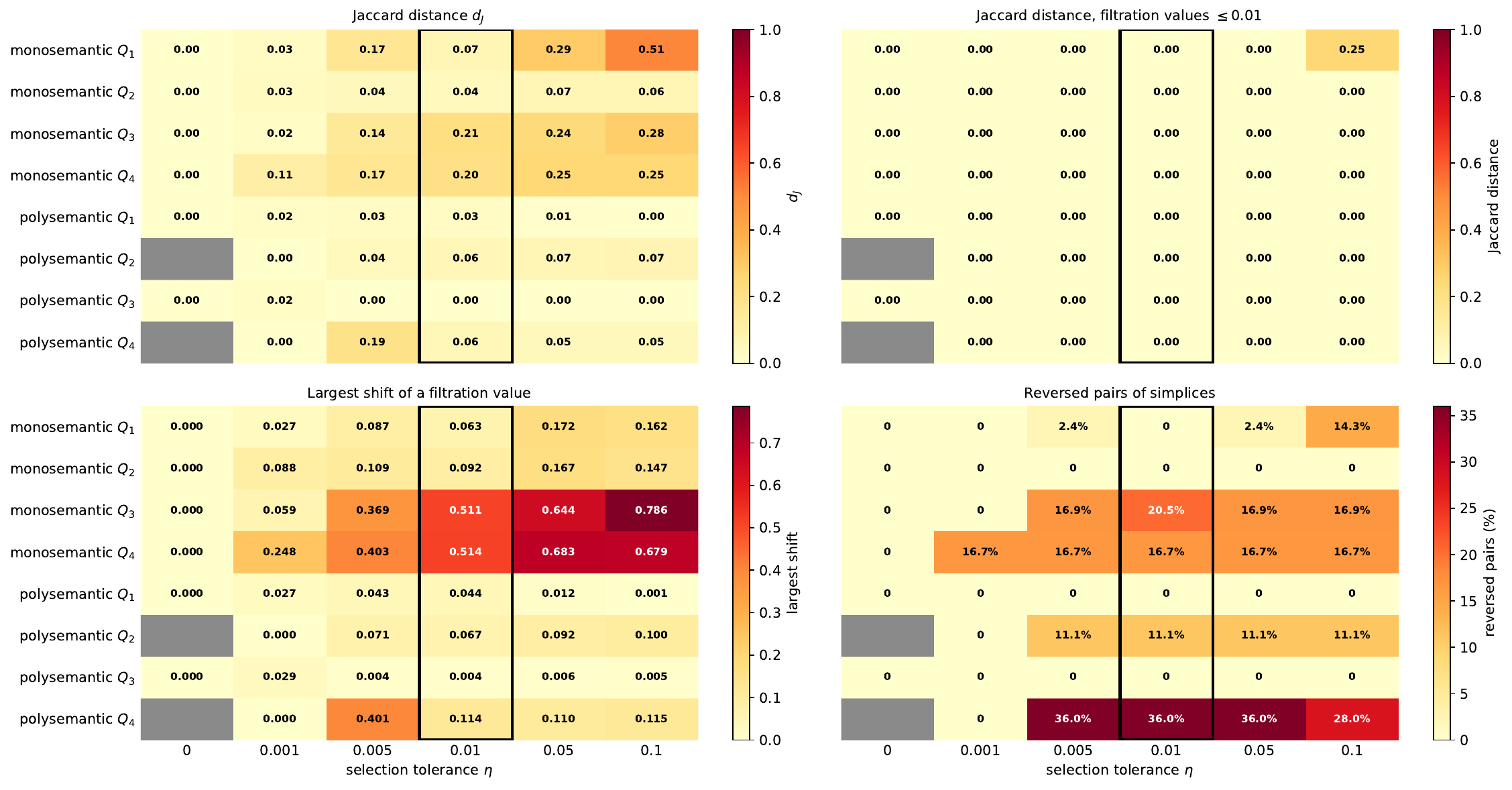}
	\caption{
		Effect of the selection tolerance $\eta$ on the complexes, relative to the complex at the smallest $\eta$ at which $B=128$ inputs were accepted for each input set.
		The top left panel shows the Jaccard distance $d_J$ (\Cref{subsec:jaccard_distance}).
		The top right panel shows the Jaccard distance between the sets of simplices with filtration value at most $0.01$, without vertices.
		The bottom left panel shows the largest shift of a filtration value.
		The bottom right panel shows the percentage of pairs of simplices whose order in the filtration is reversed, among the pairs with distinct filtration values in the reference.
		Grey cells mark $X_{Q_2}$ and $X_{Q_4}$ of the polysemantic network, for which fewer than $B=128$ inputs were accepted at $\eta=0$, and the outlined column marks the chosen $\eta=0.01$.
	}
	\label{fig:hyperparameter_probe_stability}
\end{figure}

Between $\eta=0.001$ and $\eta=0.05$, the simplex set did not change in any of the eight complexes.
Differences occurred only at the ends of the range, for $X_{Q_1}$ of the monosemantic network.
There, the complex at $\eta=0$ lacked two edges and two 2-simplices, and at $\eta=0.1$ four simplices were replaced by three others.
In contrast, the filtration values changed with $\eta$.
At $\eta=0.01$, they differed from the reference by up to $0.51$, and the order of up to $36\%$ of the pairs of simplices was reversed (\Cref{fig:hyperparameter_probe_stability}).
These changes are expected, because a larger $\eta$ accepts inputs from a larger region of input space and therefore gives a different input set.

Since the simplex sets did not depend on $\eta$ between $0.001$ and $0.05$ and the changes in the filtration values are expected, we use $\eta=0.01$ in all other experiments of this article (\Cref{sec:statistical_sensitivity,sec:quadrant_classifiers_details}).

\clearpage

\section{Computational complexity}
\label{sec:computational-complexity}

The cost of constructing a filtered DSC for neural network with $N$ internal neurons and an input set of size $B$ is dominated by the number of candidate simplices.
For each predictor set, the construction integrates the $B$ ablation trajectories once, with $O(1/\Delta t)$ steps each (\Cref{sec:sensitivity}), and reuses them for all candidates that extend it (\Cref{subsec:dsc_construction}).
Each candidate is then tested at all $K+1$ ablation values.
Each of these tests ranks the $B$ changes in target activation, which takes $O(B\log B)$ operations.
Up to dimension $k$, there are $O(N^k)$ predictor sets and $O(N^{k+1})$ candidates (\Cref{sec:introduction}).
If every candidate were admitted, the cost would therefore be
\begin{align}
	O\bigl(N^kB/\Delta t+N^{k+1}KB\log B\bigr).
	\label{eq:construction_cost}
\end{align}

In practice, far fewer candidates are tested, because a candidate is tested only if all its codimension-one faces have been admitted.
In the parameter sweep of \Cref{subsec:joint_sensitivity}, for example, no dimension required more than $104$ tests, while the monosemantic network with its $N=8$ internal neurons would require $6{,}720$ tests at dimension $4$ if every candidate were admitted.

The memory requirement is dominated by the ablation trajectories, which hold $(K+1)BN$ activation values for each predictor set and thus grow linearly in the number $K+1$ of ablation values, the sample size $B$ and the number of neurons $N$.
The tests for different predictor sets of the same dimension are independent of each other and can therefore run in parallel, as can the constructions for different input sets.

The cost is nearly the same for all relevance margins $\varepsilon>0$ but increases considerably at $\varepsilon=0$.
In this case, the relevance test admits every link that the input set can resolve, so many more candidates are tested at the next dimension.
We measure the cost in CPU-hours, that is, the wall-clock time of a construction multiplied by the number of CPUs allocated to it.
At the parameters of the main experiments, a single complex of a handcrafted network takes about a minute on a single CPU.
In the parameter sweep, the constructions for each relevance margin $\varepsilon>0$ took about $0.7$ CPU-hours in total, compared with $3.0$ CPU-hours at $\varepsilon=0$.
The whole sweep took $9.3$ CPU-hours and ran in $13$ minutes on $48$ CPUs, and the sweep of $\Delta t$ and $\eta$ in \Cref{sec:sensitivity} ran in $3$ minutes on $48$ CPUs.

Since the number of candidates grows like $N^{k+1}$, the construction as presented is constrained to small multilayer perceptrons.
With some further research, its ideas might still apply to larger networks.
The construction could, for example, be restricted to candidates of interest, such as the neurons of a few adjacent layers or of a circuit found by another method, or the simplices predicted by a hypothesis about the computation of the network.
The vertices of the complex could also be groups of neurons, which would reduce their number $N$.

\clearpage

\section{Results on the quadrant classifiers}
\label{sec:quadrant_classifiers_details}

All complexes in this section are constructed from sets of $B=128$ inputs selected with selection tolerance $\eta=0.01$ (\Cref{subsec:probe_contexts}).
The remaining parameters are set to the values chosen in \Cref{sec:statistical_sensitivity,sec:sensitivity} ($\Delta t=0.01$, $K=1000$, $\varepsilon=0.01$ and $\alpha=0.01$ with Bonferroni correction), and each complex is built up to the highest dimension that the number of neurons allows.
We write $\mathcal{K}_{Q_i}$, $\mathcal{K}_{Q_iQ_j}$ and $\mathcal{K}_{\mathrm{rand}}$ for the complexes of the one-hot inputs $X_{Q_i}$, the two-hot inputs $X_{Q_iQ_j}$ and the random inputs $X_{\mathrm{rand}}$.

\subsection{Handcrafted networks}
\label{subsec:handcrafted_dsc_logic}

For the handcrafted networks of \Cref{subsec:quadrant_networks}, the logic of the networks is simple enough to explain the simplices of their complexes.
Ablating a set of neurons moves the inputs along a path in the input plane (\Cref{subsec:counterfactual_generation}), and the relevance test detects whether the activation of a target neuron changes along this path.
Below, we describe the logic of the monosemantic and polysemantic networks and explain how it is reflected in their complexes.

\paragraph{Monosemantic network.}
In the monosemantic network, the first-layer neurons $n_1$, $n_2$, $n_3$ and $n_4$ each detect one sign of $x$ or $y$, and the second-layer neurons $n_5$, $n_6$, $n_7$ and $n_8$ construct the quadrants $Q_1$, $Q_2$, $Q_3$ and $Q_4$ by combining these signs (\Cref{subsec:quadrant_networks}).
The one-hot inputs activate the neurons as follows.

\begin{center}
	\begin{tabular}{c|c|c}
		inputs & active first-layer neurons & active second-layer neuron \\
		\hline
		$X_{Q_1}$ & $a_1=x$, $a_4=y$ & $a_5=x+y$ \\
		$X_{Q_2}$ & $a_2=-x$, $a_4=y$ & $a_6=-x+y$ \\
		$X_{Q_3}$ & $a_2=-x$, $a_3=-y$ & $a_7=-x-y$ \\
		$X_{Q_4}$ & $a_1=x$, $a_3=-y$ & $a_8=x-y$
	\end{tabular}
\end{center}
This network classifies the quadrants exactly, and its accuracy on the evaluation set is $1.0$.

\begin{figure}[p]
	\centering
	\includegraphics[width=.9\linewidth, height=.82\textheight, keepaspectratio]{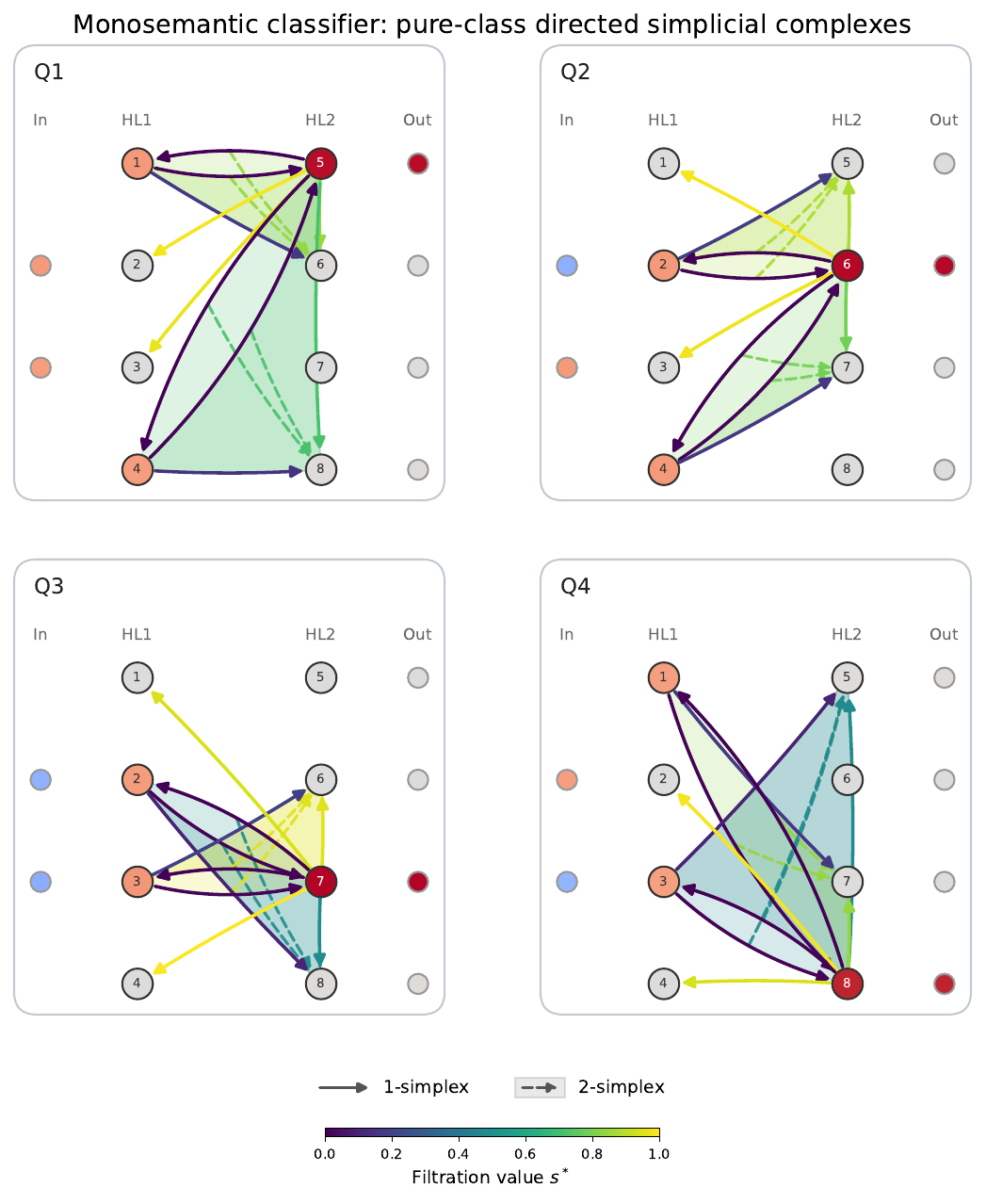}
	\caption{Complexes of the one-hot inputs of the monosemantic network.
	The fill of a node shows the mean activation of the neuron over the input set.
	Arrows are directed edges, coloured by filtration value.
	Translucent patches and dashed arrows show 2-simplices.
	After relabelling the neurons, the four complexes coincide.
	In each of them, the two active first-layer neurons are connected in both directions to the active second-layer neuron at low filtration values.
	The 2-simplices connect the active second-layer neuron and one first-layer neuron to the second-layer neuron of an adjacent quadrant.}
	\label{fig:appendix_mono_dsc_logic}
\end{figure}

The monosemantic network is symmetric under rotations of the input domain by $90^\circ$.
Such a rotation permutes $a_1$, $a_2$, $a_3$ and $a_4$ cyclically.
Up to the corresponding relabelling of all neurons, the rotation therefore maps each quadrant to the next and leaves the logic of the network computation unchanged.
Accordingly, the four complexes coincide after this relabelling, and each of them contains ten edges and four 2-simplices (\Cref{fig:appendix_mono_dsc_logic}).
Since the input sets for each quadrant are chosen independently, this agreement is a consistency check of the construction.

In $\mathcal{K}_{Q_1}$, the four edges between the active neurons $n_1$, $n_4$ and $n_5$ enter first, at filtration values below $0.01$.
The edge $n_1\to n_5$ arises because ablating $n_1$ moves the inputs horizontally towards the vertical decision boundary $x=0$, where $a_1$ vanishes.
Since $a_5=a_1+a_4$ on these inputs, the activation of $n_5$ drops along the way, and the relevance test detects this shift almost immediately.
The edge $n_4\to n_5$ arises in the same way, because ablating $n_4$ moves the inputs vertically towards $y=0$.
The backward edges $n_5\to n_1$ and $n_5\to n_4$ arise because ablating $n_5$ moves the inputs in the direction $(-1,-1)$, which is the most direct way to reduce $a_5=x+y$, so that $a_1$ and $a_4$ drop with it.
They record that $n_5$ cannot be switched off without also reducing its inputs, although the network has no connection from $n_5$ back to $n_1$ or $n_4$.

The remaining simplices describe how the ablation moves the inputs out of $Q_1$.
Ablating $n_1$ moves the inputs towards $Q_2$, and since $n_1$ is an inhibitory input of the $Q_2$ neuron $n_6$, this activates $n_6$ already before the inputs reach the decision boundary.
This gives the edge $n_1\to n_6$, and ablating $n_4$ gives the edge $n_4\to n_8$ towards $Q_4$ in the same way.
Ablating $n_5$ alone moves the inputs parallel to the diagonal, so only the inputs above the diagonal approach $Q_2$, and those close to the diagonal only shortly before $n_5$ is switched off.
The edge $n_5\to n_6$ therefore enters only at $0.84$, much later than $n_1\to n_6$ at $0.17$.
The edge $n_5\to n_8$ enters late for the same reason, through the inputs below the diagonal.
Ablating $n_1$ and $n_5$ together reduces $a_1$ and $a_5=a_1+a_4$ at equal rates and thus leaves $a_4$ unchanged, so the inputs move horizontally towards $Q_2$ as for $n_1$ alone.
This gives the 2-simplices $(n_1,n_5,n_6)$ and $(n_5,n_1,n_6)$.
Both are tested by the same joint ablation, and both are present because the complex contains both edges $n_1\to n_5$ and $n_5\to n_1$.
The effect of the joint ablation on $n_6$ sets in early, at $s^*=0.094$, but the two 2-simplices enter only together with their common face $n_5\to n_6$, because a simplex cannot enter before its faces.
In the same way, ablating $n_4$ and $n_5$ together gives the 2-simplices $(n_4,n_5,n_8)$ and $(n_5,n_4,n_8)$.
The last two edges $n_5\to n_2$ and $n_5\to n_3$ enter only close to full ablation of $n_5$, at $0.98$ and $0.97$, when the inputs cross the decision boundaries where the inactive neurons $n_2$ and $n_3$ become active.

The complex contains no 2-simplex with the predictors $n_1$ and $n_4$, although $n_5$ combines them like an AND gate, since an input lies in $Q_1$ only if $x>0$ and $y>0$.
Such a 2-simplex would require an edge between $n_1$ and $n_4$.
Since $a_1=x$ and $a_4=y$ depend on different input coordinates, ablating $n_1$ moves the inputs horizontally and leaves $a_4$ unchanged, and ablating $n_4$ leaves $a_1$ unchanged in the same way.
The relevance test therefore finds no effect of $n_1$ on $n_4$ or of $n_4$ on $n_1$ at any ablation value, and the complex contains no edge between them.
As a result, the AND gate appears only as the two edges $n_1\to n_5$ and $n_4\to n_5$.

\paragraph{Polysemantic network.}
In the polysemantic network, the first-layer neurons $n_1$, $n_2$ and $n_3$ measure how far the input lies in the direction of the quadrants $Q_1$, $Q_2$ and $Q_3$, and the second-layer neurons $n_4$, $n_5$ and $n_6$ copy them (\Cref{subsec:quadrant_networks}).
The fourth second-layer neuron $n_7$ represents $Q_4$ by being active when the total activation $a_1+a_2+a_3$ of the first layer is below $2$.
For the one-hot inputs of $Q_1$ and $Q_3$, one first-layer neuron and its copy are active.
For those of $Q_2$ and $Q_4$, the sign of $x+y$ divides the inputs into two groups of about equal size with different active neurons.
The active neurons have the following activations on these inputs.

\begin{center}
	\begin{tabular}{cc|c|c}
		\multicolumn{2}{c|}{inputs} & active first-layer neurons & active second-layer neurons \\
		\hline
		\multicolumn{2}{c|}{$X_{Q_1}$} & $a_1=x+y$ & $a_4=a_1$ \\
		\multirow{2}{*}{$X_{Q_2}$} & $x+y>0$ & $a_1=x+y$, $a_2=-x+y$ & $a_4=a_1$, $a_5=a_2$ \\
		& $x+y<0$ & $a_2=-x+y$, $a_3=-x-y$ & $a_5=a_2$, $a_6=a_3$ \\
		\multicolumn{2}{c|}{$X_{Q_3}$} & $a_3=-x-y$ & $a_6=a_3$ \\
		\multirow{2}{*}{$X_{Q_4}$} & $x+y>0$ & $a_1=x+y$ & $a_4=a_1$, $a_7=2-a_1$ \\
		& $x+y<0$ & $a_3=-x-y$ & $a_6=a_3$, $a_7=2-a_3$
	\end{tabular}
\end{center}
This network only approximates the quadrants, and its accuracy on the evaluation set is $0.866$.
Since we select the input sets based on the network's outputs, the complexes describe the logic of the network, which does not necessarily coincide with the features of the quadrants.

\begin{figure}[p]
	\centering
	\includegraphics[width=.9\linewidth, height=.82\textheight, keepaspectratio]{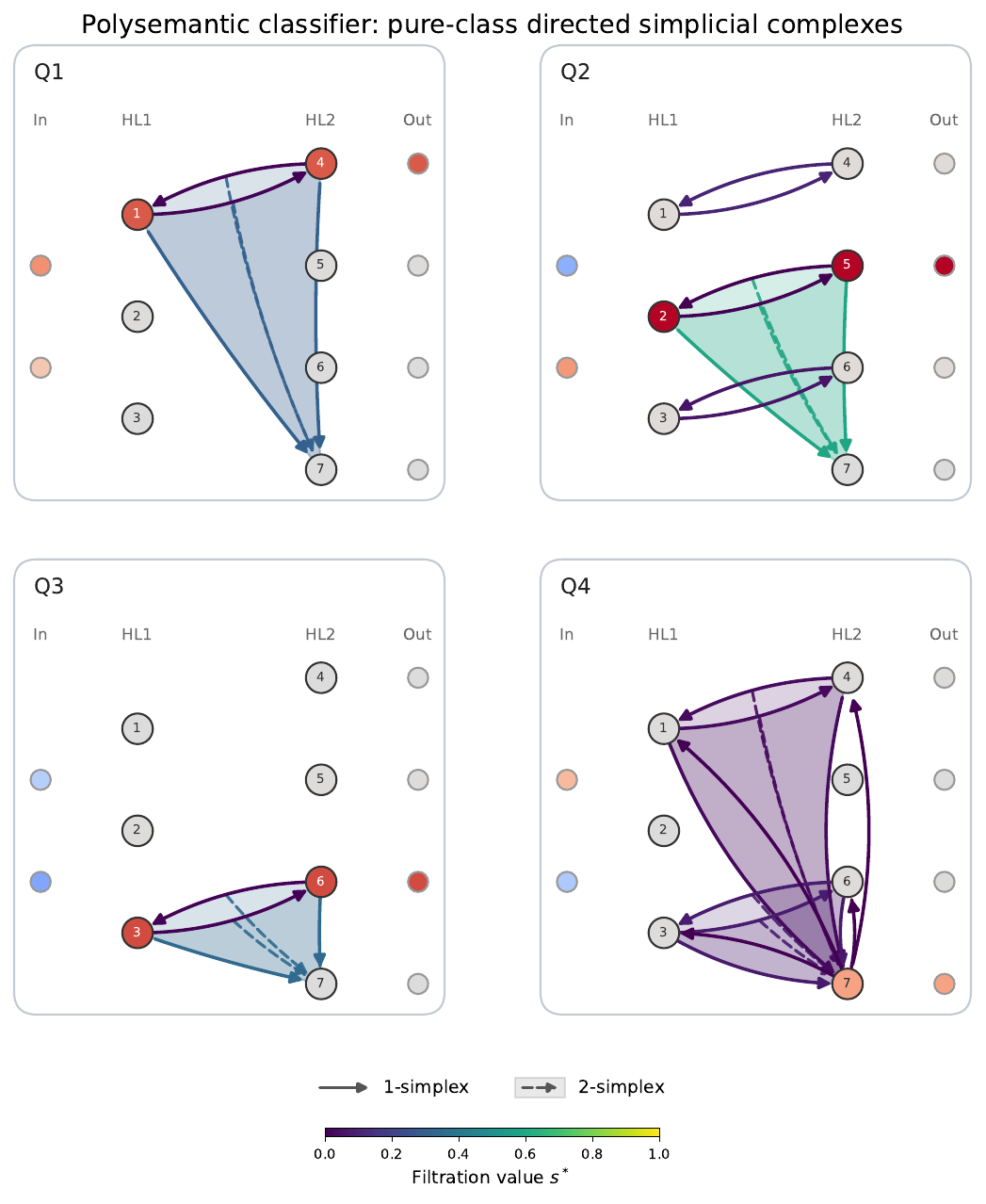}
	\caption{Complexes of the one-hot inputs of the polysemantic network, drawn as in \Cref{fig:appendix_mono_dsc_logic}.
	Each active first-layer neuron is connected in both directions with its copy in the second layer.
	Every 2-simplex has the target $n_7$, and $\mathcal{K}_{Q_4}$ contains all simplices of $\mathcal{K}_{Q_1}$ and $\mathcal{K}_{Q_3}$.}
	\label{fig:appendix_poly_dsc_logic}
\end{figure}

The polysemantic network has a smaller symmetry than the monosemantic network.
Reflecting the input domain in the line $x+y=0$ swaps $a_1$ and $a_3$ and leaves $a_2$ unchanged.
Up to a relabelling that swaps $n_1$ with $n_3$ and $n_4$ with $n_6$, the reflection therefore exchanges $Q_1$ and $Q_3$ and maps $Q_2$ and $Q_4$ to themselves.
Accordingly, $\mathcal{K}_{Q_1}$ and $\mathcal{K}_{Q_3}$ coincide after this relabelling, while $\mathcal{K}_{Q_2}$ and $\mathcal{K}_{Q_4}$ are each mapped to themselves (\Cref{fig:appendix_poly_dsc_logic}).

In $\mathcal{K}_{Q_1}$, the two edges between the active neurons $n_1$ and $n_4$ enter first, at filtration values below $0.01$.
The edge $n_1\to n_4$ arises because ablating $n_1$ moves the inputs in the direction $(-1,-1)$ towards the line $x+y=0$, where $a_1$ vanishes.
Since $a_4=a_1$, the activation of $n_4$ drops along the way, and the relevance test detects this shift almost immediately.
Ablating $n_4$ moves the inputs along the same path, so the backward edge $n_4\to n_1$ enters at the same filtration value.
It records that $n_4$ cannot be switched off without also reducing its input $n_1$.
This path leaves $y-x$ and hence $a_2$ unchanged, so the complex contains no edge to $n_2$ or $n_5$.
Along the path, $n_7$ becomes active only when $a_1+a_2+a_3$ drops below $2$, that is, when the inputs approach the region that the network assigns to $Q_4$.
The edges $n_1\to n_7$ and $n_4\to n_7$ therefore enter only at $0.31$, much later than the edges between $n_1$ and $n_4$.
Ablating $n_1$ and $n_4$ together moves the inputs along the same path again.
This gives the 2-simplices $(n_1,n_4,n_7)$ and $(n_4,n_1,n_7)$, which enter together with their faces $n_1\to n_7$ and $n_4\to n_7$.

In $\mathcal{K}_{Q_2}$, the edges between $n_2$ and $n_5$ enter first, below $0.01$, in the same way as those between $n_1$ and $n_4$ in $\mathcal{K}_{Q_1}$.
The edges between $n_3$ and $n_6$ and between $n_1$ and $n_4$ enter later, at $0.053$ and $0.10$, because $n_1$ and $n_3$ are each active on only about half of the inputs.
The other inputs do not respond to the ablation of this neuron, which delays the detection of the shift.
Ablating $n_2$ moves the inputs in the direction $(1,-1)$, which leaves $x+y$ and hence $a_1$ and $a_3$ unchanged, so the complex contains no edges from $n_2$ or $n_5$ to $n_1$, $n_3$, $n_4$ or $n_6$.
Along this path, $n_7$ becomes active only when the inputs approach the region assigned to $Q_4$, as for $X_{Q_1}$.
The edges $n_2\to n_7$ and $n_5\to n_7$ therefore enter only at $0.59$, even later than $n_1\to n_7$ in $\mathcal{K}_{Q_1}$ at $0.31$, because the inputs of $X_{Q_2}$ lie further from this region.
Ablating $n_2$ and $n_5$ together gives the 2-simplices $(n_2,n_5,n_7)$ and $(n_5,n_2,n_7)$ in the same way.

The inputs of $X_{Q_4}$ lie close to the line $x+y=0$ with $x>0$, where $n_7$ is active because of its bias.
Those above the line activate $n_1$ and $n_4$ as the inputs of $X_{Q_1}$ do, and those below it activate $n_3$ and $n_6$ as the inputs of $X_{Q_3}$ do.
Accordingly, $\mathcal{K}_{Q_4}$ contains all simplices of $\mathcal{K}_{Q_1}$ and $\mathcal{K}_{Q_3}$, and in addition the four edges from $n_7$ to $n_1$, $n_4$, $n_3$ and $n_6$.
The differences arise because $n_7$ is active on $X_{Q_4}$ and inactive on $X_{Q_1}$ and $X_{Q_3}$.
Above the line, $a_7=2-a_1$, so ablating $n_7$ moves these inputs in the direction $(1,1)$ and increases $a_1$ and $a_4$.
Below the line, it moves the inputs in the direction $(-1,-1)$ and increases $a_3$ and $a_6$.
Conversely, ablating $n_1$ moves the inputs along the same path as for $X_{Q_1}$, but $a_7$ grows immediately, whereas for $X_{Q_1}$ it stays zero until $a_1$ drops below $2$.
The simplices with target $n_7$ therefore enter below $0.08$, much earlier than the $0.31$ in $\mathcal{K}_{Q_1}$ and $0.34$ in $\mathcal{K}_{Q_3}$.
The edges between $n_1$ and $n_4$, in contrast, enter later than in $\mathcal{K}_{Q_1}$, at $0.029$ compared with $0.004$, because only about half of the inputs of $X_{Q_4}$ lie above the line.

The complex contains no 2-simplex with the predictors $n_1$ and $n_7$, although it contains the edges between them in both directions.
On the inputs above the line, $a_7=2-a_1$, so $n_1$ and $n_7$ cannot be ablated together, and their joint ablation leaves these inputs in place.
The same holds for $n_3$ and $n_7$.
$\mathcal{K}_{Q_4}$ is the largest of the four complexes, because the ablation of $n_7$ moves its inputs out of $Q_4$ in two directions.

Since $n_7$ belongs to every 2-simplex, it belongs to more maximal simplices than any other neuron, $14$ across the four complexes against two to seven for each of the six neurons that mix both coordinates.
The number of simplices a neuron belongs to therefore does not identify the polysemantic neurons.

\subsection{Trained networks}
\label{subsec:quadrant_trained}

\begin{figure}[p]
	\centering
	\includegraphics[width=.48\linewidth,height=.35\textheight,keepaspectratio]{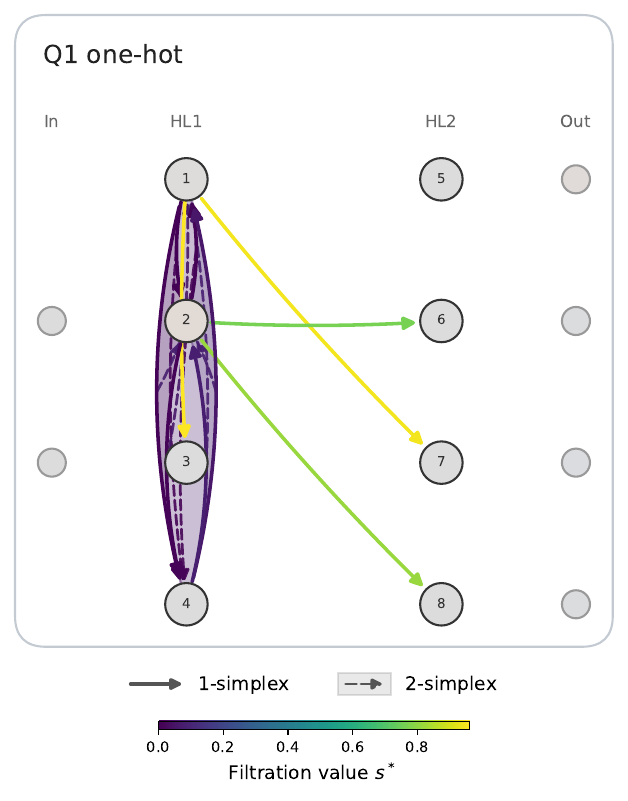}
	\includegraphics[width=.48\linewidth,height=.35\textheight,keepaspectratio]{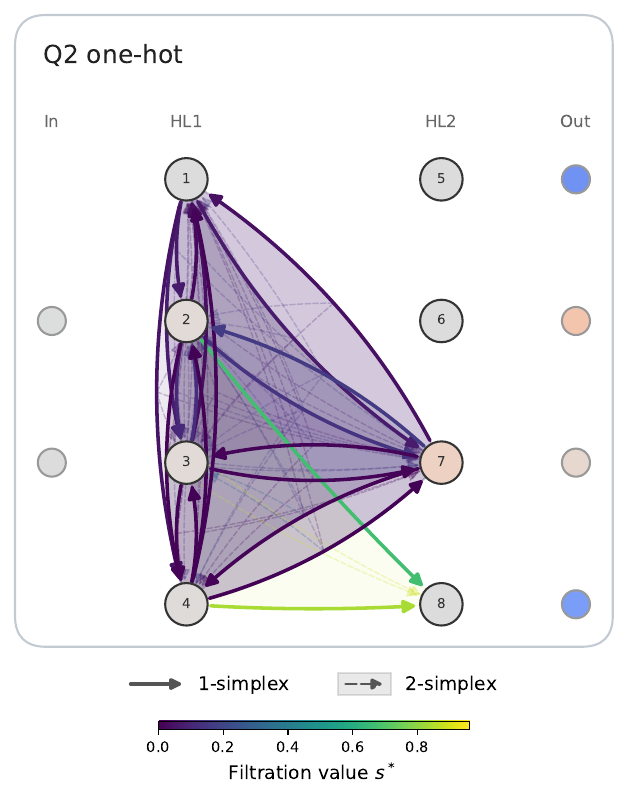}

	\includegraphics[width=.48\linewidth,height=.35\textheight,keepaspectratio]{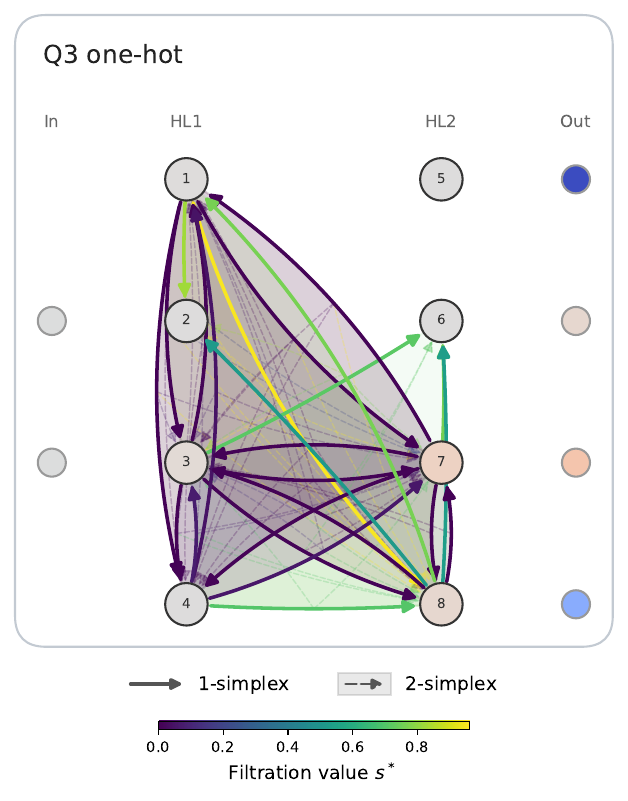}
	\includegraphics[width=.48\linewidth,height=.35\textheight,keepaspectratio]{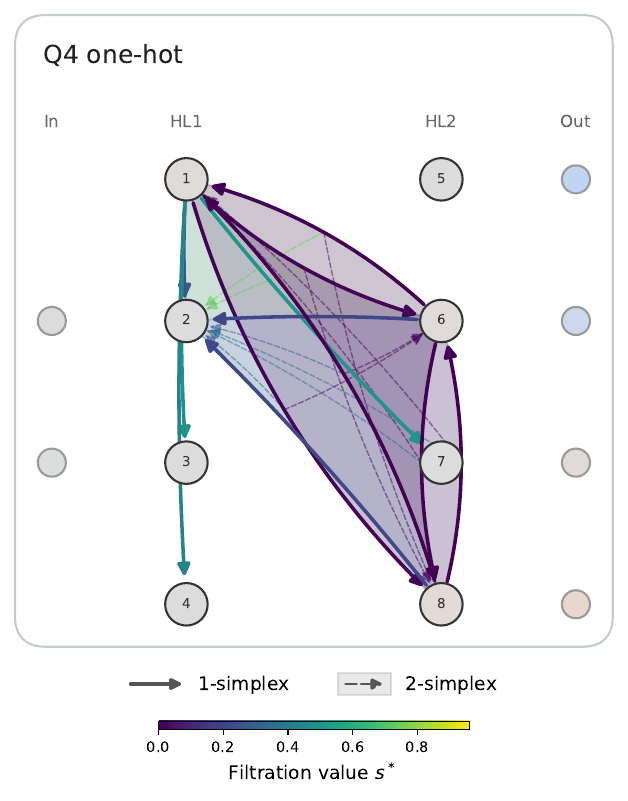}
	\caption{Complexes of the one-hot inputs of the trained $[2,4,4,4]$ network up to dimension two, drawn as in \Cref{fig:appendix_mono_dsc_logic}.
	\Cref{tab:appendix_trained_counts} gives the numbers of simplices of dimension three and four.}
	\label{fig:appendix_mono_trained_dscs}
\end{figure}

\begin{figure}[p]
	\centering
	\includegraphics[width=.48\linewidth,height=.35\textheight,keepaspectratio]{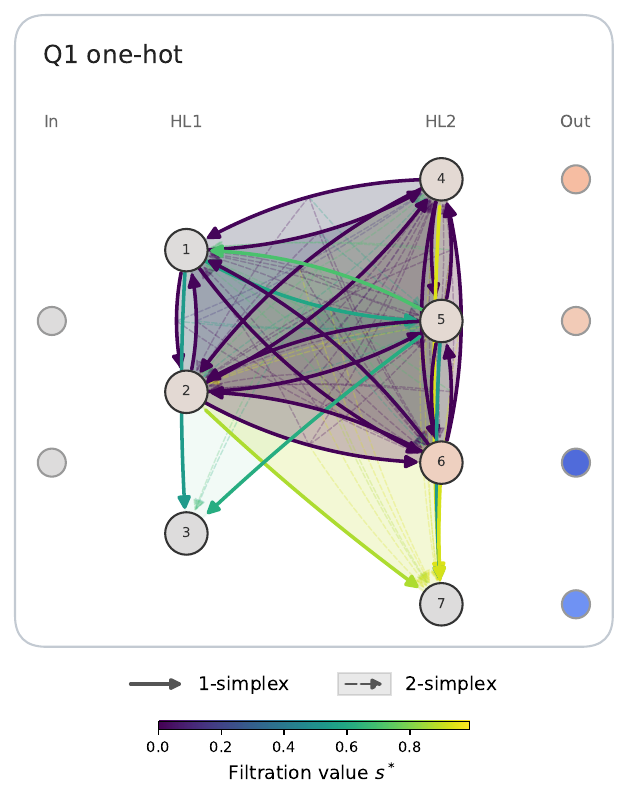}
	\includegraphics[width=.48\linewidth,height=.35\textheight,keepaspectratio]{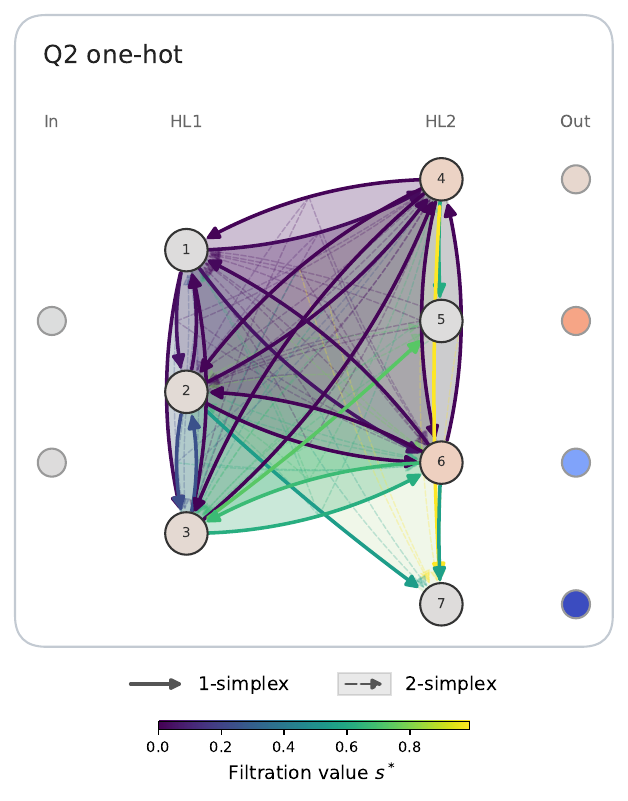}

	\includegraphics[width=.48\linewidth,height=.35\textheight,keepaspectratio]{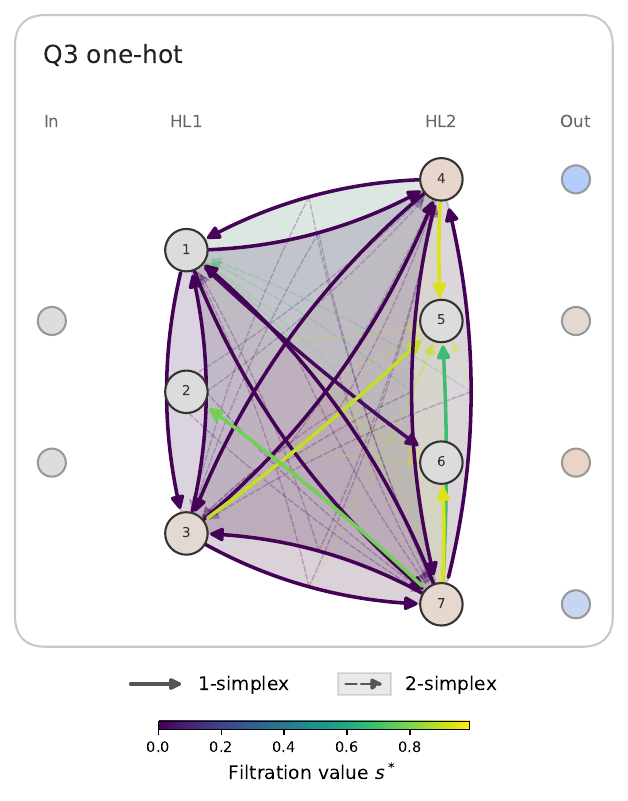}
	\includegraphics[width=.48\linewidth,height=.35\textheight,keepaspectratio]{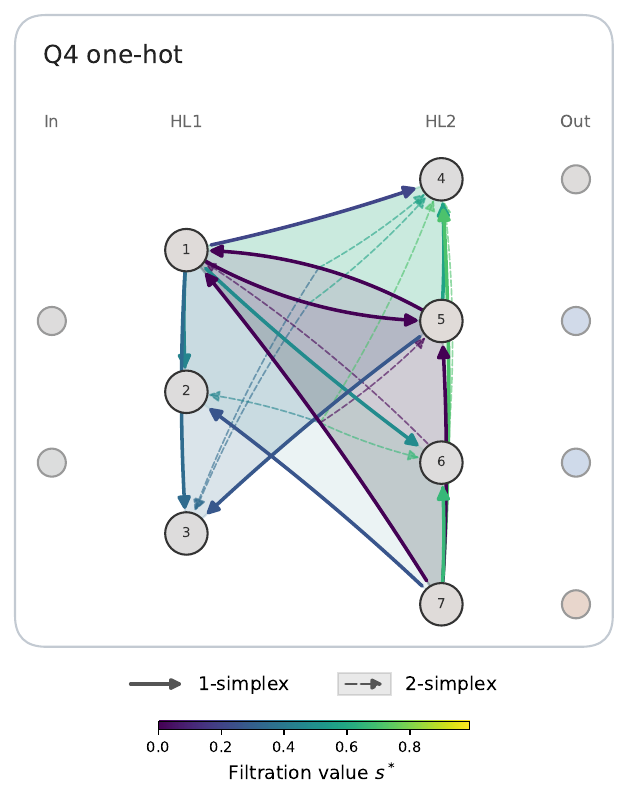}
	\caption{Complexes of the one-hot inputs of the trained $[2,3,4,4]$ network, drawn as in \Cref{fig:appendix_mono_trained_dscs}.}
	\label{fig:appendix_poly_trained_dscs}
\end{figure}

The logic of the trained networks is unknown, so we cannot explain their complexes as we did for the handcrafted networks in \Cref{subsec:handcrafted_dsc_logic}.
\Cref{tab:appendix_trained_counts} compares the sizes of their complexes $\mathcal{K}_{Q_i}$ with those of the handcrafted networks, and \Cref{fig:appendix_mono_trained_dscs,fig:appendix_poly_trained_dscs} show the trained complexes up to dimension two.

For each quadrant, the complex of a trained network is larger than that of the handcrafted network with the same architecture.
The handcrafted monosemantic network gives the same complex with $22$ simplices for all four quadrants, whereas the complexes of the trained $[2,4,4,4]$ network differ between the quadrants and contain between $24$ and $220$ simplices.
For the $[2,3,4,4]$ architecture, the handcrafted complexes contain between $13$ and $23$ simplices and the trained complexes between $32$ and $365$.

Neither handcrafted network has simplices above dimension two.
The trained $[2,4,4,4]$ network reaches dimension four for $Q_2$ and $Q_3$, and the trained $[2,3,4,4]$ network for $Q_1$ and $Q_2$.

The complexes of five of the eight trained input sets depend on the multiple-testing correction, whereas those of the handcrafted networks do not (\Cref{subsec:joint_sensitivity}).
The largest difference occurs for $\mathcal{K}_{Q_3}$ of the trained $[2,3,4,4]$ network.
Without correction, it contains $24$ edges and $196$ simplices of higher dimension, and with the Bonferroni correction $18$ edges and $62$ simplices of higher dimension.

Of the $1{,}475$ candidates tested with the Bonferroni correction, $21$ are rejected but would be admitted without correction.
Since a candidate is tested only if all its faces are admitted, these $21$ rejections keep a further $206$ candidates from being tested.
For $\mathcal{K}_{Q_3}$ of the trained $[2,3,4,4]$ network, the entire difference comes from six edges that contain $n_6$.
Their p-values lie between $9.1\times10^{-4}$ and $5.4\times10^{-3}$, below $\alpha=0.01$ but above the Bonferroni cutoff of $2.4\times10^{-4}$.
In the handcrafted networks, no p-value lies between these two cutoffs.

In the handcrafted networks, the p-value of every admitted simplex stays below the cutoff once it has crossed it.
In the trained networks, the p-value of $92$ of the $1{,}184$ admitted simplices crosses the cutoff more than once, and $s^*$ is the last of these crossings.
A further $20$ candidates pass the cutoff at an intermediate ablation value but not at full ablation, so the decision rule of \Cref{subsec:dsc_construction} does not admit them.

\begin{table}[htbp]
	\centering
	\begin{tabular}{l l | c c c c | c | c}
		& & \multicolumn{4}{c|}{simplices by dimension} & & \\
		network & quadrant & 1 & 2 & 3 & 4 & max dim & total \\
		\hline
		$[2,4,4,4]$, handcrafted & $Q_1$--$Q_4$ & 10 & 4 & --- & --- & 2 & 22 \\
		\hline
		$[2,4,4,4]$, trained & $Q_1$ & 10 & 6 & --- & --- & 2 & 24 \\
		$[2,4,4,4]$, trained & $Q_2$ & 22 & 56 & 86 & 48 & 4 & 220 \\
		$[2,4,4,4]$, trained & $Q_3$ & 24 & 50 & 42 & 8 & 4 & 132 \\
		$[2,4,4,4]$, trained & $Q_4$ & 12 & 12 & 6 & --- & 3 & 38 \\
		\hline
		$[2,3,4,4]$, handcrafted & $Q_1$ & 4 & 2 & --- & --- & 2 & 13 \\
		$[2,3,4,4]$, handcrafted & $Q_2$ & 8 & 2 & --- & --- & 2 & 17 \\
		$[2,3,4,4]$, handcrafted & $Q_3$ & 4 & 2 & --- & --- & 2 & 13 \\
		$[2,3,4,4]$, handcrafted & $Q_4$ & 12 & 4 & --- & --- & 2 & 23 \\
		\hline
		$[2,3,4,4]$, trained & $Q_1$ & 26 & 74 & 138 & 120 & 4 & 365 \\
		$[2,3,4,4]$, trained & $Q_2$ & 25 & 68 & 126 & 120 & 4 & 346 \\
		$[2,3,4,4]$, trained & $Q_3$ & 18 & 32 & 30 & --- & 3 & 87 \\
		$[2,3,4,4]$, trained & $Q_4$ & 13 & 10 & 2 & --- & 3 & 32 \\
	\end{tabular}
	\caption{Sizes of the complexes of the one-hot inputs of the handcrafted and trained networks, all constructed with the same parameters.
	Totals include the $N$ vertices.
	The handcrafted monosemantic network gives the same complex for all four quadrants, so it is listed once.}
	\label{tab:appendix_trained_counts}
\end{table}

\subsection{Untrained networks}
\label{subsec:quadrant_untrained}

An untrained network has no class structure, so selecting inputs by its outputs would pick an arbitrary region of input space.
We therefore construct the complexes of the untrained networks on the one-hot inputs $X_{Q_i}$ of the trained network with the same architecture (\Cref{subsec:probe_contexts}), so that the comparison isolates the contribution of the weights.
For each architecture, we use three untrained networks with different initialisation seeds.
\Cref{tab:appendix_untrained} compares the sizes of their complexes $\mathcal{K}_{Q_i}$ with those of the trained networks.
We compare only the sizes, because the Jaccard distance matches simplices by neuron index and therefore cannot compare networks with independent parameters.

The untrained complexes are not consistently smaller than the trained ones.
For the $[2,4,4,4]$ architecture, five of the twelve untrained complexes are larger than the trained complex of the same quadrant.
For the $[2,3,4,4]$ architecture, all twelve are smaller, but the largest of them has $327$ simplices against $346$ for the trained complex of $Q_2$.

The sizes of the untrained complexes vary strongly between seeds.
For $Q_2$ of the $[2,4,4,4]$ architecture, the three seeds give between $8$ and $419$ simplices.
Two of the $24$ untrained complexes contain only the vertices.

Three of the $24$ untrained complexes reach dimension four, all of them for $Q_2$.
The trained networks reach dimension four for four of their eight input sets.

\begin{table}[htbp]
	\centering
	\begin{tabular}{l l | r | r r r}
		& & & \multicolumn{3}{c}{untrained, by seed} \\
		architecture & quadrant & trained & 0 & 1 & 2 \\
		\hline
		$[2,4,4,4]$ & $Q_1$ & 24 & 48 & 100 & 10 \\
		$[2,4,4,4]$ & $Q_2$ & 220 & 370 & 419 & 8 \\
		$[2,4,4,4]$ & $Q_3$ & 132 & 62 & 38 & 20 \\
		$[2,4,4,4]$ & $Q_4$ & 38 & 10 & 27 & 58 \\
		\hline
		$[2,3,4,4]$ & $Q_1$ & 365 & 15 & 10 & 11 \\
		$[2,3,4,4]$ & $Q_2$ & 346 & 327 & 10 & 11 \\
		$[2,3,4,4]$ & $Q_3$ & 87 & 30 & 10 & 7 \\
		$[2,3,4,4]$ & $Q_4$ & 32 & 19 & 23 & 19 \\
	\end{tabular}
	\caption{Sizes of the complexes of the one-hot inputs of the trained networks and of three untrained networks per architecture, all constructed on the same inputs.
	Totals include the $N$ vertices.}
	\label{tab:appendix_untrained}
\end{table}

\clearpage

\section{Comparing complexes across input sets}
\label{sec:input_set_comparison}

\begin{figure}[htbp]
	\centering
	\includegraphics[width=0.49\textwidth, height=.4\textheight, keepaspectratio]{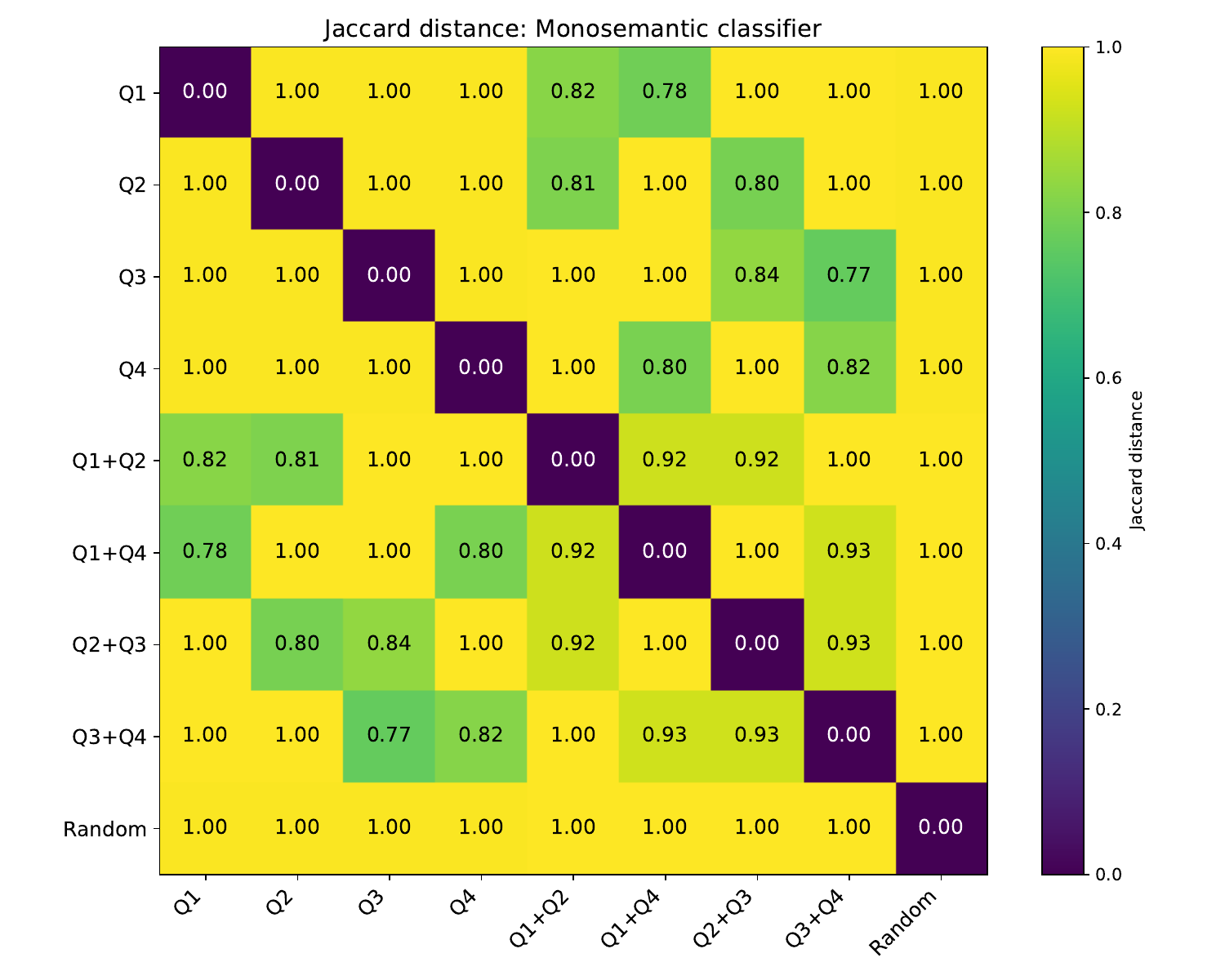}
	\includegraphics[width=0.49\textwidth, height=.4\textheight, keepaspectratio]{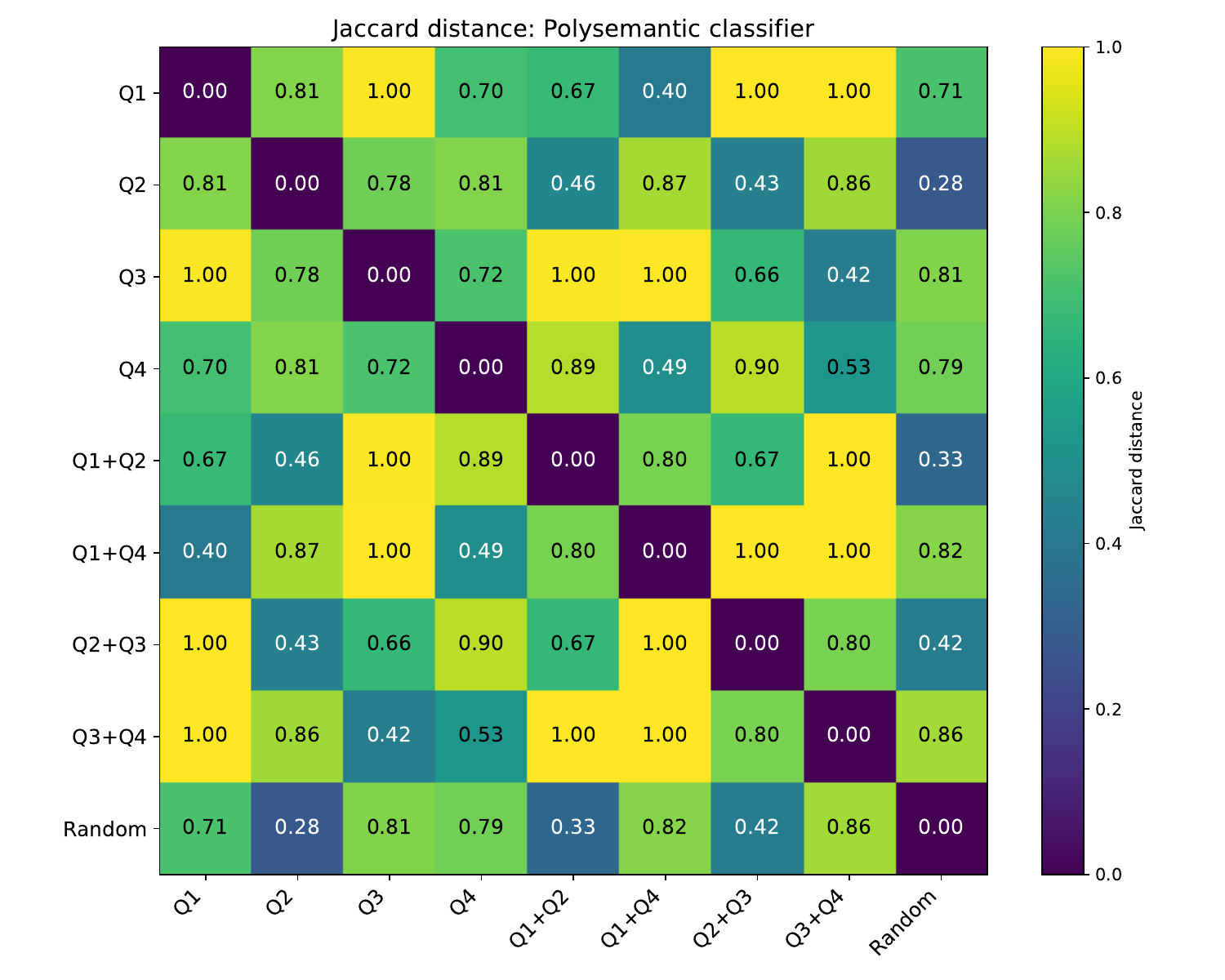}
	\caption{Jaccard distances $d_J$ between the complexes of the one-hot, adjacent two-hot and random input sets of the monosemantic (left) and polysemantic (right) networks.}
	\label{fig:appendix_jaccard_matrices_quadrant}
\end{figure}

For each network, we compute the Jaccard distance $d_J$ (\Cref{subsec:jaccard_distance}) between the complexes of all one-hot, adjacent two-hot and random input sets (\Cref{fig:appendix_jaccard_matrices_quadrant}).
All complexes are constructed as in \Cref{sec:quadrant_classifiers_details}.

\subsection{One-hot inputs}
\label{subsec:onehot_distances}

The four one-hot complexes of the monosemantic network share no edge, because the computation of each quadrant runs through its own second-layer neuron.
Their pairwise distance is nevertheless $0.996$ and not exactly $1$, because the sublevel sets of all four complexes are empty at small filtration values, where $\delta_J(\varnothing,\varnothing)=0$.

In the polysemantic network, the distances among the one-hot complexes vary between $0.70$ and $0.996$.
$\mathcal{K}_{Q_4}$ is closest to $\mathcal{K}_{Q_1}$ and $\mathcal{K}_{Q_3}$, at $0.70$ and $0.72$, because it contains both of them (\Cref{subsec:handcrafted_dsc_logic}).
$\mathcal{K}_{Q_1}$ and $\mathcal{K}_{Q_3}$ share no simplex and have the largest distance, $0.996$.

\begin{figure}[htbp]
	\centering
	\includegraphics[width=\linewidth]{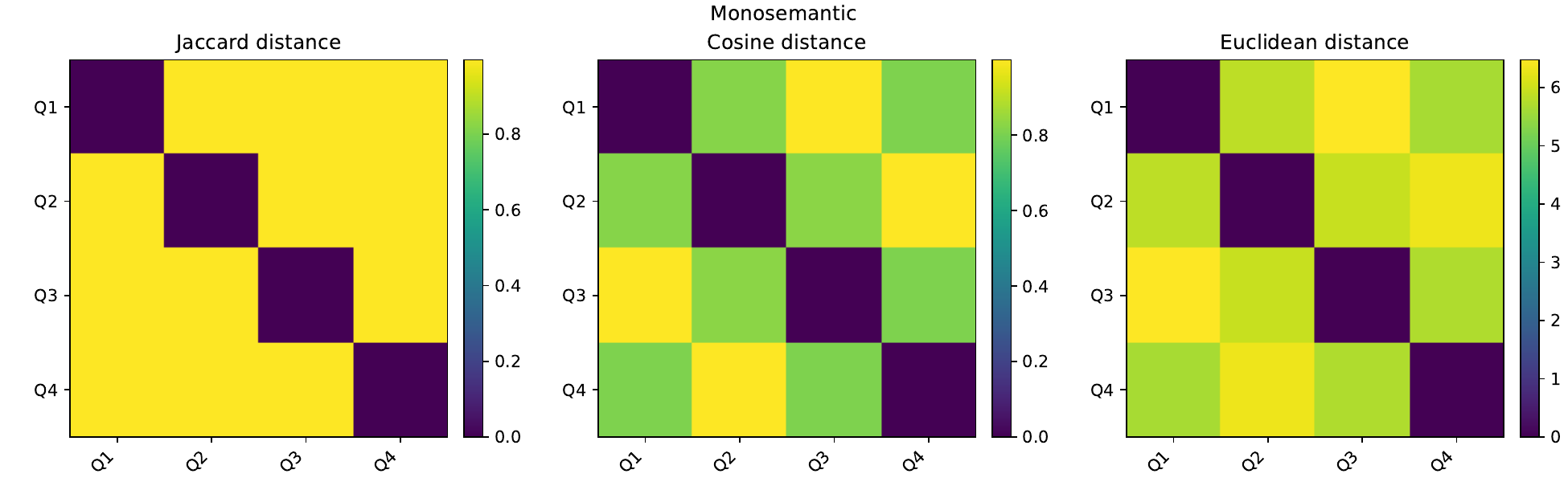}

	\vspace{0.6em}
	\includegraphics[width=\linewidth]{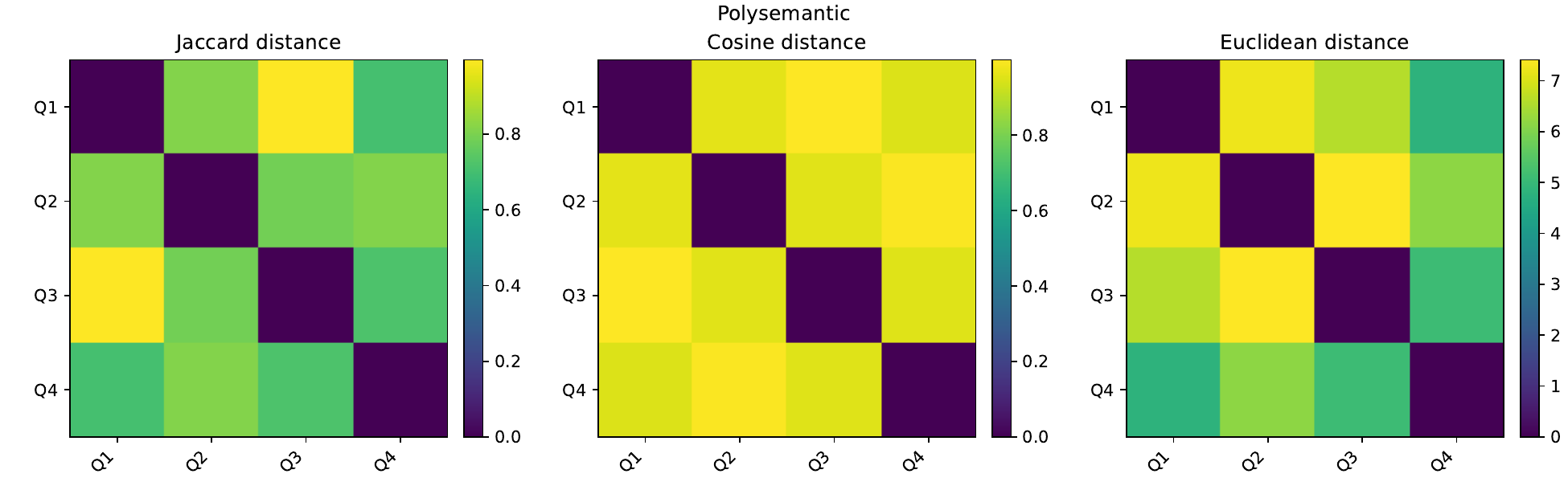}
	\caption{Distances between the one-hot input sets of the monosemantic (top) and polysemantic (bottom) networks.
	Left: Jaccard distance $d_J$ between their complexes.
	Centre and right: cosine and Euclidean distance between the mean activations of the internal neurons over the same inputs.}
	\label{fig:appendix_dsc_vs_activation}
\end{figure}

\begin{figure}[htbp]
	\centering
	\includegraphics[width=\linewidth]{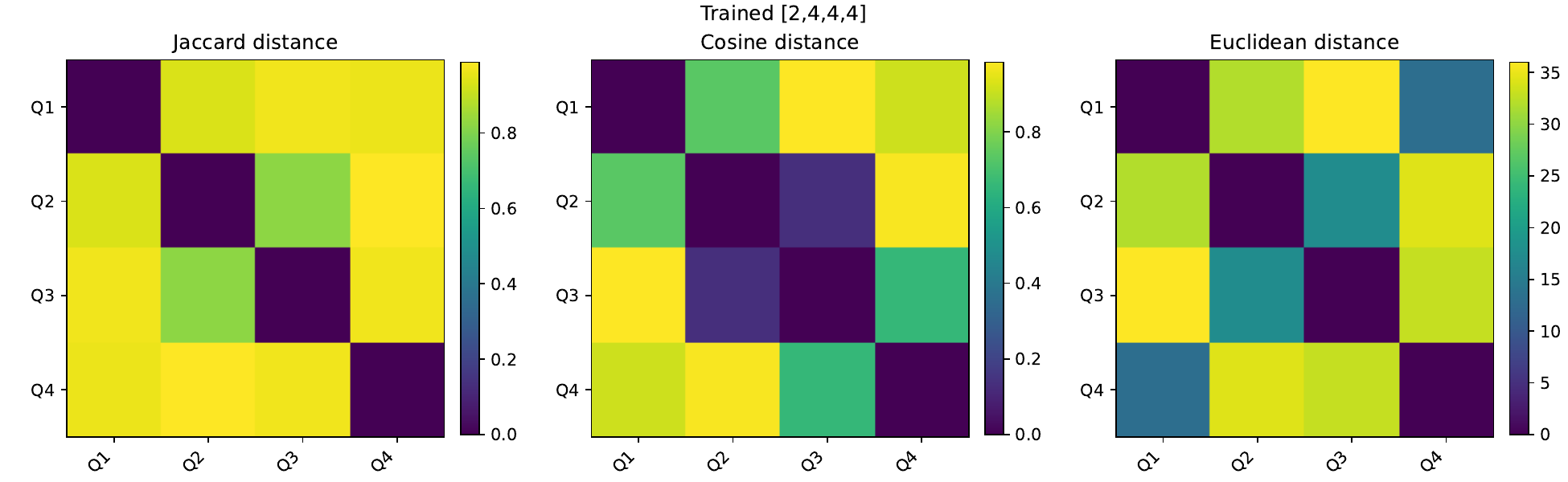}

	\vspace{0.6em}
	\includegraphics[width=\linewidth]{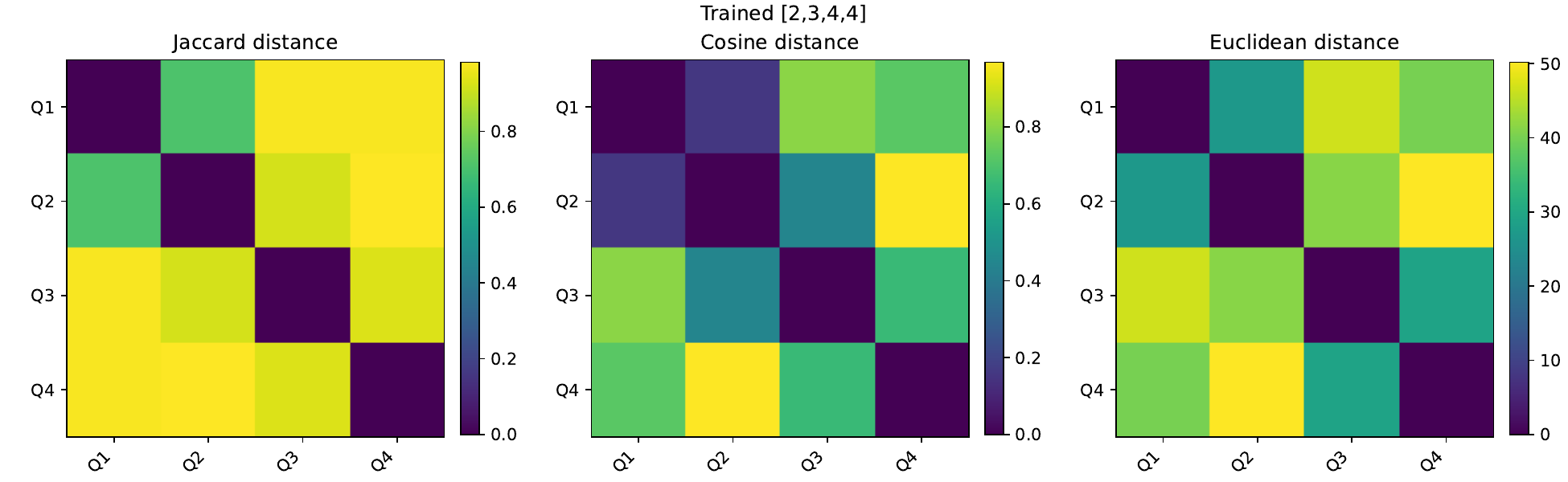}
	\caption{Distances between the one-hot input sets of the trained $[2,4,4,4]$ (top) and $[2,3,4,4]$ (bottom) networks, drawn as in \Cref{fig:appendix_dsc_vs_activation}.}
	\label{fig:appendix_dsc_vs_activation_trained}
\end{figure}

We compare these distances with distances between the mean activations of the internal neurons over the same input sets.
For each pair of one-hot input sets, \Cref{fig:appendix_dsc_vs_activation,fig:appendix_dsc_vs_activation_trained} show the Jaccard distance between their complexes and the cosine and Euclidean distances between their mean activation vectors.

In the monosemantic network, the cosine distance is between $0.81$ and $0.83$ for adjacent quadrants and $1.0$ for opposite quadrants, while the Jaccard distance is the same for all pairs.
Adjacent quadrants share one active first-layer neuron, and opposite quadrants share none.

In the polysemantic network, the Jaccard and cosine distances order the six pairs of input sets in the same way, with Spearman rank correlation $\rho=1$.
The same holds for the trained $[2,3,4,4]$ network.
For the trained $[2,4,4,4]$ network, $\rho=0.77$, so its Jaccard distance is not a monotone function of the cosine distance.

The activation distances of the trained networks are smaller than those of the handcrafted networks with the same architecture, with a mean cosine distance of $0.73$ against $0.88$ for $[2,4,4,4]$ and $0.62$ against $0.97$ for $[2,3,4,4]$.
The mean Jaccard distance of the trained network is smaller for $[2,4,4,4]$, $0.94$ against $0.996$, but larger for $[2,3,4,4]$, $0.91$ against $0.80$.

\subsection{Two-hot inputs}
\label{subsec:twohot_distances}

For each pair of adjacent quadrants $Q_i$ and $Q_j$, we compare the distances of $\mathcal{K}_{Q_iQ_j}$ to $\mathcal{K}_{Q_i}$ and $\mathcal{K}_{Q_j}$ with its distances to the other two one-hot complexes.
We say that the pair is recovered if $\mathcal{K}_{Q_i}$ and $\mathcal{K}_{Q_j}$ are both closer than the other two.
As the \emph{separation} of the pair, we report the smallest distance to the other two one-hot complexes minus the largest distance to $\mathcal{K}_{Q_i}$ and $\mathcal{K}_{Q_j}$, so a positive separation means that the pair is recovered.

\begin{table}[htbp]
	\centering
	\begin{tabular}{l | c c | c}
		network & pairs recovered & smallest separation & one-hot distances \\
		\hline
		$[2,4,4,4]$, handcrafted & 4/4 & $+0.16$ & $0.996$ \\
		$[2,3,4,4]$, handcrafted & 4/4 & $+0.22$ & $0.70$--$0.996$ \\
		$[2,4,4,4]$, trained     & 3/4 & $-0.15$ & $0.82$--$0.99$ \\
		$[2,3,4,4]$, trained     & 3/4 & $-0.037$ & $0.71$--$0.98$ \\
	\end{tabular}
	\caption{Recovery of the adjacent two-hot input sets from the Jaccard distance.
	A pair is recovered when $\mathcal{K}_{Q_iQ_j}$ is closer to both $\mathcal{K}_{Q_i}$ and $\mathcal{K}_{Q_j}$ than to either of the other two one-hot complexes.
	The smallest separation is taken over the four adjacent pairs.
	The last column gives the range of the distances among the four one-hot complexes.}
	\label{tab:appendix_twohot_composition}
\end{table}

Both handcrafted networks recover all four pairs, and the separation is larger for the polysemantic network.
For the monosemantic network, the distances of the two-hot complexes to their own one-hot complexes lie between $0.77$ and $0.84$, and those to the other two one-hot complexes are $0.997$.

Each trained network recovers three of the four pairs.
For the trained $[2,4,4,4]$ network, $\mathcal{K}_{Q_1Q_2}$ is closer to $\mathcal{K}_{Q_3}$ than to $\mathcal{K}_{Q_1}$.
For the trained $[2,3,4,4]$ network, $\mathcal{K}_{Q_1Q_4}$ is closer to $\mathcal{K}_{Q_2}$ and $\mathcal{K}_{Q_3}$ than to $\mathcal{K}_{Q_4}$.

We also compare $\mathcal{K}_{Q_iQ_j}$ with the union of $\mathcal{K}_{Q_i}$ and $\mathcal{K}_{Q_j}$, which contains every simplex of either complex with the smaller of its two filtration values.
The distance between the two-hot complex and this union lies between $0.62$ and $0.68$ for the monosemantic network and between $0.49$ and $0.54$ for the polysemantic network.
For the monosemantic network, the union is closer to $\mathcal{K}_{Q_iQ_j}$ than either one-hot complex.
For the polysemantic network, one of the two one-hot complexes is closer than the union for all four pairs.

\subsection{Random inputs}
\label{subsec:random_distances}

The random inputs come from all four quadrants, and the relevance test pools the responses of all of them.
A simplex whose target changes on many of the inputs can therefore still be admitted, while a simplex that is specific to one quadrant is diluted by the inputs on which its predictors have no effect.

In the monosemantic network, every simplex of the one-hot complexes occurs in only one of them, and $\mathcal{K}_{\mathrm{rand}}$ contains no edge.
Its distance to all other complexes is therefore close to $1$.

\begin{figure}[htbp]
	\centering
	\includegraphics[width=.9\linewidth, keepaspectratio]{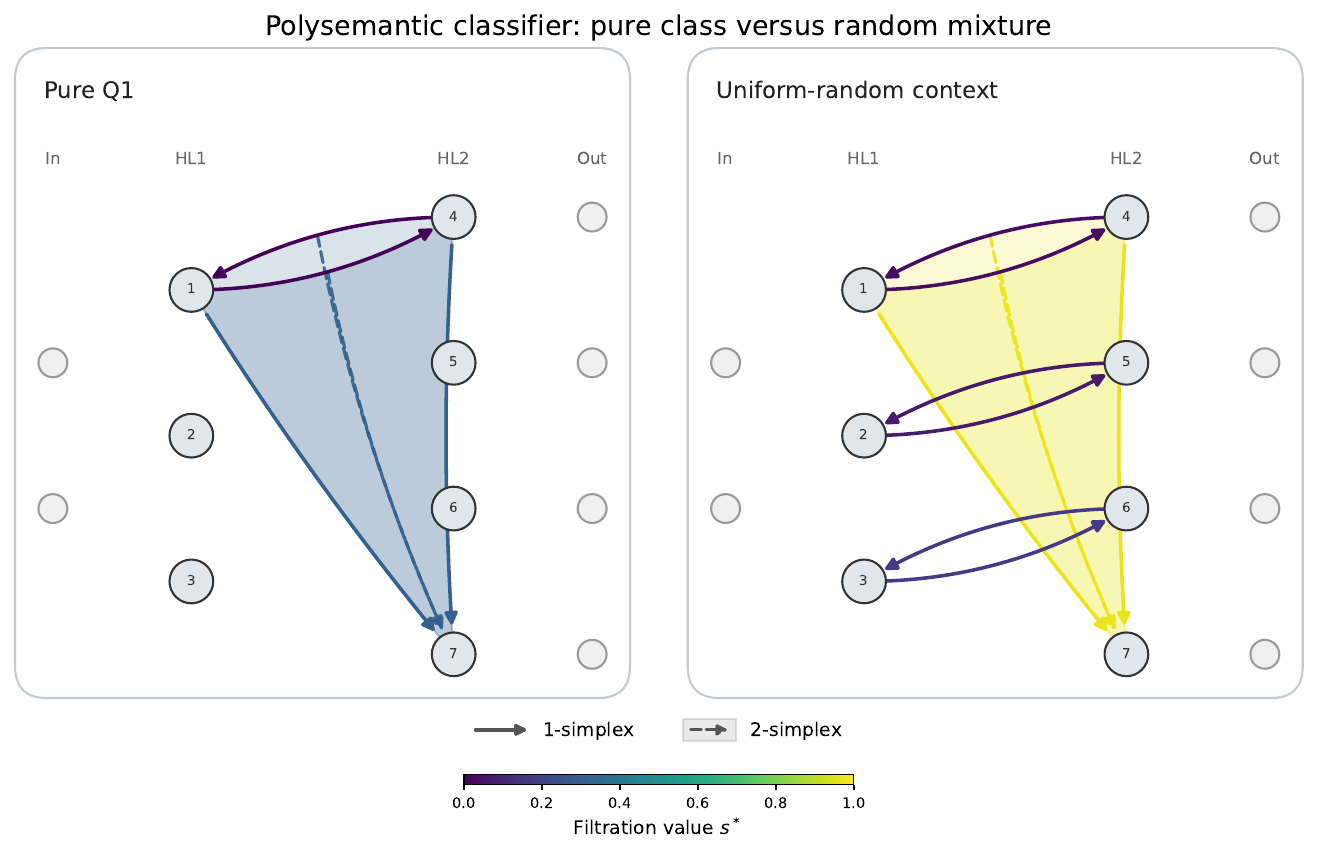}
	\caption{The complexes $\mathcal{K}_{Q_1}$ and $\mathcal{K}_{\mathrm{rand}}$ of the polysemantic network.
	Every simplex of $\mathcal{K}_{Q_1}$ also occurs in $\mathcal{K}_{\mathrm{rand}}$, which in addition contains the edges between $n_2$ and $n_5$ and between $n_3$ and $n_6$, in both directions.
	The colour shows the difference in filtration value, and the shared 2-simplices enter at $0.31$ in $\mathcal{K}_{Q_1}$ but only at $0.97$ in $\mathcal{K}_{\mathrm{rand}}$.}
	\label{fig:appendix_poly_random_superposition}
\end{figure}

In the polysemantic network, $\mathcal{K}_{\mathrm{rand}}$ contains the edges between each first-layer neuron and its copy, in both directions, as does $\mathcal{K}_{Q_2}$.
Its only other simplices are the edges $n_1\to n_7$ and $n_4\to n_7$ and the 2-simplices $(n_1,n_4,n_7)$ and $(n_4,n_1,n_7)$.
It is closest to $\mathcal{K}_{Q_2}$, at a distance of $0.28$, and contains $\mathcal{K}_{Q_1}$ as a subcomplex (\Cref{fig:appendix_poly_random_superposition}).

The simplices of $\mathcal{K}_{Q_1}$ enter much later in $\mathcal{K}_{\mathrm{rand}}$.
The edges between $n_1$ and $n_4$ move from $0.004$ to $0.033$, and the 2-simplices $(n_1,n_4,n_7)$ and $(n_4,n_1,n_7)$ from $0.31$ to $0.97$.
The distance between $\mathcal{K}_{Q_1}$ and $\mathcal{K}_{\mathrm{rand}}$ is therefore $0.71$, although one contains the other.

$\mathcal{K}_{\mathrm{rand}}$ contains the edges to $n_7$ from $n_1$ and $n_4$ but not from $n_3$ and $n_6$, although the network is symmetric under the reflection in $x+y=0$, which exchanges $n_1$ with $n_3$ and $n_4$ with $n_6$ (\Cref{subsec:handcrafted_dsc_logic}).
Since the construction is symmetric under this reflection as well, the difference comes from the input set.
Of the $128$ random inputs, $70$ lie on the side $x+y>0$, where $n_1$ is active, and $58$ on the side where $n_3$ is active.

\clearpage

\section{Comparison with other interpretability methods}
\label{sec:methods_comparisons}

Circuit-discovery methods identify the connections of a network that its behaviour on a task depends on, and they return a subgraph of the network.
We compare the filtered directed simplicial complexes (DSCs) of the handcrafted networks with four such methods.
Since a graph cannot contain simplices of dimension two or higher, we restrict the comparison to the edges of the DSCs.
The comparison therefore shows only whether the two kinds of method agree on the edges that both can express.

\subsection{Representable edges}
\label{subsec:representable_ceiling}

A circuit-discovery method intervenes on the activations of the network and measures the effect on its output.
Its graph can therefore contain an edge from $n_i$ to $n_j$ only if $n_i$ feeds $n_j$ directly.
A DSC suppresses neurons by moving the inputs (\Cref{subsec:counterfactual_generation}) and measures the effect on the other neurons.
Its edges can therefore connect any two neurons.
For example, suppressing a second-layer neuron moves the inputs and thereby changes the first-layer neurons, which can give an edge from the second to the first hidden layer.

We call an edge of a DSC forward-adjacent if its predictor feeds its target directly, within-layer if both lie in the same hidden layer, and backward if it points from the second to the first hidden layer.
Only the forward-adjacent edges can occur in a circuit.
In the monosemantic network, four of the ten edges of each one-hot complex are forward-adjacent, two are within-layer and four are backward.
In the polysemantic network, the forward-adjacent edges make up half of the edges of $\mathcal{K}_{Q_1}$, $\mathcal{K}_{Q_2}$ and $\mathcal{K}_{Q_3}$ and a third of the edges of $\mathcal{K}_{Q_4}$ (\Cref{tab:appendix_silver_standard}).

\subsection{Circuit-discovery methods}
\label{subsec:baseline_implementations}

Automatic circuit discovery (ACDC) replaces the contribution of one connection at a time by its value on a corrupted input, and it removes the connection if this changes the output by less than a threshold \citep{Conmy2023}.
Edge attribution patching (EAP) estimates the effect of every such replacement from the gradient at the clean input \citep{Syed2023}.
Edge attribution patching with integrated gradients (EAP-IG) averages this gradient along the path from the corrupted to the clean input \citep{Hanna2024}.
We apply these methods to the connections from the first to the second hidden layer.
ACDC measures the change of the output by the Kullback--Leibler (KL) divergence, EAP by the difference between the logit of the quadrant and the mean of the other logits, and EAP-IG by either of the two.

As corrupted inputs, we negate a random nonempty subset of the coordinates of each input, which moves it into another quadrant.
This lowers the accuracy from $1.0$ to $0.0$ for the four one-hot input sets of the monosemantic network and for $X_{Q_2}$ of the polysemantic network, and to between $0.17$ and $0.22$ for the other three.
EAP and EAP-IG score every connection, and we keep as many of the highest-scoring connections as ACDC keeps for the same input set, which is between three and ten.

\Cref{tab:appendix_silver_standard} compares the forward-adjacent DSC edges with the connections that each method keeps.
In $26$ of the $32$ comparisons, the method keeps every forward-adjacent DSC edge.
EAP-IG with the logit difference does so for all eight input sets, ACDC and EAP-IG with the KL divergence for seven, and EAP for four.
A random choice of the same number of connections would keep all forward-adjacent DSC edges with a probability between $0.002$ and $0.12$, depending on the input set.

\begin{figure}[p]
	\centering
	\includegraphics[width=\linewidth]{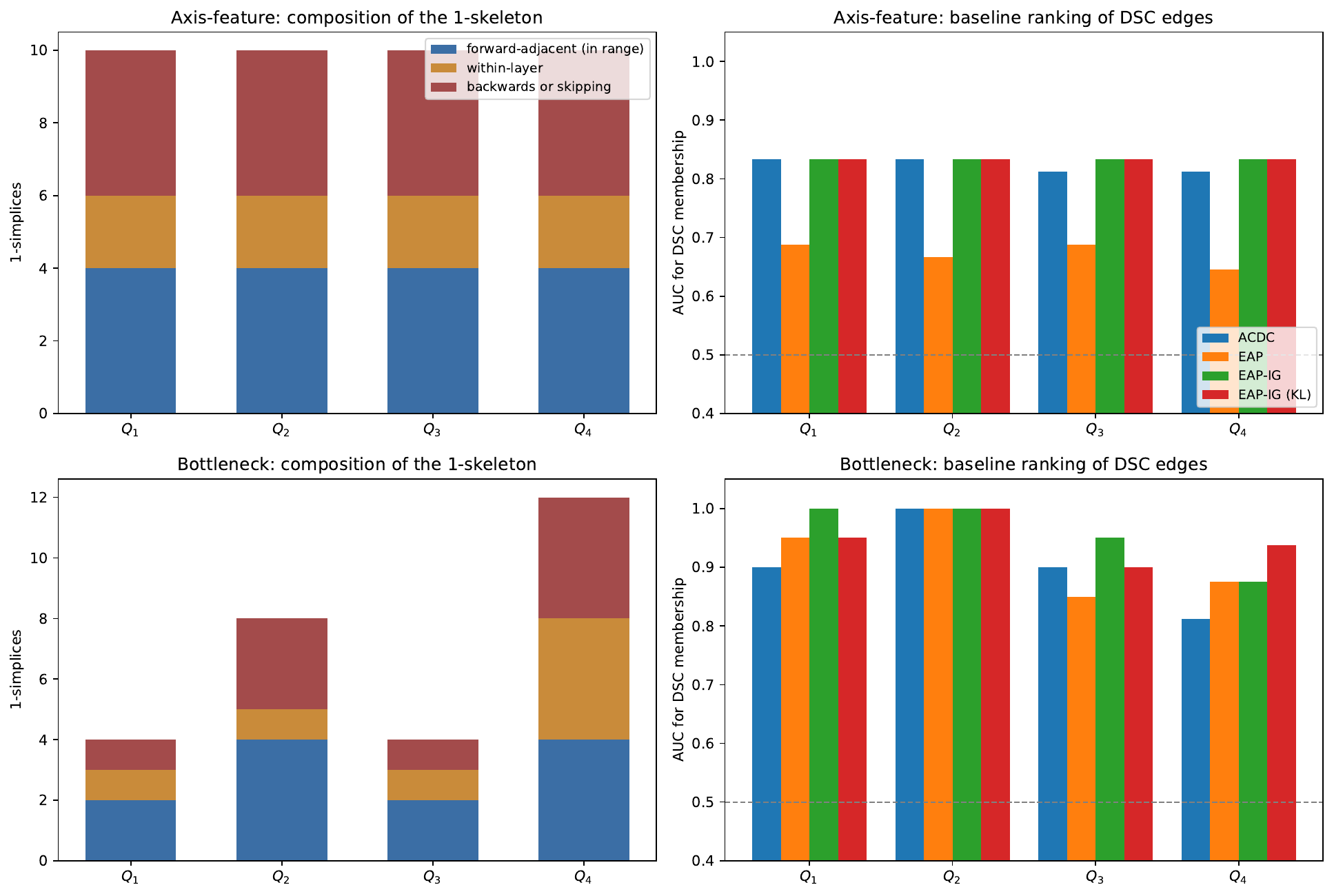}
	\caption{Edges of the one-hot complexes of the monosemantic (top) and polysemantic (bottom) networks.
	Left: numbers of forward-adjacent, within-layer and backward edges.
	Right: AUC of the scores of the four circuit-discovery methods for membership in the DSC, where a random score gives $0.5$.}
	\label{fig:appendix_silver_standard}
\end{figure}

The scores of the methods also rank the forward-adjacent DSC edges above the other connections (\Cref{fig:appendix_silver_standard}).
The area under the receiver operating characteristic curve (AUC) of the scores for membership in the DSC is $0.86$ on average and at least $0.65$, where a random score gives $0.5$.

Except for $X_{Q_2}$ of the polysemantic network, the methods keep more connections than the DSC has forward-adjacent edges.
Of the $103$ connections that a method keeps and the DSC lacks, summed over the $32$ comparisons, $68$ start at a first-layer neuron that is inactive on the whole one-hot input set, and $16$ at one that is active on at most a fifth of it.
An inactive neuron cannot be suppressed, so the DSC contains no edge from it.
The corrupted inputs lie in other quadrants, where these neurons can be active.

\begin{table}[htbp]
	\centering
	\footnotesize
	\begin{tabular}{l c c c c | c c | c c | c c | c c}
		& & \multicolumn{3}{c|}{DSC} & \multicolumn{2}{c|}{ACDC}
		& \multicolumn{2}{c|}{EAP} & \multicolumn{2}{c|}{EAP-IG}
		& \multicolumn{2}{c}{EAP-IG (KL)} \\
		network & & $\abs{E}$ & f/w/b & share
		& rec & AUC & rec & AUC & rec & AUC & rec & AUC \\
		\hline
		mono & $Q_1$ & 10 & 4/2/4 & 40\% & 1.00 & 0.83 & 1.00 & 0.69 & 1.00 & 0.83 & 1.00 & 0.83 \\
		mono & $Q_2$ & 10 & 4/2/4 & 40\% & 1.00 & 0.83 & 0.50 & 0.67 & 1.00 & 0.83 & 1.00 & 0.83 \\
		mono & $Q_3$ & 10 & 4/2/4 & 40\% & 1.00 & 0.81 & 0.75 & 0.69 & 1.00 & 0.83 & 1.00 & 0.83 \\
		mono & $Q_4$ & 10 & 4/2/4 & 40\% & 1.00 & 0.81 & 0.75 & 0.65 & 1.00 & 0.83 & 1.00 & 0.83 \\
		poly & $Q_1$ & 4 & 2/1/1 & 50\% & 1.00 & 0.90 & 1.00 & 0.95 & 1.00 & 1.00 & 1.00 & 0.95 \\
		poly & $Q_2$ & 8 & 4/1/3 & 50\% & 1.00 & 1.00 & 1.00 & 1.00 & 1.00 & 1.00 & 1.00 & 1.00 \\
		poly & $Q_3$ & 4 & 2/1/1 & 50\% & 0.50 & 0.90 & 0.50 & 0.85 & 1.00 & 0.95 & 0.50 & 0.90 \\
		poly & $Q_4$ & 12 & 4/4/4 & 33\% & 1.00 & 0.81 & 1.00 & 0.88 & 1.00 & 0.88 & 1.00 & 0.94 \\
	\end{tabular}
	\caption{Forward-adjacent edges of the one-hot complexes against four circuit-discovery methods that keep the same number of connections.
	$\abs{E}$ is the number of edges of the DSC, f/w/b their split into forward-adjacent, within-layer and backward edges, and ``share'' the share of forward-adjacent edges.
	``rec'' is the fraction of the forward-adjacent DSC edges that the method keeps, and AUC that of its scores for membership in the DSC.}
	\label{tab:appendix_silver_standard}
\end{table}

\subsection{Other interpretability methods}
\label{subsec:methods_not_compared}

Besides circuit-discovery methods, several other interpretability methods explain a network through interventions or attributions.
Causal mediation analysis measures how much of an effect on the output passes through individual neurons by setting their activations to the values for another input \citep{Vig2020}, and integrated gradients attribute an output to the inputs by integrating gradients along a path \citep{Sundararajan2017}.
Applied to the connections between neurons, these become activation patching and integrated gradients on connections, which ACDC and EAP-IG already implement.
Subnetwork probing learns a sparse mask over the weights such that the remaining subnetwork still solves a given task \citep{Cao2021}, so it returns a subgraph and would enter the comparison like the four methods above.
Distributed alignment search tests whether the variables of a hypothesised causal model are represented in subspaces of the activations \citep{Geiger2024,Geiger2025}, and testing with concept activation vectors measures how sensitive an output is to a concept given by labelled examples \citep{Kim2018a}.
Both locate a given variable or concept in the activations and do not relate neurons to each other.
A comparison with DSCs would first need a rule that translates their results into edges and simplices, which we leave for future work.

\end{document}